%% file: main.tex
\documentclass{article} 
\usepackage{collas2023_conference,times}

\usepackage{easyReview}

\input{math_commands.tex}

\usepackage[utf8]{inputenc} 
\usepackage[T1]{fontenc}    
\usepackage{url}            
\usepackage{booktabs}       
\usepackage{amsfonts}       
\usepackage{nicefrac}       
\usepackage{microtype}      
\usepackage{xcolor}         

\usepackage{float}
\usepackage{hyperref}
\usepackage{graphicx}
\usepackage{caption}
\usepackage{subcaption}

\usepackage{algorithm}
\usepackage{algorithmic}
\usepackage{wrapfig}

\usepackage[ruled, lined, longend,algo2e]{algorithm2e}

\SetCommentSty{mycommfont}
\newcommand\negspace[1]{\vspace{#1}}
\usepackage{setspace}
\usepackage{amsmath}
\usepackage{amssymb}
\usepackage{mathtools}
\usepackage{amsthm}
\usepackage{multirow}
\usepackage{color}

\usepackage[capitalize]{cleveref}
\crefname{section}{Sec.}{Secs.}
\Crefname{section}{Section}{Sections}
\Crefname{table}{Table}{Tables}
\crefname{table}{Tab.}{Tabs.}
\hypersetup{
    colorlinks=true,
    linkcolor=red,
    filecolor=magenta,
    urlcolor=blue,
    citecolor=purple,
    pdftitle={Overleaf Example},
    pdfpagemode=FullScreen,
    }

\input{custom_shortcut}

\title{Continual Learning Beyond a Single Model}

\author{Thang Doan  \thanks{Work done at McGill University \\ corresponding author: \texttt{thang.doan@mail.mcgill.ca}} \\
Bosch Research North America $\&$  \\
Bosch Center for Artificial Intelligence (BCAI)
\And 
Seyed Iman Mirzadeh  \\
Washington State University \\
\And 
Mehrdad Farajtabar\thanks{Work done at DeepMind.} \\
Apple \\
}

\collasfinalcopy 

\begin{document}

\maketitle

\begin{abstract}
A growing body of research in continual learning focuses on the catastrophic forgetting problem. While many attempts have been made to alleviate this problem, the majority of the methods assume a \textit{single model} in the continual learning setup. In this work, we question this assumption and show that employing \textit{ensemble models} can be a simple yet effective method to improve continual performance. However, ensembles' training and inference costs can increase significantly as the number of models grows. Motivated by this limitation, we study different ensemble models to understand their benefits and drawbacks in continual learning scenarios. Finally, to overcome the high compute cost of ensembles, we leverage recent advances in neural network subspace to propose a computationally cheap algorithm with similar runtime to a single model yet enjoying the performance benefits of ensembles.
\end{abstract}

\input{1-introduction.tex}
\input{1a-preliminary.tex}

\input{2-beyond-single-model.tex}
\input{4-methodology.tex}

\input{5-experiments.tex}

\input{6-related_works.tex}
\input{7-conclusion.tex}

\clearpage

\input{main.bbl}
\appendix
\input{appendix}

\end{document}

%% file: math_commands.tex
\usepackage{amsmath,amsfonts,bm}

\def\eqref#1{equation~\ref{#1}}

\def\1{\bm{1}}

\DeclareMathAlphabet{\mathsfit}{\encodingdefault}{\sfdefault}{m}{sl}
\SetMathAlphabet{\mathsfit}{bold}{\encodingdefault}{\sfdefault}{bx}{n}



%% file: custom_shortcut.tex
\newcommand{\cB}{\mathcal{B}}

\newcommand{\cT}{\mathcal{T}}

\newcommand{\algoname}{Subspace-Connectivity }

%% file: 1-introduction.tex
\vspace{-3mm}
\section{Introduction}
\label{sec:introduction}
Continual learning (CL) and Lifelong learning~\citep{thrun} have recently gained popularity since many real-world applications fall into that setting. It describes the scenario where not only a stream of data arrives sequentially, but their distribution also changes over time. This setup induces Catastrophic Forgetting (CF)~\citep{MCCLOSKEY1989109} which is a degradation of performances on previous data due to distribution shift between tasks~\citep{doan2021theoretical}.

One fundamental goal in continual learning is to learn from the new incoming tasks while retaining knowledge from the past and avoiding interference that can lead to poor performance~\citep{lesort2021continual}. This becomes particularly challenging when the stream of data increases because all the burden is left to a single model. A simple yet effective solution is to rely on an ensemble method that improves performance over a single model. Inspired by bootstrapping~\citep{breiman1996bagging}, deep ensembles initialize and train multiple neural networks independently~\citep{lakshminarayanan2017simple,fort2019deep}. Not only does this improve their robustness, but it also boosts the overall performance thanks to a higher diversity in the solutions and their de-correlated predictions ~\citep{goodfellow2014explaining,havasi2020training}. This simple mechanism allows ensemble methods to improve performance over single models \citep{huang2017snapshot}. From now on, we will refer to this method as Vanilla Ensemble (VE). It is well known that ensemble methods perform well in supervised learning~\citep{dietterich2000ensemble,fort2019deep}. However, their functionality has not been fully studied in continual learning scenarios.

\begin{wrapfigure}{h!}{0.4\textwidth}
\centering
\includegraphics[width=0.4\textwidth]{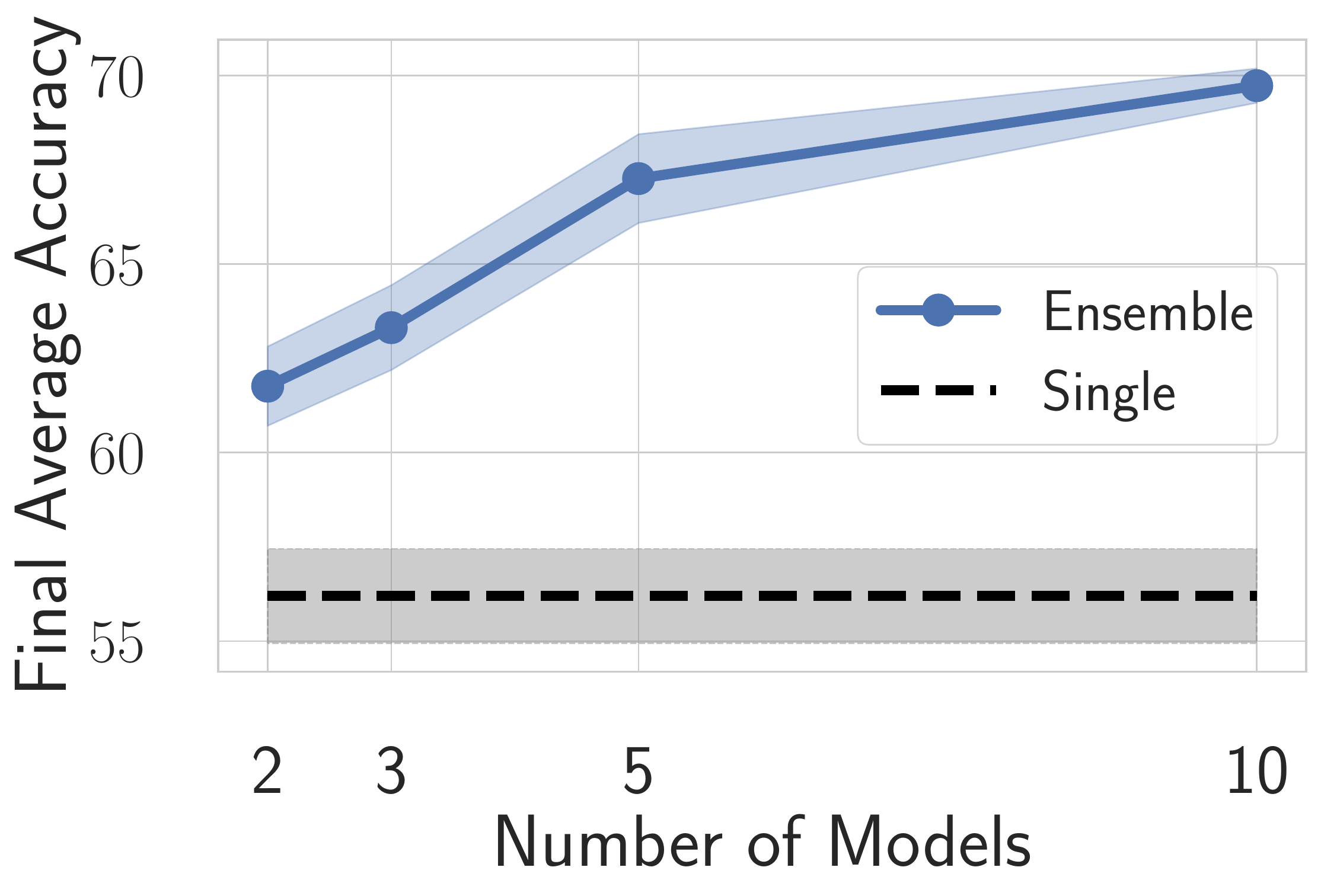}
\caption{Split CIFAR-100: As the ensemble size grows, the continual learning performance improves.}
\label{fig:intro-acc-cifar}
\vspace{-20pt}
\end{wrapfigure}

\begin{figure*}[htbp]
     \centering
     
  \hspace*{-12pt} 
   \includegraphics[scale=0.34,keepaspectratio=True]{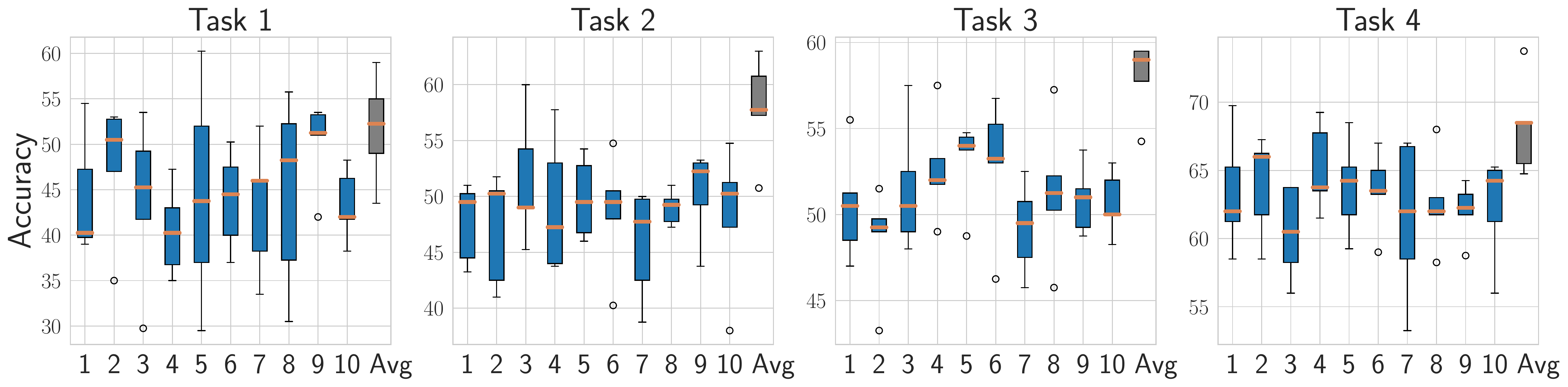}
    \caption{ Final accuracy of continual learning after finishing all the tasks for each of the $10$ members of the ensemble (blue) and the final ensemble model (gray) for Split CIFAR-100. The final ensemble prediction accuracy is almost always higher than the accuracy of the best model. Moreover, each model specializes in different tasks throughout the training contributing to the performance of the ensemble method.}
   
    \label{fig:ensemble_diversity_cifar}
\end{figure*}

In the context of continual learning, using an ensemble method and having diversity in the solutions allows each model to have a good solution in different tasks and lead to good performance for the ensemble as shown in Fig.~\ref{fig:ensemble_diversity_cifar}. Given that only some of the individual models may need a drastic or a slight change to learn the incoming tasks, it leads to an attenuation of forgetting and a boost in the overall performance (Fig.~\ref{fig:intro-acc-cifar}). Recently by~\cite{caccia2021anytime} also showed empirically good performance of ensemble methods in their ``anytime learning`` framework where data arrive by batches instead of a whole dataset.

However, the computational cost of ensembles grows linearly with the number of models. This limits their usage for real-world applications due to computational or environmental concerns.

This issue becomes significantly more challenging on resource-constrained systems such as edge devices~\citep{habitat19iccv,szot2021habitat}. While it is possible to focus solely on the computational cost problem to find an efficient parameter sharing algorithm~\citep{Wen2020BatchEnsemble}, in this work, we focus on both the computational efficiency and continual learning performance together. Our work is motivated by the recent advances in the deep learning optimization literature, such as Neural Network Subspace~\citep{learning-neural-network-subspaces} and Mode Connectivity~\citep{modeconnectivity_main,draxler2018essentially,mirzadeh2021MCSGD}.

\textbf{Contributions.} 
In this work, we study continual learning with various ensemble methods. Our contributions can be summarized as three-fold:
we show a) that ensembling is a simple technique to boost continual learning performance, but significantly increasing the computation cost. To alleviate this issue, we borrow insights from the recent advances mode-connectivity to propose b) a method with a similar computation cost to a single model yet enjoys the high-performance benefits of ensembles. Our proposed efficient ensembling method learns tasks continually in the subpaces of neural networks~\citep{learning-neural-network-subspaces} to divide the learning capacity between tasks without causing interference, and relies on the connectivity of the tasks' optima~\citep{modeconnectivity_main} for better retention of the previous knowledge. 

%% file: 1a-preliminary.tex
    \negspace{-2mm}

\section{Preliminaries}
    \negspace{-2mm}


\textbf{Notation.} Let $\mathcal{X}$ be some features and $\mathcal{Y}$ the labels space ($\mathcal{Y}=\mathbb{R}$ for a regression problem and $\mathcal{Y} \in \Delta^{K}$ for classification problem\footnote{$\Delta^{K}$ denotes the vertices of the $K$-dim probability simplex}). In CL, a stream of supervised learning tasks indexed by $\tau \in [T]$, $\mathcal{T}_{\tau}$, ( where $T \in \mathbb{N}^{*}$ is the total number of tasks) arrives sequentially. The goal is to learn a predictor $f_{\omega}: \mathcal{X} \times \mathcal{T} \rightarrow \mathcal{Y}$ (where $\omega \in \mathbb{R}^{p}$ are the learnable parameters of size $p$) that performs a prediction as accurate as possible with respect to a loss function $\mathcal{L}_{\tau}$. We denote the weight learned after task $\tau$ as $\omega^{*}_{\tau}$. In the framework of CL, one cannot recover samples from previous tasks unless storing them in a replay buffer~\citep{Chaudhry2019OnTE}. A pseudo-code of the Vanilla Continual Learning Algorithm (single model) is provided in Appendix (Alg.~\ref{alg:single_cl}).

\subsection{Experimental Setup}
For our experiments in Sec.~\ref{sec:2-beyond-one-model}, we will be using the following setup:\\
\textbf{Benchmarks.}
For this ablation, we use two standard benchmarks following~\citep{Goodfellow2013Forgetting}, and \citep{Chaudhry2019OnTE}. Rotated MNIST \citep{farajtabar2020orthogonal} consists of a series of MNIST classification tasks, where the images are rotated with respect to a fixed angle, monotonically. We increment the rotation angle by $22$ degrees at each new task like in \cite{mirzadeh2021MCSGD} to worsen the catastrophic forgetting phenomenon. The task ID does not need to be provided since it is a domain incremental task. Split CIFAR-100 \citep{Chaudhry2019OnTE} is constructed by splitting the original CIFAR-100 dataset \citep{Krizhevsky2009LearningML} into 20 disjoint subsets, where each subset is formed by sampling without replacement of 5 classes out of 100. For the sake of our experiment, we fine-tune on a sequence of $5$ tasks for each benchmark, seeing the tasks only once without relying on a replay buffer. For this dataset, the task ID is provided to the model. While for brevity, we include the main results in this section, more detailed plots can be found in Appendix~\ref{sec:ablation_details}.\\
\textbf{Architectures.}
The neural network architectures used are respectively fully connected layers with two hidden layers of $100$ hidden units (Rotated MNIST) and a reduced Resnet-18 with three times fewer filters map across all layers as in \cite{understanding_continual}.\\
\textbf{Metrics.}
To assess the performance of each baseline, we report the Final Accuracy and Forgetting Measure defined as follows.
The Final Accuracy after $T$ tasks is the average validation accuracy over all the tasks $\tau=1...T$ defined as: $A_T=\frac{1}{T}\sum_{\tau=1}^{T}a_{T,\tau}$ where $a_{T,\tau}$ is the validation accuracy of task $\tau$ after the model finished learning on task $T$ (at test time). The Learning Accuracy is defined as $a_{\tau,\tau}$, this describes how well a model learns a task $\tau$ the first time it sees it. The Forgetting Measure is defined as:
$F_T=\frac{1}{T-1}\sum_{\tau=1}^{T-1} \max_{t=\{1..T-1 \}} (a_{t,\tau}-a_{T,\tau})$. Finally, we define the Forgetting Improvement (FI) simply as the difference between the Forgetting Measure of the single model and the ensemble (or subspace) method as: $FI_{T}=F_{T}(\textit{single model})-F_{T}(\textit{ensemble/subspace model})$. Intuitively, the higher this value the less forgetting a method has compared to the single model case.\\

\begin{figure*}[h!]
\centering
\begin{subfigure}{.49\textwidth}
    \centering
    \includegraphics[width=0.9\linewidth,height=1.0\linewidth,keepaspectratio=True]{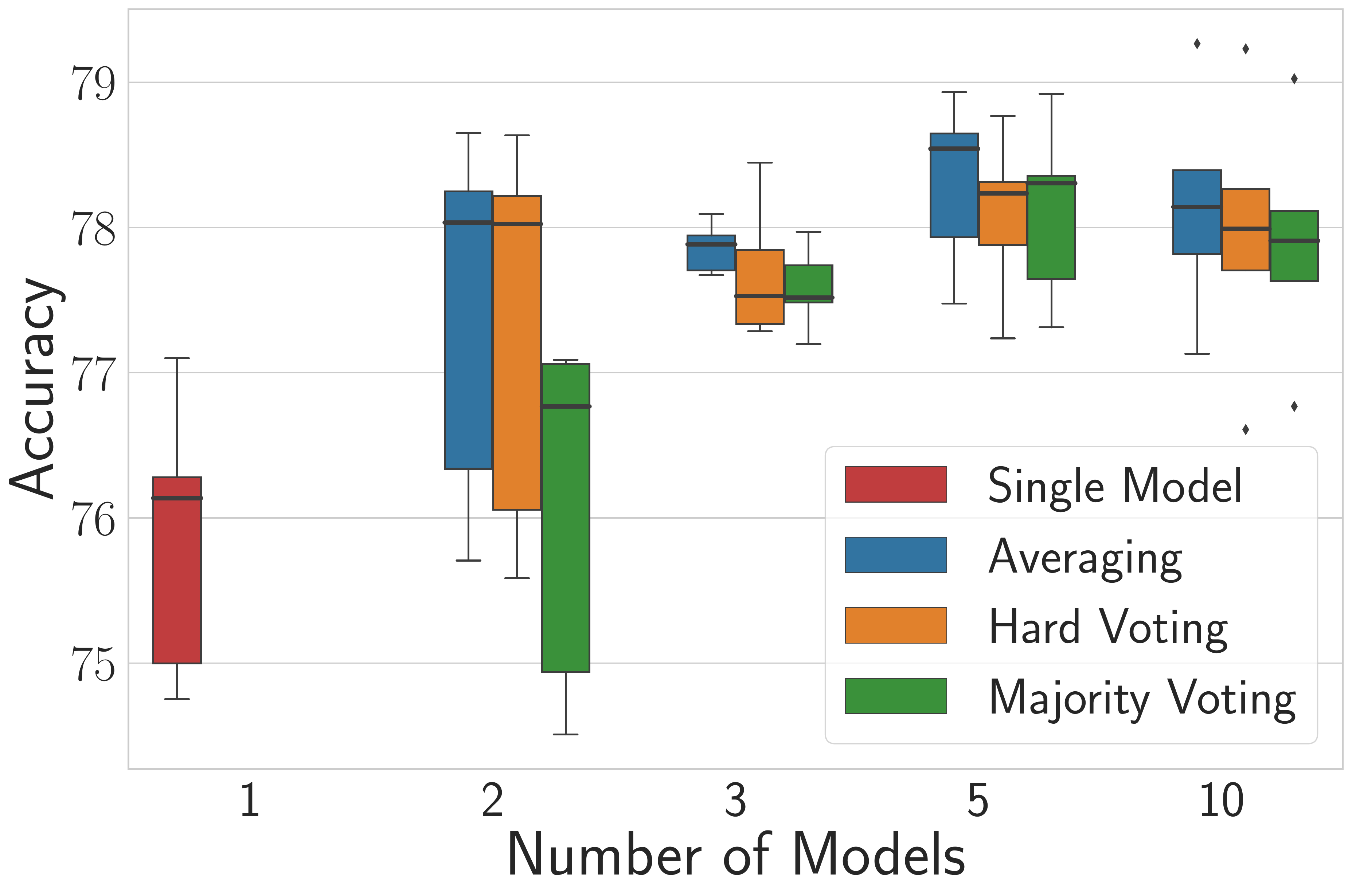}
    \caption{Rotated MNIST}
    
\end{subfigure}\hfill
\begin{subfigure}{.49\textwidth}
    \centering
    \includegraphics[width=0.9\linewidth,height=1.0\linewidth,keepaspectratio=True]{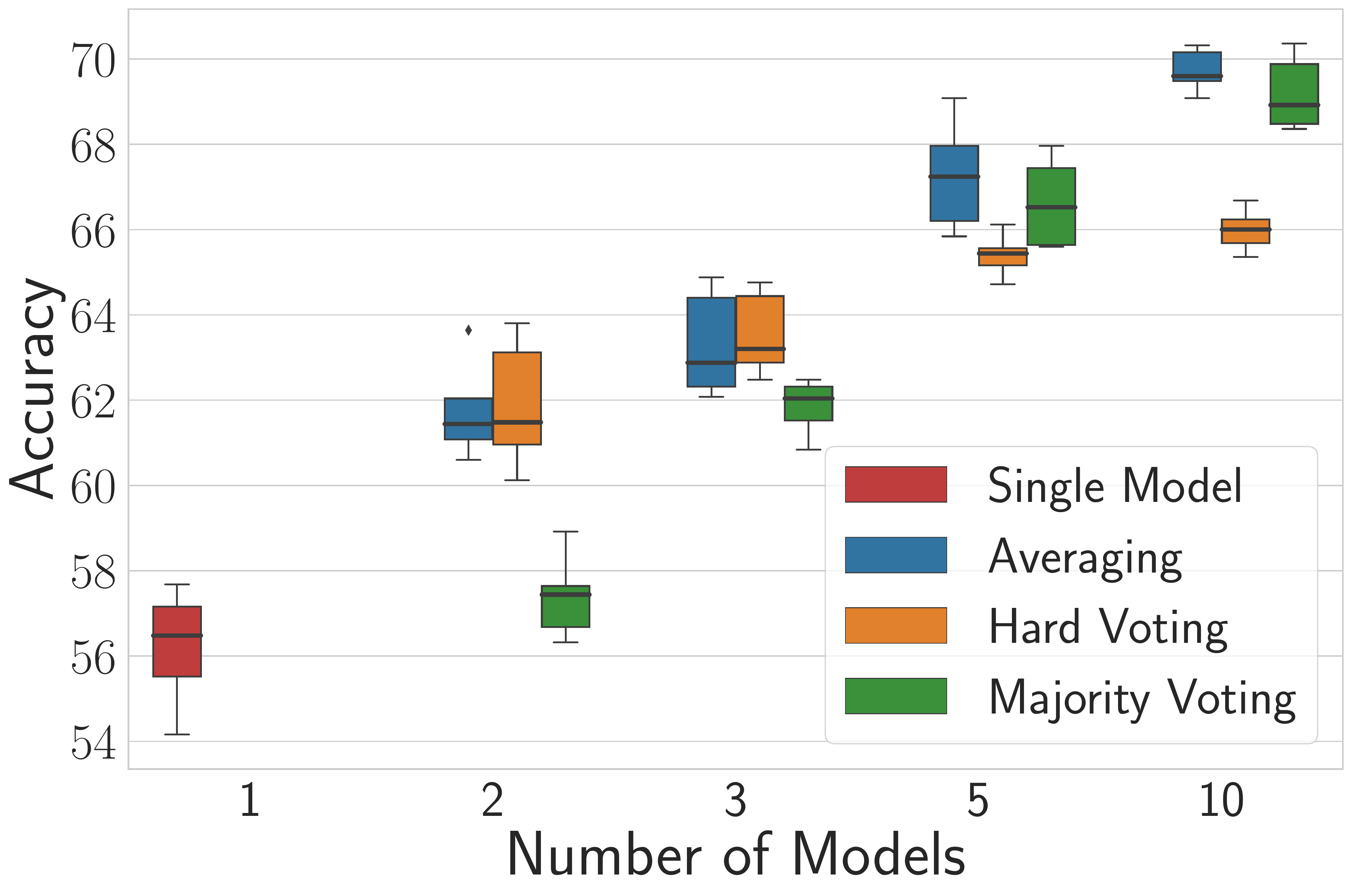}
    \caption{Split CIFAR-100}
   
\end{subfigure}
\caption{Performance of different prediction strategies for Vanilla Ensemble over $5$ seeds for $n=1,2,3,5$ and $10$ models ($x$ axis). Averaging strategy often outperforms the other strategies.}
\label{fig:prediction_strategy}
\end{figure*}

\section{Continual Learning With Ensembles}
\label{sec:2-beyond-one-model}

This section studies the implications of having multiple models for continual learning. We start by comparing basic ensembling strategies and then we study more efficient ensembling techniques. For brevity, we postpone detailed discussion and additional results to Appendix~\ref{sec:ensemble_method_for_cl} and ~\ref{sec:ablation_details}.

\textbf{Algorithms.}
For this analysis, we compare Vanilla CL (Alg.~\ref{alg:single_cl}), Ensemble CL (Alg.~\ref{alg:enemble_cl}), Batch Ensemble CL and Subspace CL (Alg.~\ref{alg:subspace_cl}).

%% file: 2-beyond-single-model.tex
\subsection{Does the Ensembling Strategy Matter?}
\label{sec:prediction_strategies}

Ensembles can use various strategies for learning and prediction. Here, we study three common choices:
\begin{itemize}
    \item \textbf{Averaging}: Average of predictions for each ensemble member.
    \item \textbf{Hard Voting}: Selecting the label for which a member is the most confident.
    \item \textbf{Majority Voting}: Label that gathers the most vote among the members.
\end{itemize}

As shown in Fig.~\ref{fig:prediction_strategy}, averaging strategy consistently outperforms (slightly) other strategies on both benchmarks. In non continual learning scenarios this is consistent with previous findings on the benefits of averaging mechanism~\citep{caccia2021anytime,huang2017snapshot,caruana2004ensemble}. In the context of continual learning, we hypothesize that averaging considers each member's diversity, which is a good fit for multiple tasks and distribution and can boost performance in continual learning. Figure~\ref{fig:prediction_evolution}  provides a visualization of Vanilla ensemble's prediction evolution throughout the learning highlighting the diversity in solution.

\subsection{Different Ensembling Methods}

Although Vanilla Ensemble (VE)\footnote{For the rest of the paper, unless otherwise stated, VE employs the averaging strategy.} shows promising performance benefits, its computation cost grows linearly with the ensemble size. Both Batch Ensemble (BE~\citep{Wen2020BatchEnsemble}) and Subspace Ensemble (SE~\citep{learning-neural-network-subspaces}) provide more efficient ensembling techniques. Hence, for the rest of this section, we study all three methods from performance and computation perspectives.

\begin{figure*}[h!]
\begin{subfigure}{.32\textwidth}
  \centering
  \includegraphics[width=00.99\textwidth,height=1.0\textwidth,keepaspectratio=True]{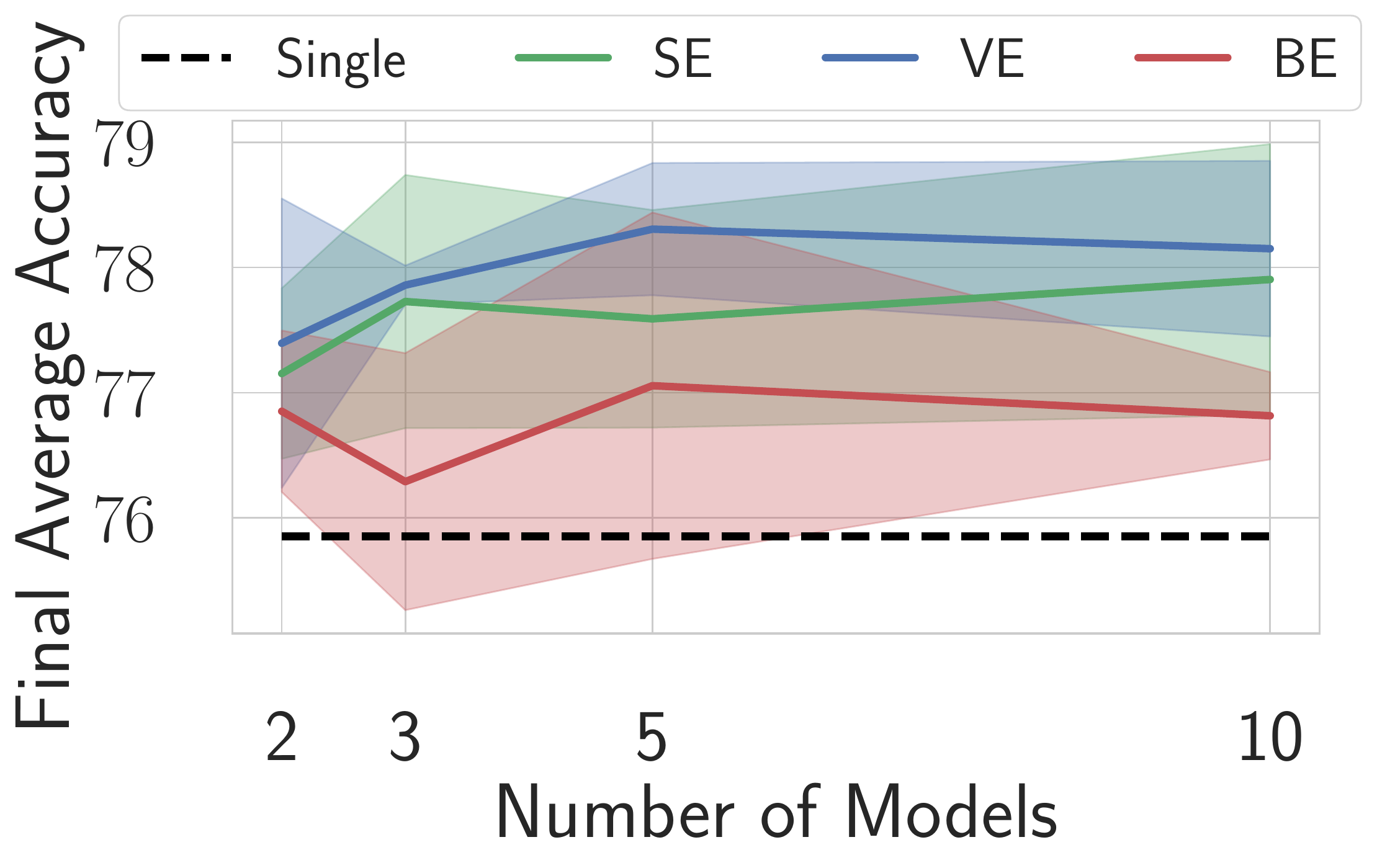}  
  \caption{Final average accuracy}
  \label{fig:intro_accs_rotated}
\end{subfigure}
\begin{subfigure}{.32\textwidth}
  \centering
  \includegraphics[width=00.99\textwidth,height=1.0\textwidth,keepaspectratio=True]{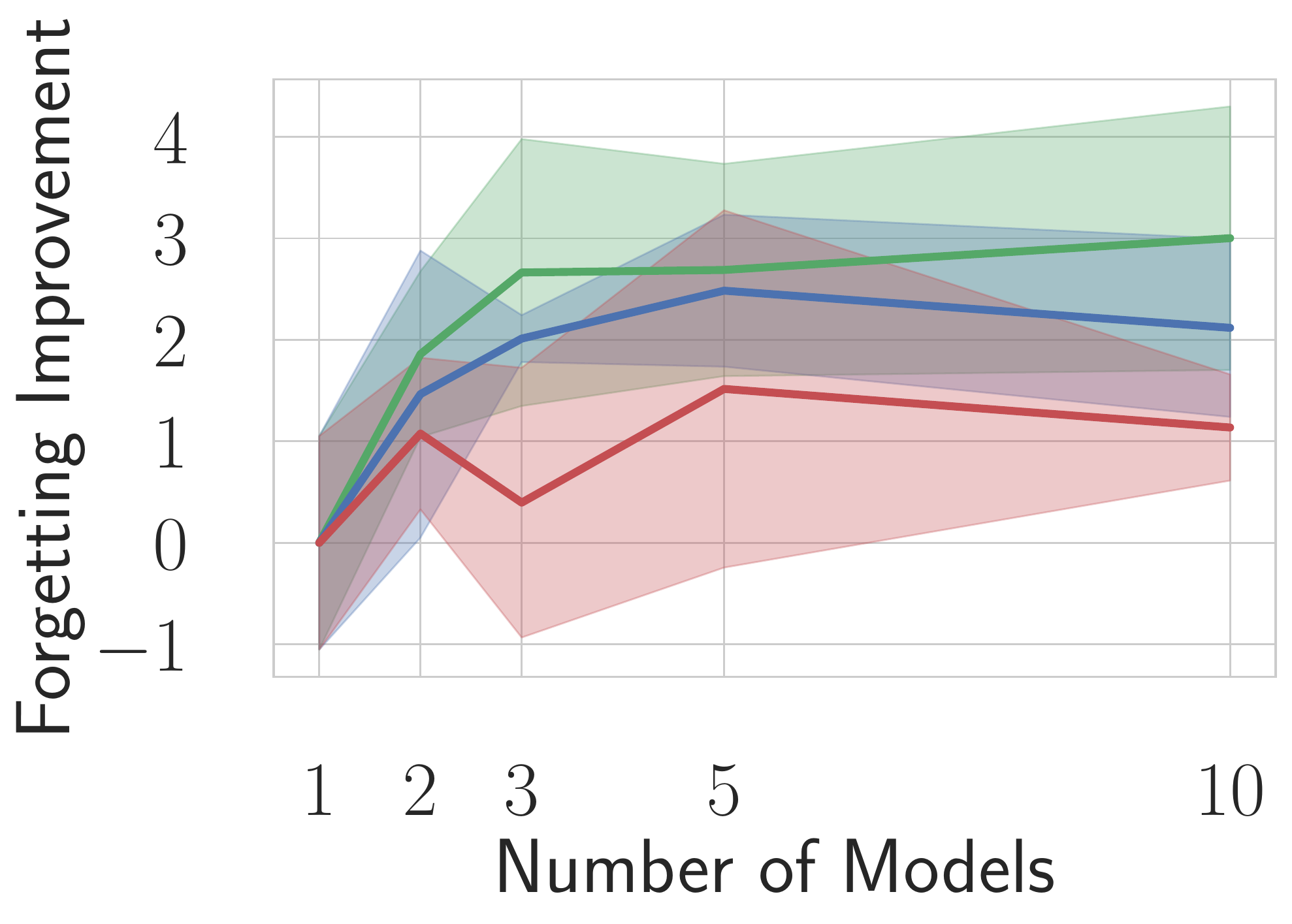}  
  \caption{Forgetting improvement}
  \label{fig:intro_forgetting_rotated}
\end{subfigure}
\begin{subfigure}{.32\textwidth}
  \centering
 
  \includegraphics[width=00.99\textwidth,height=1.0\textwidth,keepaspectratio=True]{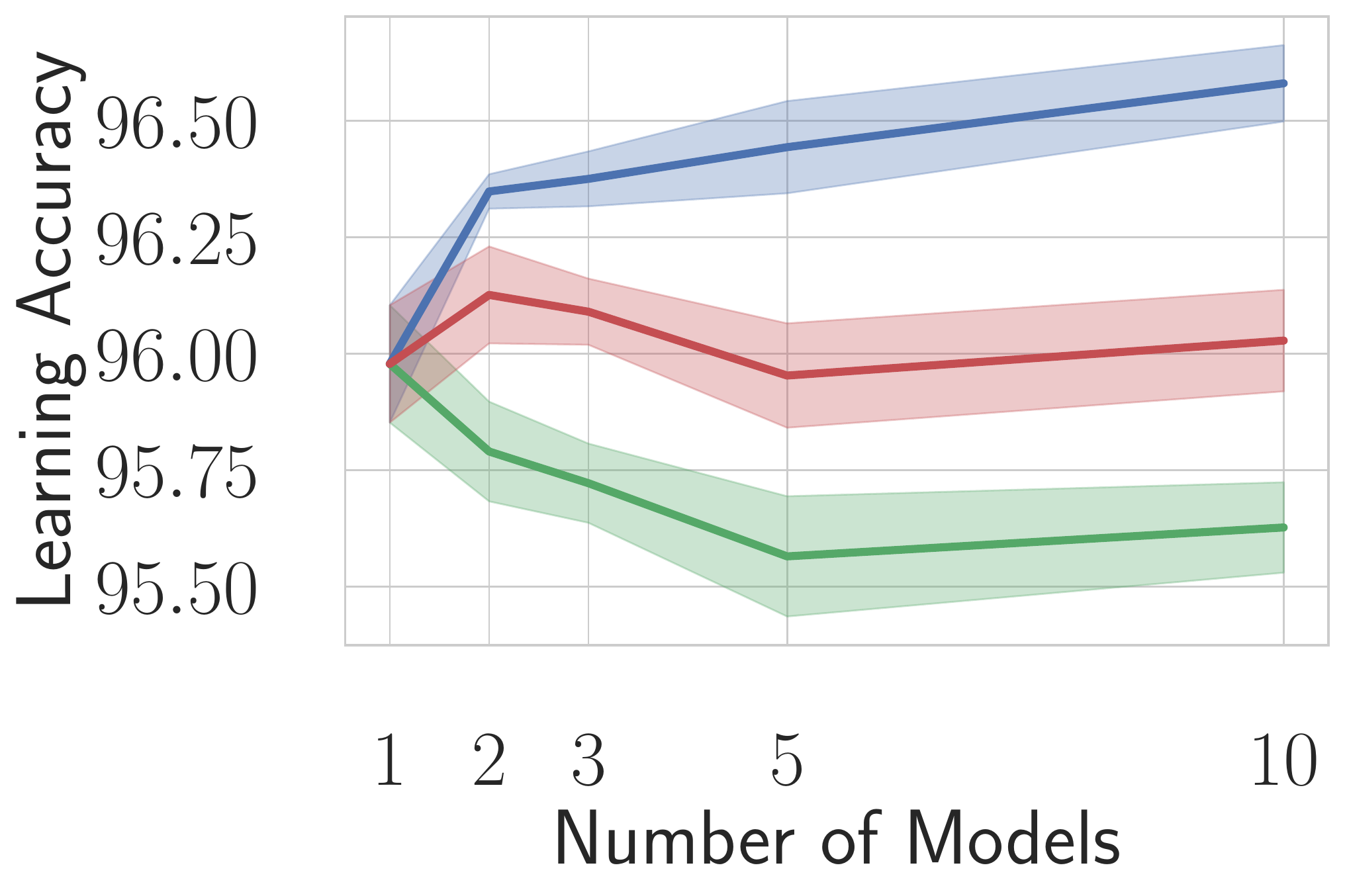}  
  \caption{Learning accuracy}
  \label{fig:intro_learning_acc_rotated}
\end{subfigure}

\begin{subfigure}{.32\textwidth}
  \centering
  \includegraphics[width=00.99\textwidth,height=1.0\textwidth,keepaspectratio=True]{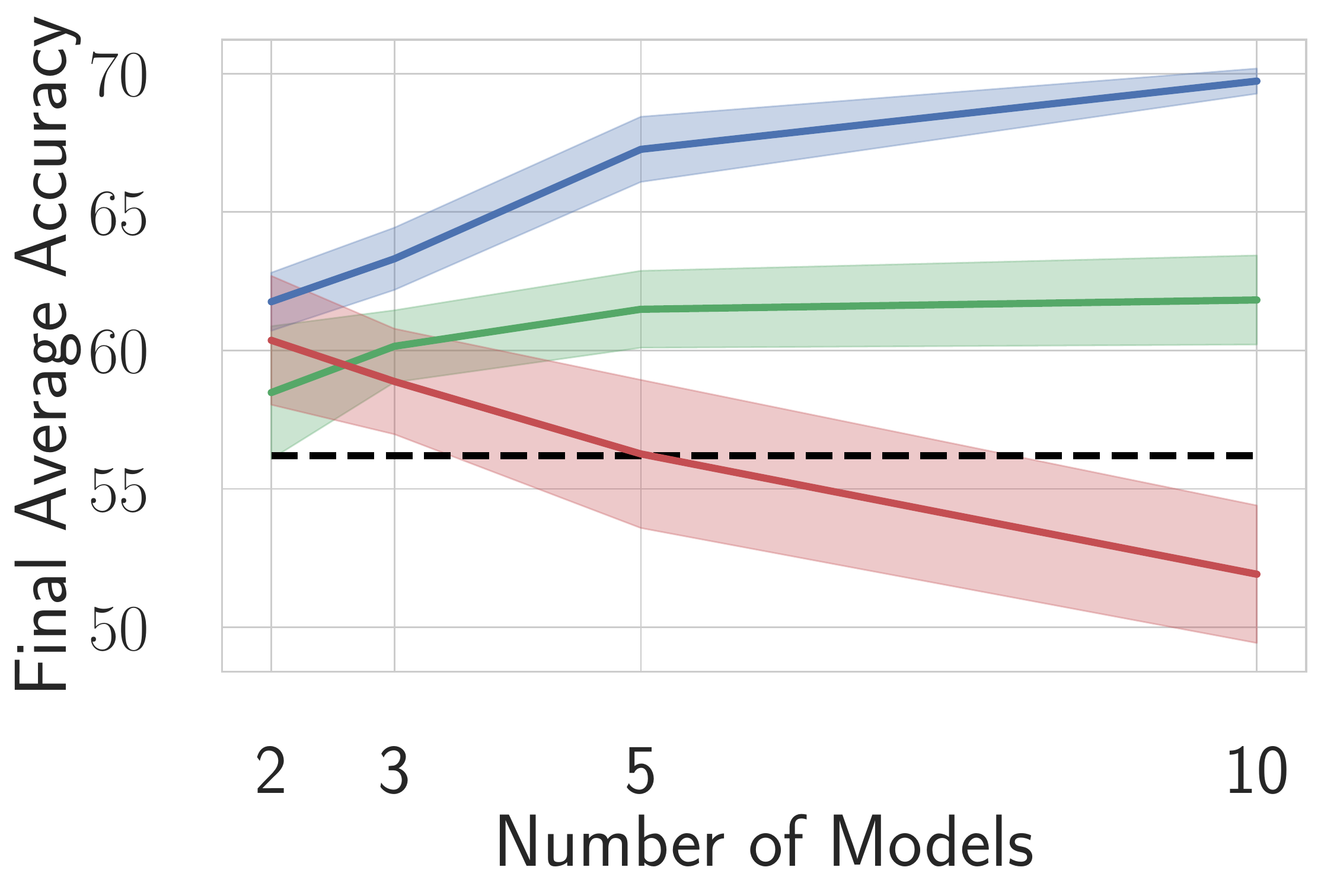}  
  \caption{Final average accuracy}
  \label{fig:intro_accs_cifar}
\end{subfigure}
\begin{subfigure}{.32\textwidth}
  \centering
  \includegraphics[width=00.99\textwidth,height=1.0\textwidth,keepaspectratio=True]{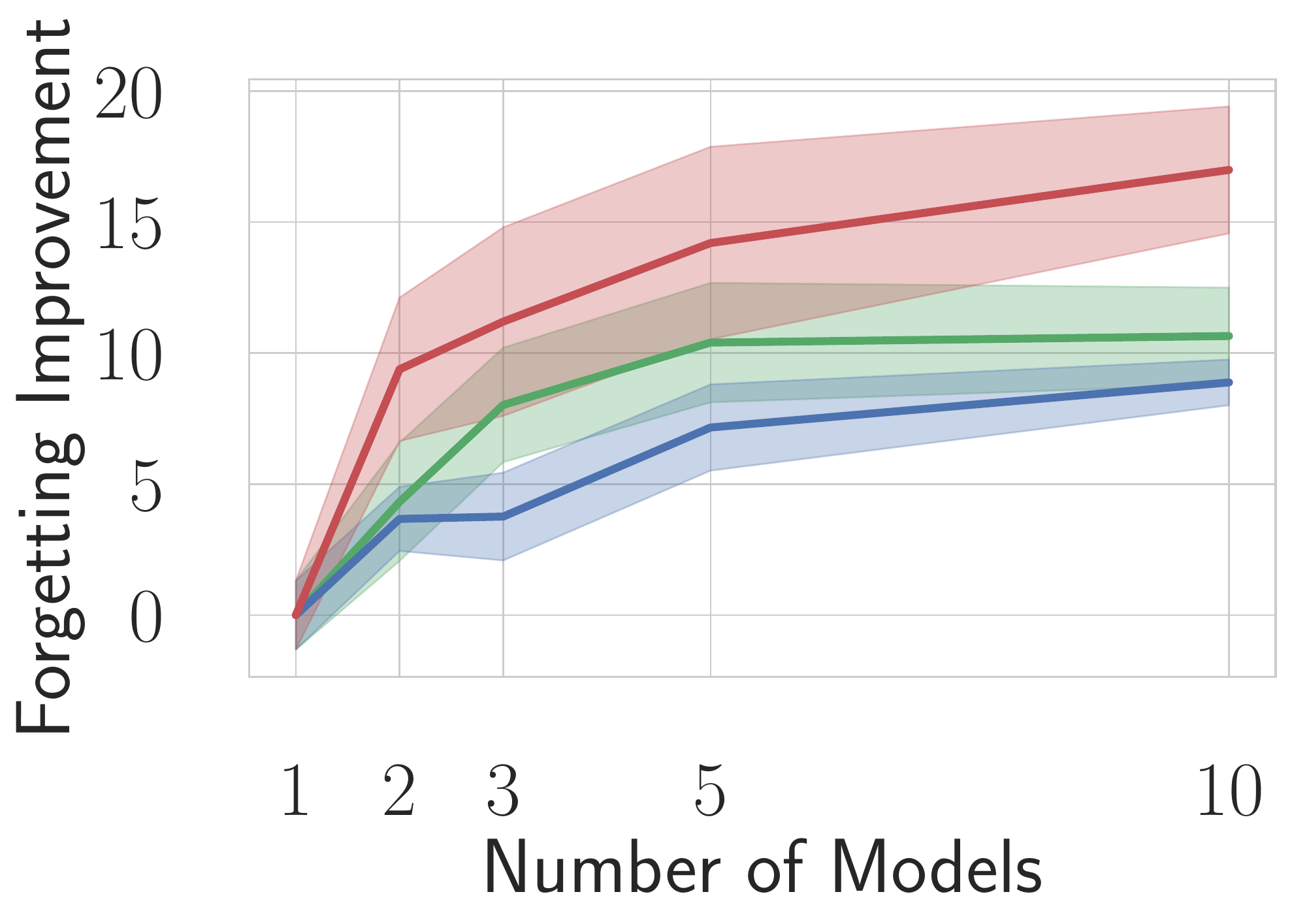}  
  \caption{Forgetting improvement}
   \label{fig:intro_forgetting_cifar}
\end{subfigure}
\begin{subfigure}{.32\textwidth}
  \centering
 
  \includegraphics[width=00.99\textwidth,height=1.0\textwidth,keepaspectratio=True]{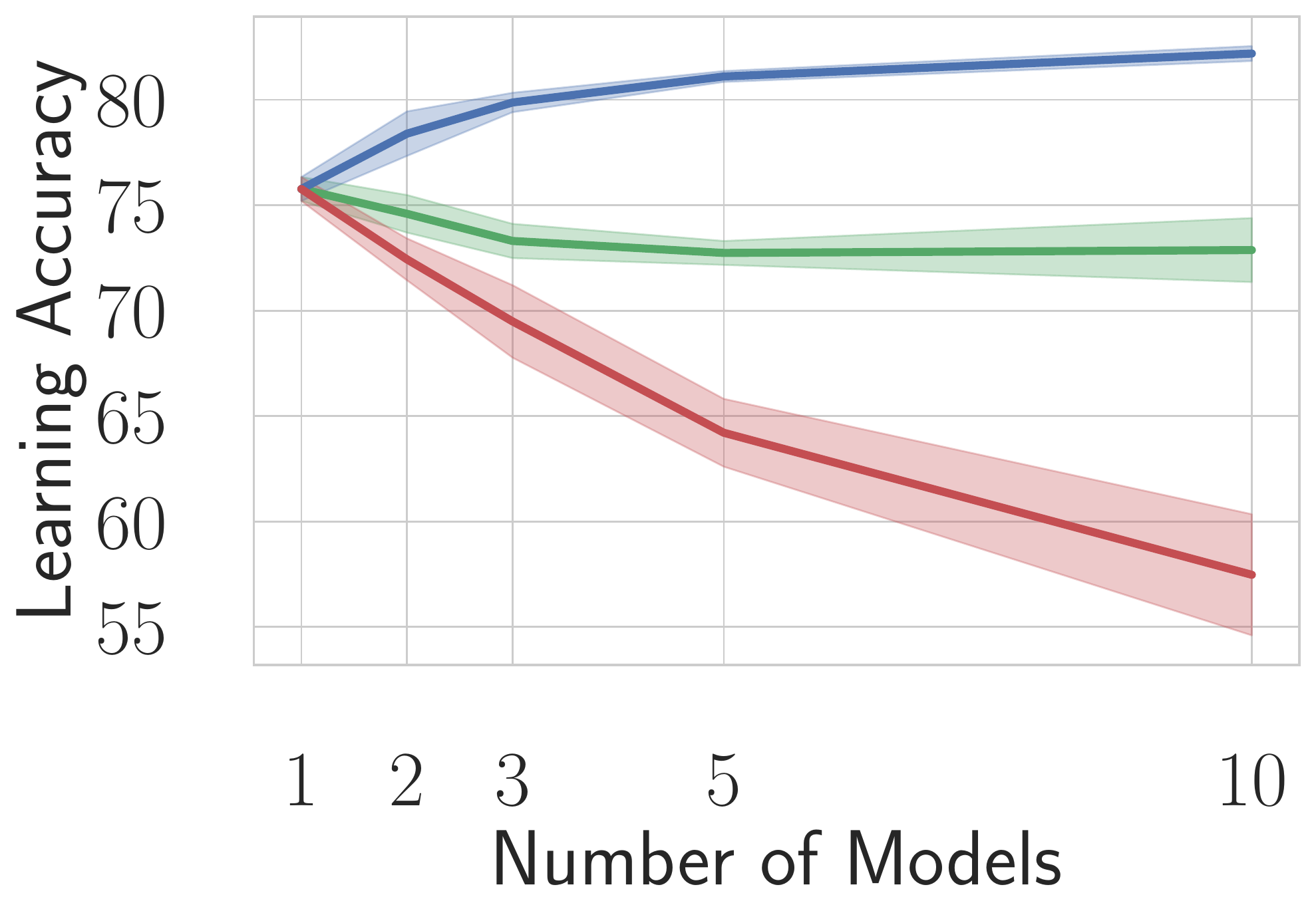}  
  \caption{Learning accuracy}
  \label{fig:intro_learning_acc_cifar}
\end{subfigure}
\caption{\looseness=-2 Continual Learning metrics for Rotated MNIST (first row) and Split CIFAR-100 (second row). Having more models improve the final average accuracy over a single model (\textbf{left}). While SE and BE methods mitigate forgetting (stability) by tying the weights of its member (\textbf{middle}), the VE method enjoys better learning accuracy (plasticity) as each model moves independently (\textbf{right}).}
\end{figure*}

\paragraph{Vanilla Ensemble (VE)} Given a set of weights $\{ \omega_i \}_{i=1}^{n}$, we train independently $n$ models by optimizing $\frac{1}{n}\sum_{i=1}^{n} \mathcal{L}_{\tau}(f_{\omega_i}(x),y)$. Therefore, the models' training differs only in the weights initialization of each model. For the prediction we use the average of each member $\frac{1}{n}\sum_{i=1}^{n}f_{\omega_i}(x)$.

\paragraph{Batch Ensemble (BE)} \looseness=-1 Let's consider a neural network layer weights $W \in \mathbb{R}^{k \times l}$ where $k$ and $l$ are the input and output dimension. BE factorizes the weights $\omega$ such that each member of the ensemble $i$ has weights $\omega_i = \omega \circ f_i$ with $f_i=r_i s_i^{T}$, $r_i \in \mathbb{R}^{k}, s_i \in \mathbb{R}^{l}$. In other words, while they share a common weights $\omega$ they have their own tuple $\{r_i,s_i\}$. During training, each element of the incoming batch $(x,y)$ is shared uniformly among the member of the ensemble. For the prediction we use the ensemble's average prediction as suggested in~\citep{Wen2020BatchEnsemble}.

\paragraph{Subspace Ensemble (SE)}   Given a set of weights $\{ \omega_i \}_{i=1}^{n}$, SE trains a predictor $f_{\hat{\omega}}$ such that $\hat{\omega}=\sum_{i=1}^{n}\alpha_i \omega_i$ with $\sum_{i=1}^{n}\alpha_i=1$, i.e forming a convex combination of the weight of each member. For the prediction we use the midpoint defined as $\frac{1}{N}\sum \omega_i$.

We refer the reader to Appendix~\ref{sec:ensemble_method_for_cl} for more details about each of the ensemble methods as well as their pseudo-code in Appendix~\ref{sec:pseudo_codes}.



\begin{figure*}[h!]
\centering
\begin{subfigure}{.48\textwidth}
    \centering
    \includegraphics[width=1.0\textwidth,height=0.95\textwidth,keepaspectratio=True]{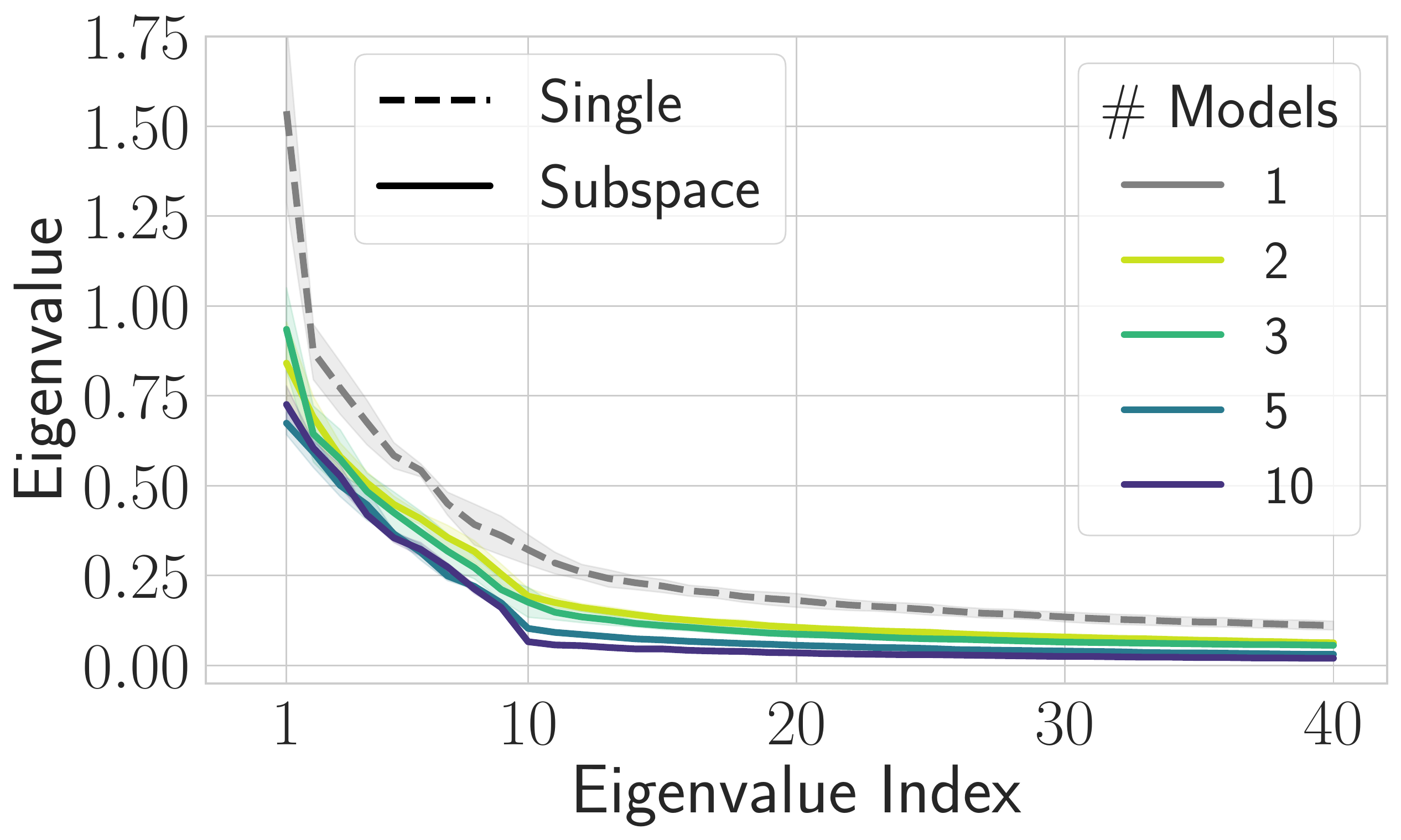}
    
\end{subfigure}\hfill
\begin{subfigure}{.48\textwidth}
    \centering
    \includegraphics[width=0.9\linewidth,height=0.7\linewidth,keepaspectratio=True]{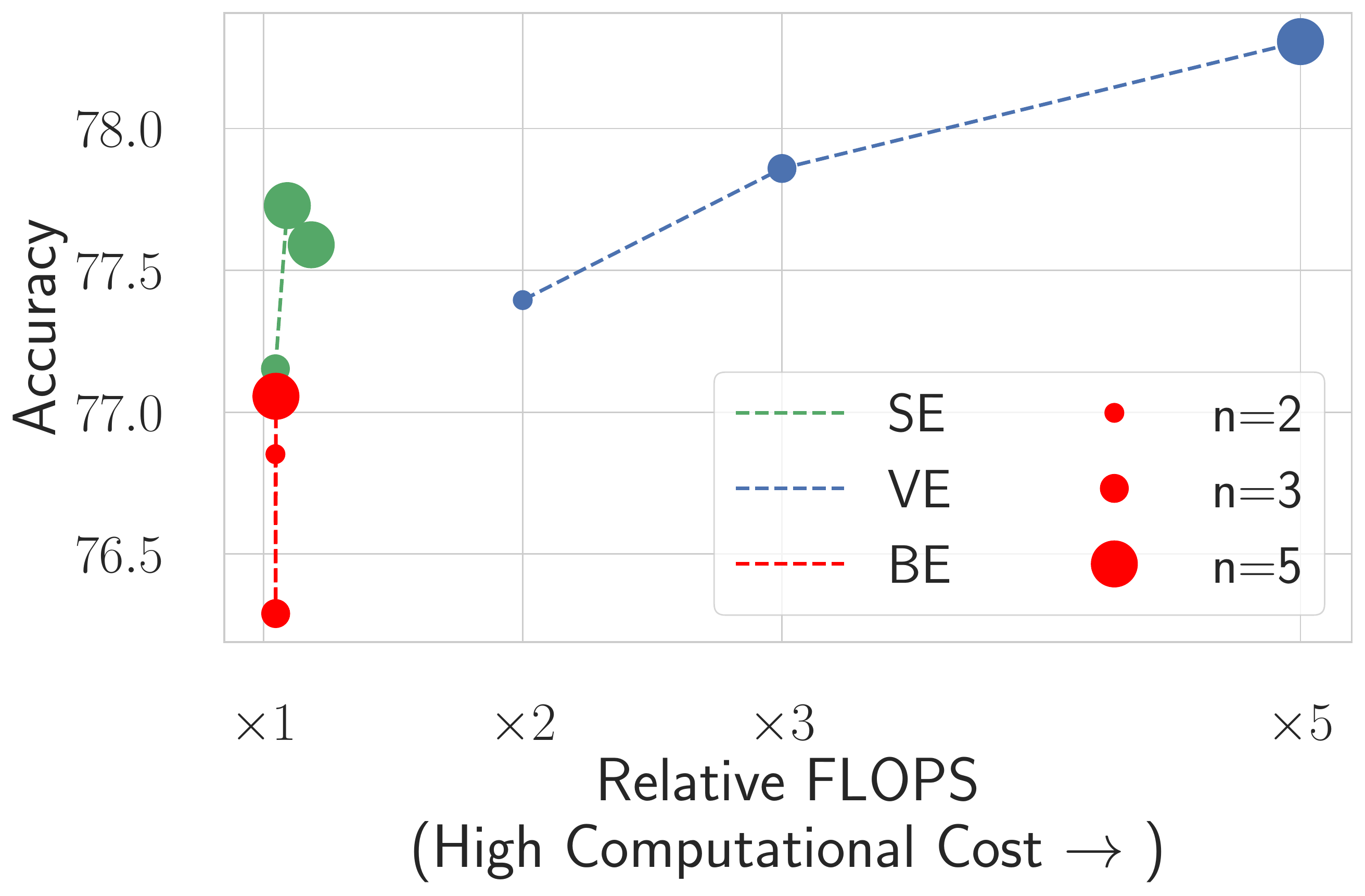}

   
\end{subfigure}
\caption{\textbf{(left)} The Hessian spectrum of each model for the loss of task $1$ for Rotated MNIST. For SE methods, the more models, the lower the eigenvalues. Lower eigenvalues imply flatter minima which is a proxy to describe how much forgetting will be incurred when learning subsequent tasks. \textbf{(right)} Accuracy versus cost in $x$ axis (wrt to single model) for each Ensemble method for Rotated MNIST on $5$ tasks. Although VE shows better performance, it also implies linear growth in cost, unlike SE and BE, which have close to single model compute cost.}
\label{fig:intro_eigen}
\end{figure*}

\subsubsection{Performance Comparison}
We first focus on the continual learning performance of each of the aforementioned methods. More specifically, we focus on the average accuracy, forgetting improvement, and learning accuracy that provide interesting insights on the learning and retention capabilities of each ensembling method.

\textbf{Average Accuracy.} Fig.~\ref{fig:intro_accs_rotated} and \ref{fig:intro_accs_cifar} show the final average accuracy on both Rotated MNIST and Split CIFAR-100 benchmarks for all three methods. We can see the immediate benefits of both ensemble methods compared to the single model (black dashed line). 

\textbf{Average Forgetting.} To better understand these results, we show the \textit{forgetting improvement} (FI) which describes how much a model forgets compared to a single model. Overall, all ensembling methods provide a better forgetting improvement compared to a single model (Fig.~\ref{fig:intro_forgetting_rotated} and \ref{fig:intro_forgetting_cifar}).

\textbf{Learning Accuracy.}
Figs.~\ref{fig:intro_learning_acc_rotated} and~\ref{fig:intro_learning_acc_cifar} show the learning accuracy for each method which demonstrate their ability to learn new tasks. Not surprisingly, ensemble methods with the best forgetting improvement (SE and BE) trade their stability for less plasticity since the models in the ensemble are tightly tied together. On the other hand, since VE trains each of its members independently, it enjoys a better learning accuracy and a better final accuracy. 

SE mitigates CF by tying the weight of each member (through the weight's convex combination). This is illustrated by the eigenvalues of the Hessian of the loss function (Fig.~\ref{fig:intro_eigen}). On the other hand, VE trains its member independently. We believe the power of ensembles and their diversity-enhancing approach (due to random initialization) is very pronounced in the context of continual learning. For BE, we adapted the algorithm from~\cite{Wen2020BatchEnsemble} where originally one member is added for each new task (with a total of $250$ epochs) to our minimal setting of $1$ epoch per task where members are responsible for all tasks similarly to VE. Since the batch is divided equally among all members, this looks similar to bagging strategy and finally, each member only gets to see $1/n$ of the whole dataset.

Given the good performance of ensembles, one may keep adding more models for continued improvement. However, it is either not always helpful (MNIST dataset) or the marginal improvement is not worth the linear compute cost. See Figs~\ref{fig:intro_accs_cifar} and~\ref{fig:intro_forgetting_cifar} for the relevant results. For the subspace method, increasing the number of models (hence the number of parameters) requires more training epochs to update them. That can explain why after a certain threshold (e.g., $n=5$ models for Split CIFAR-100), the accuracy and forgetting metrics degrade or plateau given a fixed training budget. We now elaborate on the compute cost implied by these additions of models.

\subsection{Performance-Cost Trade-off}

\looseness=-1 Another critical factor for comparing ensembles is their compute cost. For convenience, we will refer to the (Inference) FLOPS as the compute cost (The backward pass can be approximated as two times the forward pass). The relative FLOPS is the compute cost ratio wrt to the single model. Table~\ref{tab:computational_cost} provides the compute cost implication of each ensemble method for a single layer as a proof of concept. VE requires the most compute cost with linear growth while SE and BE have an addition and Hadamard product as overhead costs. Fig.~\ref{fig:intro_eigen} (right) shows the accuracy-cost trade-off for VE, SE, and BE. We can observe that VE is the least cost-efficient method while SE and BE have roughly similar costs to a single model. However, SE enjoys higher accuracy compared to BE.

\begin{table}[h!]
\centering

\resizebox{0.5\textwidth}{!}{%
\begin{tabular}{l|c|c}
\hline
\textbf{Method}     & \textbf{Inference Step} & \textbf{Overhead Cost}  \\  \cline{1-3} 
Single Model  & $x \cdot \omega $ &  N/A  \\ \hline
Vanilla Ens.  & $x \cdot \omega_i , \forall i=1..n$ & $(n-1)$ forward pass  \\ \hline
  Subspace Ens.  &  $x \cdot (\sum_{i=1}^{n}\alpha_i \omega_i)$ & 
   $(n-1)$ additions   \\ \hline
Batch Ens.        &  $((x \circ R ) \cdot W )\circ S$ &  $2$ Hadamard products   \\  
\end{tabular}%

}
\caption{ Computational cost comparison between different methods for a given layer $\omega$ and input $x$. SE only has addition operations as an overhead cost, while BE has Hadamard products independently of the size of the ensemble.The first line corresponds to the single model as a reference.}
\label{tab:computational_cost}
\end{table}

Overall, each of the above ensembling methods has its own mechanism to mitigate CF (either with diversity enhanced, tying its member's weights together). However, VE is not computationally efficient, and BE does not provide strong performance. For these reasons, we focus in the next section on improving Subspace Ensemble to provide an efficient alternative approach to the high compute cost of Vanilla Ensemble.

%% file: 4-methodology.tex
\section{Improving Subspace Learning with Connectivity}
\label{sec:4-methods}
In Sec.\ref{sec:2-beyond-one-model}, we observed that although SE enjoys a nice compute cost, it cannot match the performance of the VE method. In this section, we aim to find out if we can improve the performance of the SE method in continual learning scenarios while keeping the cost roughly the same.

To improve the performance, Experience Replay (ER)~\cite{riemer2018learning} is one of the most popular and practical CL methods that come to mind at first. However, naively adding ER method to subspace will not increase the performance significantly. To explain, we note that the subspace formed by models of the SE method is subject to drift as the models' parameters still change as the task changes. As a result, the optimal subspace found by the SE method will drift as the number of tasks increases. Appendix~\ref{sec:subspace_properties} provides a more detailed analysis of this subject.

To prevent the subspace drift problem and thus exploit the benefits of the ER method, we note that SE method~\cite{learning-neural-network-subspaces} is originally motivated by mode connectivity of optima~\cite{modeconnectivity_main}. In the context of continual learning, \cite{mirzadeh2021MCSGD} has shown that enforcing the linear mode connectivity of each task's optima is equivalent to mimicking the multitask (i.e., joint) training and hence, a key factor in preventing forgetting. Given the shared origins between~\cite{learning-neural-network-subspaces} and~\cite{mirzadeh2021MCSGD}, it is natural to think about both works together to overcome the drift challenge. However, \cite{mirzadeh2021MCSGD} studies continual learning with a ``single'' model, while here, we work with ``multiple'' models that form a convex region. Intuitively, our algorithm is the generalization of the proposed algorithm by~\cite{mirzadeh2021MCSGD} where we prevent the drift of an optimal subspace rather than a single optimal (i.e., model). 

Our \algoname algorithm proceeds in two steps: we first naively finetune to the incoming task to learn a solution $\omega^{*}_{\tau}$ before creating a low-loss path to the former task's solutions $\omega^{*}_{\tau-1}$ that enforces the connectivity between subspaces found for each task. A pseudo-code can be found in Alg.~\ref{alg:subspace_connectivity}.

\textbf{(1) Learning a subspace solution for the incoming task}: The first step consists in learning a subspace solution by fine-tuning on task $\tau$ leading to the solution $\hat{W}_{\tau}=\{ \hat{w}_{\tau,i}\}_{i=1}^{n}$  obtained by optimizing:
\begin{equation}
\{ \hat{\omega}_{\tau,i} \}_{i=1}^{n}=\underset{ W }  {\mathrm{argmin}}  \quad \mathbb{E}_{\pmb{\alpha} \sim \mathcal{U}[\Delta^{n}]}[ \mathcal{L}_{\tau}(W^{T}\pmb{\alpha})]
\label{eq:w_hat}
\end{equation}
At the end of this step, we save in a buffer memory $\mathcal{B}$, $m_{\mathcal{B}}$ samples per class per task that will be used to connect linearly two subspace's solutions.

\textbf{(2) Connecting the new subspace to previous subspaces}: This step aims at connecting subspaces from prior task's solutions together as in MC-SGD \cite{mirzadeh2021MCSGD}. First, we use the midpoint of subspace $\tau$ denoted $\hat{\omega}^{*}_{\tau,mid}$ as its proxy since it gives the best performance (Sec.~\ref{sec:inside-subspace}). We then connect $\hat{\omega}^{*}_{\tau,mid}$ and previous tasks midpoint solution $\omega^{*}_{\tau-1,mid}$ via a low loss path. The loss over the connecting path acts as a penalty or regularizer term. The rationale behind choosing the middle point as a proxy for subspace is that the middle point is the most stable point of the subspace~\citep{learning-neural-network-subspaces}. In our experiments in the appendix, we also confirm this observation in the context of continual learning.

Thus, for the second step of our algorithm, we optimize the following objective with saved elements from the buffer:
\begin{equation}
 \begin{aligned}
&\{ \omega^{*}_{\tau,i}\}_{i=1}^{n}=\underset{W}{\mathrm{argmin}} \quad \mathbb{E}_{\pmb{\alpha}\sim \mathcal{U}(\Delta^{n+1})} [\displaystyle{\sum_{j=1}^{\tau-1}}\mathcal{L}_{j}(W^{T}\pmb{\alpha}^{n}+ 
&\alpha_{n+1}\omega_{\tau-1,mid}^{*})]+   \mathcal{L_{\tau}}(W^{T}\pmb{\alpha}^{n}+\alpha_{n+1}\hat{\omega}_{\tau,mid})]
\label{eq:w_bar}
\end{aligned}
\end{equation}
where $\pmb{\alpha}=(\underbrace{\alpha_1,..,\alpha_n}_{\pmb{\alpha}^{n} \in \mathbb{R}^{n}},\alpha_{n+1}) \in \mathbb{R}^{n+1}$ and $\mathcal{L}_{j}$ corresponds to the loss of task $j$ (using element of task $j$ from the buffer). The rationale behind this loss function is to create a linear path of low-loss between two subsequent solutions $\omega_{\tau-1,mid}^{*}$ and $\omega_{\tau,mid}^{*}$.

A few remarks worth noting here. Intuitively speaking, the subspace method ties the models for a single task, while the mode connectivity regularization ties the Subspaces together.
Note that the original subspace method is only developed for single-task settings. Although SE is an efficient ensembling technique, it still does not have any mechanism to learn from a sequence of tasks.

\begin{algorithm}[t]
\footnotesize
\begin{spacing}{0.8}
    \SetKwInOut{Input}{Input}
    \SetKwInOut{Output}{Output}
    \SetKw{KwBy}{by}
    \SetAlgoLined
    \Input{A task sequence $\cT_1, \cT_2, \ldots , \cT_{T}$, number of models $n$ , buffer $\cB$ and memory size $m_{\cB}$ }
    
         Initialize set of weigths $S^{n}=\{\omega_{0,i} \}_{i=1}^{n}$, buffer $\mathcal{B} \leftarrow \{  \}$  \\
         \For{tasks  $\tau=1,2,3,\ldots T  $}{
            Get $\{ \hat{\omega}^{*}_{\tau,i} \}_{i=1}^{n}$ with Eq \ref{eq:w_hat} \:   \tcp{Learn subspaces solution for task $\tau$} 
           
                  \For{ $(x, y) \in \mathcal{T}_{\tau}$}
                {   $\mathcal{B} \leftarrow \cB \bigcup \{x,y \}$  \tcp{\textcolor{blue}{Collect 1 sample per class at the end of each class}} \hskip3.5em  }
               
             Get $\{ \omega^{*}_{\tau,i} \}_{i=1}^{n}$ with Eq \ref{eq:w_bar} using samples from $\mathcal{B}$  \:  \tcp{Connect previous solution' subspaces} 
           
        }
    
    \Output{ $\{ \omega^{*}_{T,i} \}_{i=1}^{n}$}
    \caption{\algoname CL}
    \label{alg:subspace_connectivity}
\end{spacing}
\end{algorithm}

%% file: 5-experiments.tex
\section{Experiments and Results}
\label{sec:5-experiments-results}

In this section, before comparing \algoname to Ensemble baselines, we want to address a subsidiary yet important question: ``Will increasing the number of parameters (for instance, the number of hidden units) boost performance, or is it the way parameters work together as an ensemble that matters?''
To address this hypothesis, we compare the Ensemble model with a Scaled version of a single model where we either increase the number of hidden units (for dense networks) or the number of filters (for convolutional networks) to match the number of parameters in Ensemble models. Finally, we compare our algorithm with an Ensemble version of MC-SGD~\citep{mirzadeh2021MCSGD} (dubbed Ens MC-SGD) and provide a discussion between the trade-off accuracy and compute cost. As a note, we use MC-SGD for the experiments in this section since it is one of the strongest CL methods and shares a similar mechanism to our proposed method, so the comparison is fairer. Although the main focus of our work is ensemble methods, we additionally report the results for other ensemble methods (e.g., Batch Ensemble) and also single models with various learning algorithms (e.g., A-GEM, ER, etc.) in Appendix~\ref{sec:final_exp_parameters}. Since we want to simulate an online setting, which is relevant for small AI embedded devices, we ran all algorithms with only 1 epoch per task.

\subsection{Does Increasing the Number of Parameters Help?}
We first compare Ensemble MC-SGD (with $n=3$) to Scaled MC-SGD, where the number of hidden units has been increased to $600$ and the number of kernel filters to $35$, to match the parameters and compute cost. Table~\ref{tab:results_cnn} showcase the performance of both models. We can observe that across all benchmarks, even with the same number of parameters and thus computation, the Ensemble MC-SGD model matches or outperforms the Scaled MC-SGD. Similar to our discussion in Sec.~\ref{sec:introduction}, we can see that the benefit of ensembles is not just the increase of parameters but the way the models communicate with each other, which can lead to more diverse solutions.

\begin{table*}[ht!]
\centering

\resizebox{\textwidth}{!}{%
\begin{tabular}{lccccc}
\hline
\multirow{2}{*}{\textbf{Method}} &
  \multicolumn{2}{c}{\textbf{Permuted MNIST}} &
  \multicolumn{2}{c}{\textbf{Rotated MNIST}}& Relative FLOPS ratio 
    \\ \cline{2-6} 
        & Accuracy $\uparrow$ & Forgetting $\downarrow$& Accuracy   $\uparrow$ & Forgetting  $\downarrow$ & \\ \hline

MC-SGD (Mirzadeh et. al \text{\cite{mirzadeh2021MCSGD}})      & 82.9 ($\pm$0.40) & 0.10 ($\pm$0.01) & 81.9 ($\pm$0.46) & 0.08 ($\pm$0.01) & 1 \\ \hline
Scaled MC-SGD       & 88.03 ($\pm$0.36) & 0.06 ($\pm$0.01) & 83.67 ($\pm$0.40) & 0.07 ($\pm$0.01) & $3.11$ \\
Ensemble MC-SGD              & 88.30 ($\pm$0.48) & 0.06 ($\pm$0.01) & 83.63 ($\pm$0.39) & 0.07 ($\pm$0.01) &  $3$\\ 
\algoname  & \multicolumn{1}{l}{87.8 ($\pm$0.30)} &0.07 ($\pm$0.01) & \multicolumn{1}{l}{86.7 ($\pm$0.67)} & 0.07 ($\pm$0.01)      & $1.03$ \\\hline

 Multitask Learning & 89.5 ($\pm$0.21) & 0.0              & 89.8($\pm$0.37)  & 0.0      &     NA  \\ \hline
\end{tabular}%

}

\end{table*}

\begin{table*}[ht!]
\centering

\resizebox{\textwidth}{!}{%
\begin{tabular}{lccccc}
\hline
\multirow{2}{*}{\textbf{Method}} &
  \multicolumn{2}{c}{\textbf{Split CIFAR-100}} &
  \multicolumn{2}{c}{\textbf{Split miniImageNet}}& Relative FLOPS ratio 
    \\ \cline{2-6} 
        & Accuracy $\uparrow$ & Forgetting $\downarrow$& Accuracy   $\uparrow$ & Forgetting  $\downarrow$ & \\ \hline

MC-SGD (Mirzadeh et. al \text{\cite{mirzadeh2021MCSGD}})      & 58.22 ($\pm$0.91) & 0.08 ($\pm$0.01) & 54.80 ($\pm$1.04) & 0.03 ($\pm$0.01) & 1 \\ \hline
Scaled MC-SGD    & 60.55 ($\pm$0.84) & 0.06 ($\pm$0.01) &55.44 ($\pm$1.36) & 0.05 ($\pm$0.01) &  $3.01$ \\ 
Ensemble MC-SGD              & 64.12 ($\pm$1.16) & 0.06 ($\pm$0.01) & 59.10 ($\pm$1.1) & 0.04 ($\pm$0.01) & $3$ \\
\algoname            (ours) & 61.7 ($\pm$0.80) & 0.05 ($\pm$0.01) & 58.17 ($\pm$0.84) & 0.03 ($\pm$0.01)  & $\approx 1$  \\ \hline
 Multitask Learning & 66.8 ($\pm$1.42) & 0.0              & 62.82($\pm$1.77)  & 0.0      &  NA   \\ \hline
\end{tabular}%

}
\caption{Comparison ensemble methods performance with $n=3$ models or the equivalent number of parameters for Scaled MC-SGD. Each benchmark includes 20 tasks.}
\label{tab:results_cnn}
\end{table*}

\begin{figure*}[h!]
\centering
\begin{subfigure}{.245\textwidth}
    \centering
    \includegraphics[width=1.0\textwidth,keepaspectratio=True]{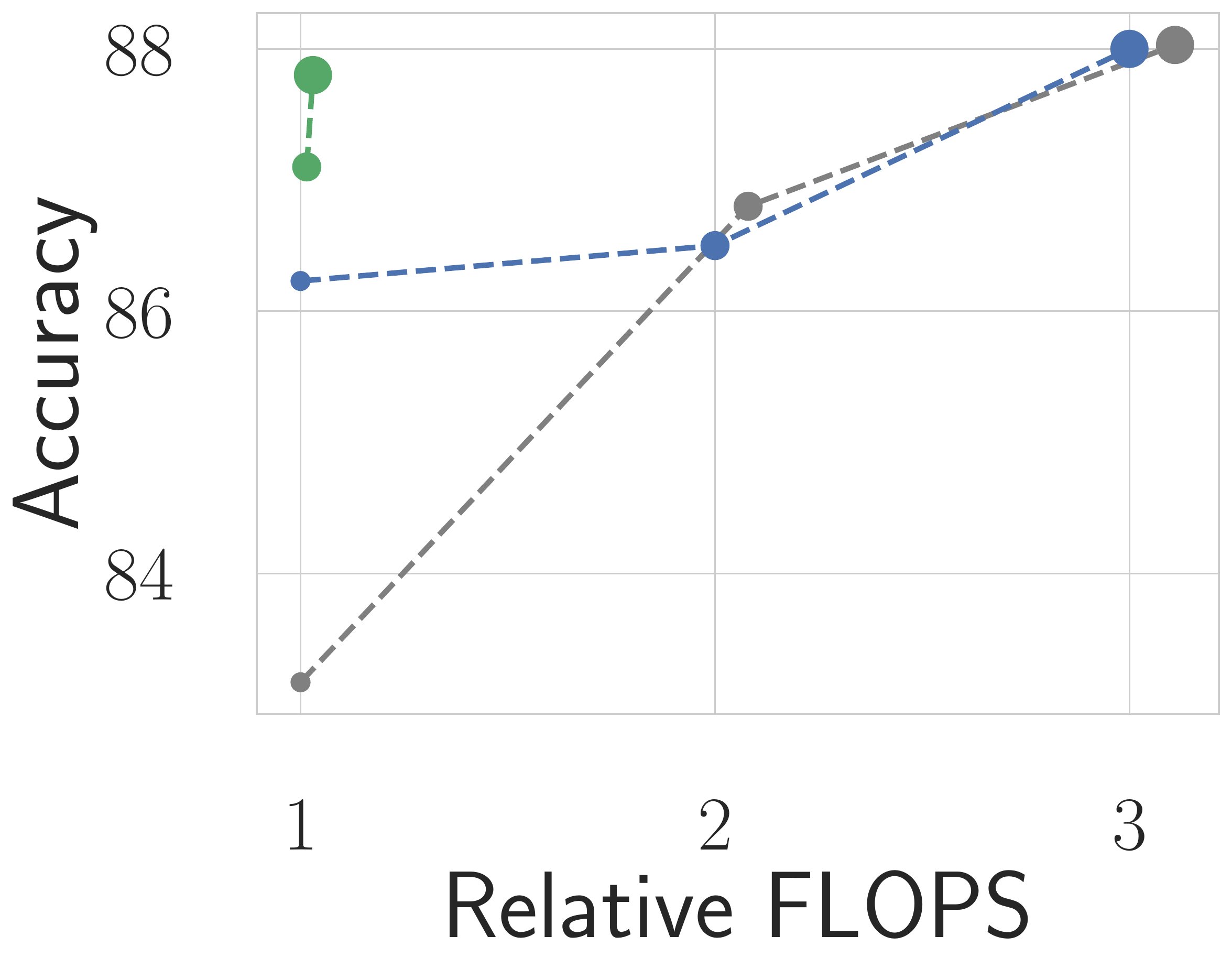}
    \caption{Permuted MNIST}
\end{subfigure}\hfill
\begin{subfigure}{.245\textwidth}
    \centering
    \includegraphics[width=1.0\textwidth,keepaspectratio=True]{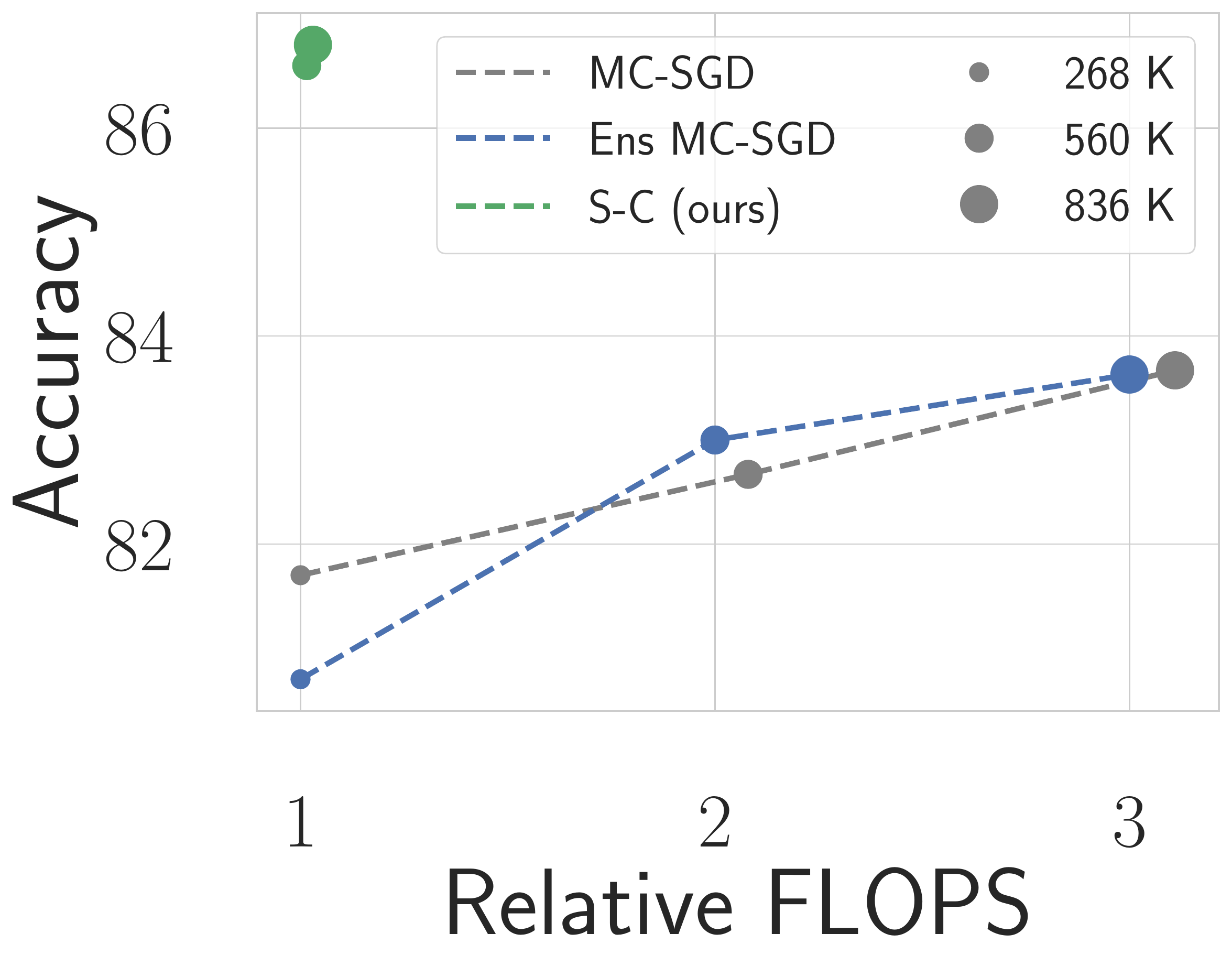}
    \caption{Rotated MNIST}
\end{subfigure}
\begin{subfigure}{.245\textwidth}
    \centering
    \includegraphics[width=1.0\textwidth,keepaspectratio=True]{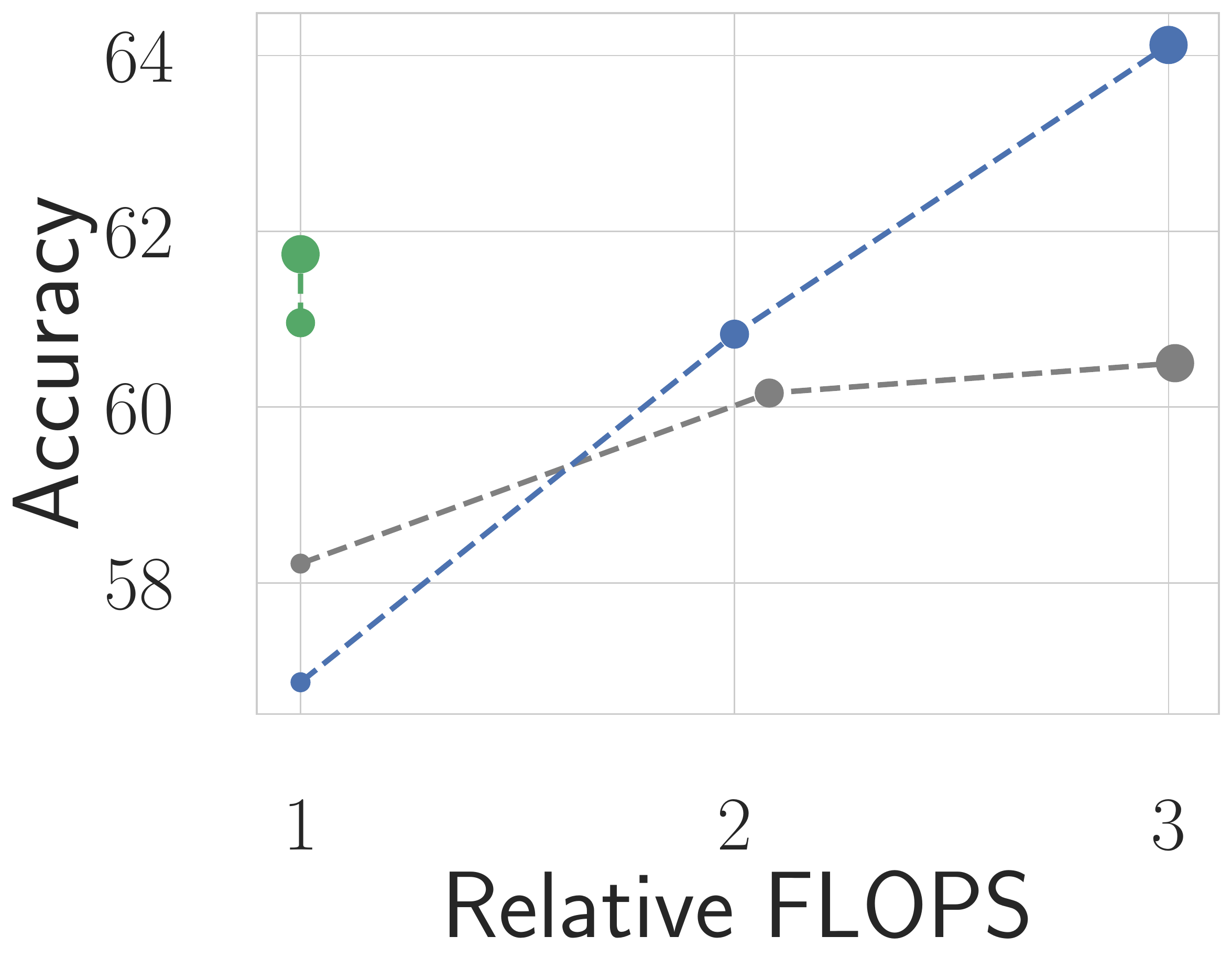}
    \caption{Split CIFAR-100}
\end{subfigure}\hfill
\begin{subfigure}{.245\textwidth}
    \centering
    \includegraphics[width=1.0\textwidth,keepaspectratio=True]{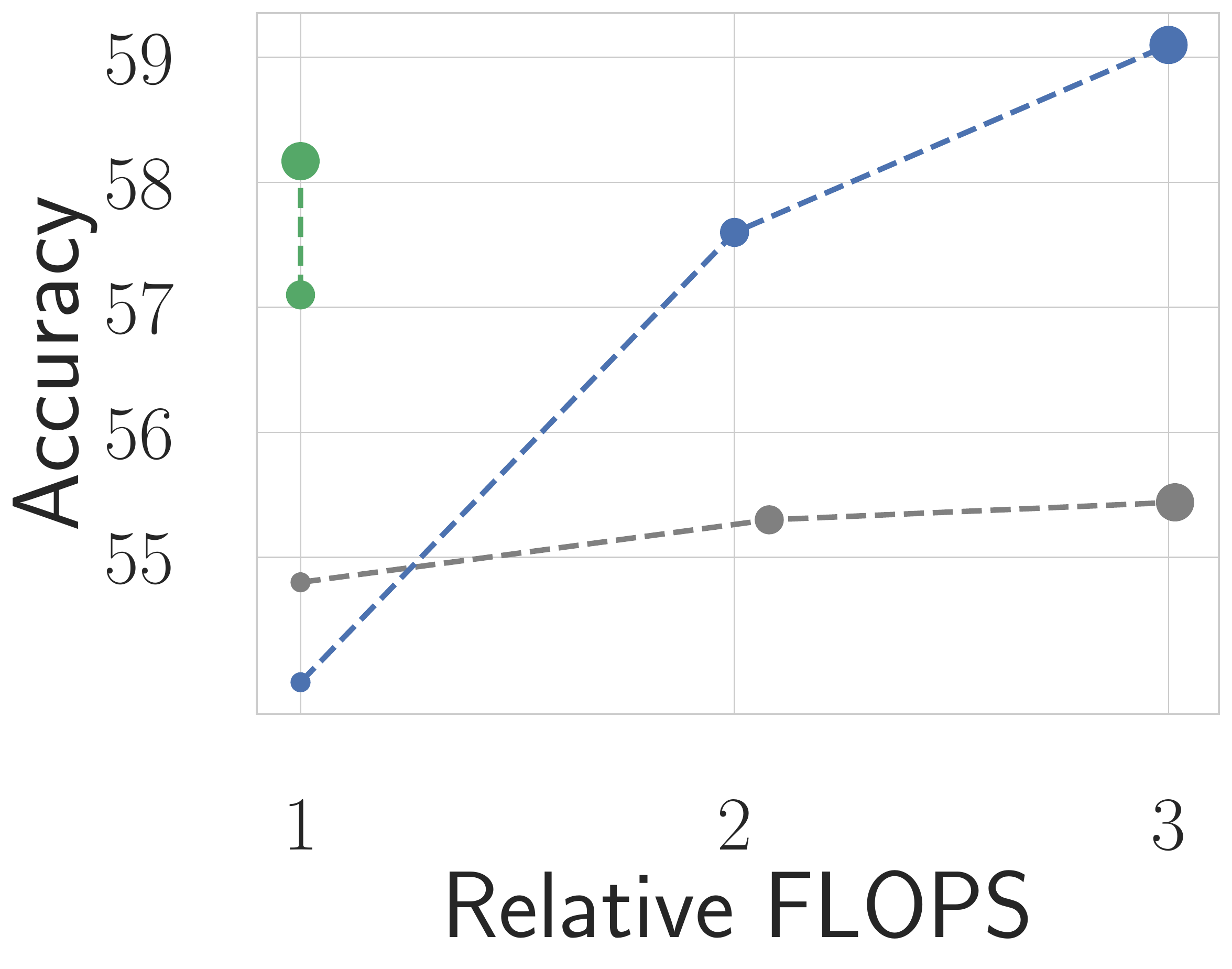}
    \caption{Split miniImageNet}
\end{subfigure}

\caption{Performance of (Scaled) MC-SGD, Ensemble MC-SGD (Ensemble) and \algoname (S-C) with respect to the inference cost (FLOPS) and number of parameters/models $n$ (circles) on 20 tasks.}
\label{fig:trade_off_performance_compute_cnn}
\end{figure*}

\subsection{Performance Vs. Compute Trade-off}
Fig.~\ref{fig:trade_off_performance_compute_cnn} reports the performance versus the relative FLOPS (w.r.t to MC-SGD). We notice that for fully connected models (MNIST) \algoname is at least on par with Ensemble MC-SGD and Scaled MC-SGD but enjoys a much better compute cost (close to Single MC-SGD). However, for convolutional models, Ensemble MC-SGD has the best performance but at a high computational cost. One reason might be due to the small dataset size ($500$ per task) on which Subspace models are not performing well. Note that the subspace method incurs almost no computational cost overhead in this case. This can be explained by the high reuse of kernel filters in convolutional layers (once the weights convex combination are summed). Note that the Ensemble MC-SGD with computing cost corresponding to $1$ on the x-axis corresponds to the bagging strategy where incoming batches are randomly assigned to $1$ one the member of the ensemble (each of them eventually gets to see $1/n$ of the total dataset).

Overall, depending on the limiting factor and use case, one can either use Vanilla Ensemble for its simplicity and performance regardless of the compute cost. However, \algoname has the best cost-performance trade-off among other methods.

%% file: 6-related_works.tex
\negspace{-2mm}
\section{Related Work}
\negspace{-2mm}
The methodology to tackle catastrophic forgetting for continual learning has been extensively populated with three main groups.

\textbf{Continual learning.} Among pioneer methods to alleviate the catastrophic forgetting, we can name \emph{regularization-based} methods that limit the drift in important parameters of features of past tasks~\citep{EWC,zenke2017continual,nguyen2017variational,yin2020optimization}. For instance, EWC~\citep{EWC} uses the Fisher information to identify the important parameters. A major drawback of the regularization methods is that they often need multiple passes over data to perform well~\citep{AGEM} and when the number of tasks is large, they suffer more from the feature drift~\citep{titsias2020functional}.

The second group of methods in continual learning is the \emph{memory-based} methods that keep a small episodic memory of data from past tasks to either replay those examples~\citep{Chaudhry2019OnTE,rebuffi2017icarl,riemer2018learning,lesort2020continual} or use them for improving the optimization procedure such as projection methods~\citep{farajtabar2020orthogonal,bennani2020generalisation,saha2021gradient}, or train a generative model to serve that purpose~\citep{shin2017continual, kirichenko2021task}. A-GEM~\citep{AGEM} is a notable example of these methods that use gradients of past tasks to modify the gradients of the new task and alleviate the forgetting.

Finally, \emph{parameter isolation} methods focus on the neural network modules that can be either be expanded for each new task~\citep{aljundi2017expert} or a sub-network will be allocated for each task~\citep{wortsman2020supermasks,fernando2017pathnet}, or create implicit gateways for different tasks~\citep{mirzadeh2020dropout}. However, the expansion-based methods' memory and compute requirement grows as the number of tasks grows. In addition, these methods rely on the task identifiers for selecting the appropriate module for prediction and often cannot operate without this information.

Perhaps our work is mainly related to regularization- and memory-based methods. Our proposed algorithm maintains a memory of past data for regularization purposes (i.e., encouraging the connectivity between subspaces across tasks).

\textbf{Ensemble methods for Continual Learning.}
Ensemble models have been utilized in the literature on continual learning to mitigate catastrophic forgetting, as demonstrated in studies such as \cite{ROSE, coscl, dual_distillation}. For instance, \cite{dual_distillation} utilized distillation to maintain knowledge from previous experiences by ensembling past predictions with the current model. In \cite{ROSE}, a module was designed to detect data drifts, and an ensemble cooperation was used to distribute the data knowledge among each member. \cite{coscl} demonstrated that an ensemble of small models can outperform a larger one, but they mainly concentrated on parameter efficiency, which still incurs a linear compute cost, whereas our approach focuses on decreasing the compute cost. Although these studies also utilized ensemble methods, the direct comparison of compute cost is not always straightforward, and we aim to provide a comprehensive perspective on ensemble models dynamic in continual learning rather than ensembling specific algorithm.

\textbf{Mode connectivity.}
\cite{draxler2018essentially,modeconnectivity_main} studied the loss landscape and existence of connectivity of neural network solutions. They discovered the existence of pathways/curves of non-increasing loss between solutions optima. In the context of continual learning, \cite{mirzadeh2021MCSGD} have recently shown that multitask and continual minima are connected via low-loss paths and leveraged this property to connect tasks' minima in continual learning and proposed the MC-SGD algorithm, which encourages the linear connectivity between tasks' minima via path regularization. While our work is inspired by their findings, we note that MC-SGD is developed with a single model continual learning in mind, while we extend their work for the continual learning setup with multiple models.

\textbf{Neural network subspaces.} \cite{learning-neural-network-subspaces,benton2021} recently proposed to connect solution of an ensemble of models through a region in the weight space known as \textit{subspace}. That region is a low loss surface and shown to outperform the solution of ensemble models. \cite{gaya2021learning} learned a subspace of policies in the reinforcement learning context for fast adaptation. While our work is directly motivated by the subspace literature, we note that these methods have been studied in single-task settings. In contrast, our work extends them for continual learning with a sequence of tasks rather than a single task.

%% file: 7-conclusion.tex
\section{Conclusion}

While continual learning literature focuses mainly on studying the problem with a single model, we have extended it to the multiple models' case in this work. To the best of our knowledge, we are the first to investigate the behavior of ensemble methods in continual learning. Due to their enhanced diversity power, vanilla ensemble achieves good performance, but this comes at the expense of high computational cost growth. To overcome this challenge, and inspired by the recent advances in the mode connectivity literature, we have proposed a simple yet computationally efficient algorithm to improve the performance of subspace ensemble method. 
We believe it is important not to consider this seemingly simple method as the main contribution of our work but rather the understanding and implication of ensemble methods for continual learning. 

Despite the simplicity, ensemble methods provide a reliable mechanism to mitigate catastrophic forgetting in continual learning. The choice of an ensemble method for continual learning highly depends on the use case and limiting factors (e.g., performance and cost trade-off). 
We believe our work can be a stepping-stone for several future works, such as further studying the interactions between multiple models in continual learning and designing more efficient ensembling techniques.

\textbf{Limitation.} One limitation of our work is the dependence on a (small) sample of previous tasks to ensure the connectivity. Another limitation may arise due to the nature of mode connectivity and subspace methods because they decrease the capacity of the model for learning new tasks as an expense for increased stability. Finding the right trade-off between forgetting improvement (aka stability of previous knowledge) and learning accuracy (plasticity for learning new knowledge) is very interesting line of work for future. In this regard, one may look for explicitly employing parameter isolation or dynamic expansion methods (like adding new member to the ensemble) to cover for the decreased capacity of learning combined with a non-forgetting continual learning algorithm.

%% file: appendix.tex
\onecolumn
\section*{Appendix}
The appendix is organized as follows:
\begin{itemize}
   
    \item Appendix~\ref{sec:ensemble_method_for_cl} provides details on each of the Ensemble method (Vanilla, Subspace and Batch Ensemble)
     \item Appendix~\ref{sec:computation_cost} compares the computational cost implications of Vanilla and Subspace Ensemble method.
     \item Appendix~\ref{sec:pseudo_codes} provides pseudo-code of algorithms used in our ablations analysis and final experiments section.
    \item Appendix~\ref{sec:ablation_details} shows setting of our additional ablation analysis and provides additional insights on the diversity enhanced of Vanilla Ensemble method (Section~\ref{sec:2-beyond-one-model}).
     \item Appendix~\ref{sec:subspace_properties} provides an in-depth investigation of subspace learning.
    \item Appendix~\ref{sec:comparison_ensembles} provides detailed ablation results and comparison among Ensemble and Scaled models.
       \item Appendix~\ref{sec:final_exp_parameters} details  the experimental setup of our final results on $20$ tasks for MNIST and CIFAR-100 dataset (Section~\ref{sec:appendix-experiments-results})
\end{itemize}

\newpage


    
   


\section{Ensemble Methods for Continual Learning}
\label{sec:ensemble_method_for_cl}

\subsection{Training Vanilla Ensemble Models for Continual Learning}
In this work, we adapt the learning of ensemble methods to continual learning scenarios. Unless mentioned, we use the standard training procedure, which consists of training independently each of the $n$ models on the same dataset\footnote{We also compare against bagging strategy in Appendix~\ref{sec:comparison_ensembles}.}. Given a set of weights $\{ \omega_i \}_{i=1}^{n}$, a task $\tau$, when a data batch $(x,y)$ arrives, we optimize $\frac{1}{n}\sum_{i=1}^{n} \mathcal{L}_{\tau}(f_{\omega_i}(x),y)$. Therefore, the models' training differs only in the weights initialization of each model. The average prediction of each model is used as the output of the ensemble for evaluation ($\frac{1}{n}\sum_{i=1}^{n}f_{\omega_i}(.)$)\footnote{Classical majority voting strategies has been tried without significant difference in performance}. That being said, the training cost simply grows linearly with the number of models $n$. A pseudo-code of the Ensemble CL Algorithm is provided in Alg.~\ref{alg:enemble_cl} in the Appendix.

\subsection{Learning a Subspace Solution for  Continual Learning}
For the subspace method training, we proceed similarly as in \cite{learning-neural-network-subspaces}. Given a predictor $f$, a set of $n$ learnable parameters $\{ \omega_{i}\}_{i=1}^{n}$, learning a subspace of dimension $n$ consists in training the predictor $f_{\bar{\omega}}$ as accurate as possible (with $\bar{\omega}=\sum_{i=1}^{n} \alpha_i \omega_{i}$, $\pmb{\alpha} \in \Delta^{n}$). Simply put, given a task $\tau$, when a data batch $(x,y)$ arrives, we optimize $\mathcal{L}_{\tau}(f_{\bar{\omega}}(x),y)$ where $\bar{\omega}=\sum_{i=1}^{n} \alpha_i \omega_{i}$, with $\pmb{\alpha} \sim \mathcal{U}(\Delta^{n})$. Although, one can learn a distribution over the $\pmb{\alpha}$'s, we consider the standard case like in \cite{wortsman2020supermasks} where we sample it uniformly in the simplex $\Delta^{n}$. The prediction steps differs only in the choice of $\pmb{\alpha}$ and will be discussed in the following sections. A pseudo-code of the Subspace Continual Learning Algorithm is provided in Alg.~\ref{alg:subspace_cl} in the Appendix.

When it comes to backpropagation, subspace methods enjoy slightly similar computation cost as the single model. Given a loss $\mathcal{L}$, the update w.r.t. each $\omega_{i}$, is (assuming standard SGD optimizer): $\frac{\partial \mathcal{L}}{\partial \omega_{i}}=\frac{\partial \mathcal{L}}{\partial \bar{\omega}}\frac{\partial \bar{\omega}}{\partial \omega_{i}}=\alpha_{i}\frac{\partial \mathcal{L}}{\partial \bar{\omega}}$, $\forall i=1...n$. Then, only a \textit{unique} backpropagation through the model $f$  (w.r.t. to $\bar{\omega}$) is needed. A detailed discussion about the computational cost implication between ensemble and subspace methods is provided in Appendix~\ref{sec:computation_cost}.

\subsection{Batch Ensemble Solution for  Continual Learning}
Let's denote a weight $\omega \in \mathbb{R}^{m \times p}$ where $m$ and $p$ are respectively the input and output dimension. \cite{Wen2020BatchEnsemble} define a (common) slow weights $\omega$  and a set of fast weights $\{ s_i , r_i \}_{i=1}^{n}$ (with $s_i \in \mathbb{R}^{m}, r_i \in \mathbb{R}^{p}, \forall i=1...n$. Each member of the ensemble $i=1...n$ owns its set of weights $\omega_i= \omega \circ f_i$ with $f_i=r_i s_i^{T}$. Whenever an incoming batch arrive $(x,y)$, each element of the batch is randomly assigned to a member of the ensemble such that the forward pass (for a given layer) can be vectorized as:
$\phi( (x \circ R)\omega \circ S)$ where rows of $R$ (respectively $S$) consists of the vector $r_i$ (respectively $s_i$) for all examples of the batch $x$ and $\phi$ the non-linear activation layer.  We refer reader to Section 3.1 of ~\cite{Wen2020BatchEnsemble} for more details.

\newpage
\section{Algorithms pseudo-code}
\label{sec:pseudo_codes}
\begin{algorithm}
\footnotesize
 \SetKwInOut{Input}{Input}
    \SetKwInOut{Output}{Output}
    \Input{A task sequence $\cT_1, \cT_2, \ldots , \cT_{T}$, loss function $\mathcal{L}$ }
     Initialize weights $\omega_{0}^{*}$  \\
      \For{tasks  $\tau=1,2,3,\ldots T  $}{
      
        \For{$(x,y) \in \mathcal{T}_{\tau}$}{
          \,  \, \,   Get $f_{\omega_{\tau-1}}(x)$ \\
        Optimize $\mathcal{L}(f_{\omega_{\tau-1}}(x),y)$  \:  \tcp{Fine-tune on task $\tau$}
        }
        Return $\omega_{\tau}^{*}$

        }
    
    \Output{ $\omega_{T}^{*}$}
     \caption{Vanilla Continual Learning (Single Model)}
     \label{alg:single_cl}
\end{algorithm}

\begin{algorithm}
\footnotesize
 \SetKwInOut{Input}{Input}
    \SetKwInOut{Output}{Output}
    \Input{A task sequence $\cT_1, \cT_2, \ldots , \cT_{T}$, loss function $\mathcal{L}$, number of models $n$ }
     Initialize set of weights $\{ \omega_{i,0}\}_{i=1}^{n}$  \\
      \For{tasks  $\tau=1,2,3,\ldots T  $}{
      
        \For{$(x,y) \in \mathcal{T}_{\tau}$}{
         \,  \, \,   Get $f_{\omega_{i,\tau-1}}(x)$, $\forall i=1...n$ \\
        Optimize $\frac{1}{n}\sum_{i=1}^{n}\mathcal{L}(f_{\omega_{i,\tau-1}}(x),y)$  \:   \tcp{Update each model independently: requires $n$ backward passes}
        
        }
        Return $\{ \omega_{i,\tau}^{*} \}_{i=1}^{n}$

        }
    
    \Output{ $\{ \omega_{i,T}^{*} \}_{i=1}^{n}$}
     \caption{Vanilla Ensemble Continual Learning }
     \label{alg:enemble_cl}
\end{algorithm}

\begin{algorithm}
\footnotesize
 \SetKwInOut{Input}{Input}
    \SetKwInOut{Output}{Output}
    \Input{A task sequence $\cT_1, \cT_2, \ldots , \cT_{T}$, loss function $\mathcal{L}$, number of models $n$ }
      Initialize set of weights $\{ \omega_{i,0}\}_{i=1}^{n}$  \\
      \For{tasks  $\tau=1,2,3,\ldots T  $}{
      
        \For{$(x,y) \in \mathcal{T}_{\tau}$}{
        \; $\pmb{\alpha} \sim \mathcal{U}(\Delta^{n})$, $\bar{\omega}_{\tau-1}=\sum_{i}\alpha_i \omega_{i,\tau-1}$ \;  \tcp{Sample uniformly convex combination of weights}
         \: Get $f_{\bar{\omega}_{\tau-1}}(x)$ \\
        Optimize $\mathcal{L}(f_{\bar{\omega}_{\tau-1}}(x),y)$
        }
        Return $\{ \omega_{i,\tau}^{*} \}_{i=1}^{n}$

        }
    
    \Output{ $\{ \omega_{i,T}^{*} \}_{i=1}^{n}$}
     \caption{Subspace Ensemble Continual Learning}
     \label{alg:subspace_cl}
\end{algorithm}

\begin{algorithm}
\footnotesize
 \SetKwInOut{Input}{Input}
    \SetKwInOut{Output}{Output}
    \Input{A task sequence $\cT_1, \cT_2, \ldots , \cT_{T}$, loss function $\mathcal{L}$, number of models $n$ }
      Initialize set of weights $\{ \omega_{i,0}\}_{i=1}^{n}$  \\
      \For{tasks  $\tau=1,2,3,\ldots T  $}{
        \For{$(x,y) \in \mathcal{T}_{\tau}$}{
            \For{$each layer$}{
         
             Get $R$ and $S$ using $r_i$ and $s_i$, $i=1...n$\;  \tcp{Assign each element of the batch randomly to a member of the ensemble}   
         Compute $W=((x_{out} \circ R)\omega \circ S)$  \; \tcp{$x_{out}$ being the output of the previous layer}
             }
            Optimize $\mathcal{L}(f_{W}(x),y)$
            }
        Return $\{ \omega_{\tau}^{*} ,s_{\tau}, r_{\tau} \}_{i=1}^{n}$
        }
    \Output{ $\{ \omega_{i,T}^{*} \}_{i=1}^{n}$}
     \caption{Batch Ensemble Continual Learning
}
     \label{alg:batch_ensemble_cl}
\end{algorithm}

\begin{algorithm}
\footnotesize

 \SetKwInOut{Input}{Input}
    \SetKwInOut{Output}{Output}
    \Input{A task sequence $\cT_1, \cT_2, \ldots , \cT_{T}$, loss function $\mathcal{L}$, number of models $n$, replay buffer $\mathcal{B}$, $m_{\mathcal{B}}$ memory size per task }
      Initialize set of weights $\{ \omega_{i,0}\}_{i=1}^{n}$  \\
      \For{tasks  $\tau=1,2,3,\ldots T  $}{
      
        \For{$(x,y) \in \mathcal{T}_{\tau}$}{
        \; $\pmb{\alpha} \sim \mathcal{U}(\Delta^{n})$, $\bar{\omega}_{\tau-1}=\sum_{i}\alpha_i \omega_{i,\tau-1}$ \;  
        \tcp{Sample uniformly convex combination of weights} 
        \; $(x',y') \sim \mathcal{B}$ \;  \tcp{Samples elements from the buffer} 
        \; $x \leftarrow \text{Concat}(x,x')$ \\
        \; $y \leftarrow \text{Concat}(y,y')$ \\
         \: Get $f_{\bar{\omega}_{\tau-1}}(x)$ \\
        Optimize $\mathcal{L}(f_{\bar{\omega}_{\tau-1}}(x),y)$
        }
        Return $\{ \omega_{i,\tau}^{*} \}_{i=1}^{n}$ \\
          \For{$(x, y) \in \mathcal{T}_{\tau}$}
         {$\mathcal{B} \leftarrow \cB \bigcup \{x,y \}$ \;  \tcp{Store $m_{\mathcal{B}}$ elements per class per task in the buffer}
         }
        }
    \Output{ $\{ \omega_{i,T}^{*} \}_{i=1}^{n}$}
     \caption{Subspace + Experience Replay (ER)}
     \label{alg:subspace_er_cl}
\end{algorithm}

\begin{algorithm}[t]
\footnotesize
\begin{spacing}{0.8}
    \SetKwInOut{Input}{Input}
    \SetKwInOut{Output}{Output}
    \SetKw{KwBy}{by}
    \SetAlgoLined
    \Input{A task sequence $\cT_1, \cT_2, \ldots , \cT_{T}$, number of models $n$ , buffer $\cB$ and memory size $m_{\cB}$ }
    \begin{enumerate}
        \item Initialize set of weigths $S^{n}=\{\omega_{0,i} \}_{i=1}^{n}$, buffer $\mathcal{B} \leftarrow \{  \}$
        \item \For{tasks  $\tau=1,2,3,\ldots T  $}{
            Get $\{ \hat{\omega}^{*}_{\tau,i} \}_{i=1}^{n}$ with Eq \ref{eq:w_hat} \\
        \tcp{Learn subspaces solution for task $\tau$}
          \For{$(x, y) \in \mathcal{T}_{\tau}$}
         {$\mathcal{B} \leftarrow \cB \bigcup \{x,y \}$ }
      Get $\{ \omega^{*}_{\tau,i} \}_{i=1}^{n}$ with Eq \ref{eq:w_bar} using samples from $\mathcal{B}$ \\
      \tcp{Connect previous solution' subspaces}
        }
    \end{enumerate}
    \Output{ $\{ \omega^{*}_{T,i} \}_{i=1}^{n}$}
    \caption{\algoname CL}
\end{spacing}
\end{algorithm}
\clearpage
\section{Computational cost implications for ensemble and subspace methods}
\label{sec:computation_cost}
This section discusses the implication cost of Vanilla and Subspace Ensemble methods.

For the ensemble methods, since each model is trained independently on the same dataset, the cost is $n$ times higher ($n$ times more forward and backward passes).

Compared to the single model, the subspace method contains one additional operation for the inference and backpropagation operation. Before the inference, a convex combination of the weights is sampled: $\bar{\omega}=\displaystyle{\sum_{i=1}^{n}}\alpha_i \omega_{i}$, $\alpha \sim \mathcal{U}(\Delta^{n})$ (weights mixing). For the backpropagation, only one backward pass is needed ($\frac{\partial \mathcal{L}}{\partial \bar{\omega}}$) before assigning the new weights $\omega_{i}=\omega_{i}- l_{r} \alpha_{i}\frac{\partial \mathcal{L}}{\partial \bar{\omega}}$, $\forall i=1 ...n$, $l_r$ being the learning rate.

Overall, the subspace method has only two supplementary addition operations as an overhead cost compared to the single model which is much cheaper than the $n-1$ additional forward and backward pass of the ensemble methods. This makes the subspace method more efficient.

\begin{table*}[h!]
\centering

\resizebox{0.8\textwidth}{!}{%
\begin{tabular}{l|c|c}
\hline
\multirow{2}{*}{Method}     & \multirow{2}{*}{forward pass} & \multirow{2}{*}{backward pass}  \\ 
& & \\\cline{1-3} 
Single model  & $1$ inference pass ($f(x,\omega$) & $1$ backpropagation pass ( $\frac{\partial \mathcal{L}}{\partial \omega}$ )\\ \hline
  Ensemble  &  \textcolor{red}{$n$ inference passes ($f(x,\omega_{i})$, $\forall i=1...n$)} & 
  \textcolor{red}{$n$ backpropagations passes  ( $\frac{\partial \mathcal{L}}{\partial \omega_i}, \; \forall i=1...n$ ) }  \\ \hline
\multirow{2}{*}{Subspace}         &  \textcolor{red}{$\bar{\omega}=\displaystyle{\sum_{i=1}^{n}}\alpha_i \omega_i$} &  $1$ backpropagation pass  ( $\frac{\partial \mathcal{L}}{\partial \bar{\omega}}$ )  \\  
& $1$ inference pass ($f(x,\bar{\omega}$) & \textcolor{red}{$\omega_{i}\leftarrow \omega_{i}-l_{r} \alpha_i \frac{\partial \mathcal{L}} {\partial \bar{\omega}}, \forall i=1...n$}  \\\hline

\end{tabular}%
}
\caption{ Computational cost comparison between different methods. In \textcolor{red}{red} are the additional operations compared to the single model. While ensemble methods have $(n-1)$ additional backward passes (backward pass), subspace methods only incur addition operations as an overhead cost which are much cheaper.}
\end{table*}
\newpage
\section{Setup of ablation and additional results}
\label{sec:ablation_details}

\subsection{Experimental parameters}
For the ablation of Section~\ref{sec:2-beyond-one-model}, we train on $5$ tasks for Rotated MNIST and Split CIFAR-100.  The details of the setup are the following:

\paragraph{Rotated MNIST}: The incremental angle is $22.5^{\circ}$ for a total of $90^{\circ}$. This made the benchmark more challenging as done in~\cite{understanding_continual} with one training epoch per task. 

\paragraph{Split CIFAR-100}: We use the $25$ first tasks of CIFAR-100 and train for $5$ epochs per task to highlight the properties of ensemble methods.

The hyperparameters for each baseline are the following:
\subsubsection*{Naive SGD and Vanilla Ensemble (VE) and Batch Ensemble (BE)}
\begin{itemize}
    \item learning rate:  [0.2, 0.15, \textbf{0.1} (MNIST), \textbf{0.05} (CIFAR-100), 0.01]
    \item learning rate decay: [\text{1.0} (CIFAR-100)\footnote{A value of $1$ means no decay has been applied to the learning rate}, 0.95, 0.9, \text{0.5} (MNIST)]
    \item batch size: [10,32,64,128]
\end{itemize}

\subsubsection*{Subspace Ensemble (SE)}
\begin{itemize}
    \item standardized learning rate\footnote{This is an average learning rate per model, i.e we use a learning rate of $0.3$ when using $n=3$ models for MNIST}:  [0.2, \textbf{0.15} (CIFAR-100), \textbf{0.1} (MNIST)] 
    \item learning rate decay: [\text{1.0} (CIFAR-100), 0.95, 0.9, \text{0.5} (MNIST)]
    \item batch size: 10
\end{itemize}
An other important aspect of subspace methods are how close each model is initialized. The further away from each other they are, the more iteration will be needed to update the whole volume. To control this volume, we initialize each model's weight with respect to the first one as: $\omega_{i}= \omega_{1} * \mathcal{N}(1,\sigma_{init})$, $\forall i=1...n$, $\omega_1$ being the initialized weight of the first model. The variance $\sigma_{init}$ controls this volume. For MNIST, we use $\sigma_{init}=1.0$ ($n=2,3$) and $\sigma_{init}=1.5$ ($n=5,10$), for CIFAR-100 we use $\sigma_{init}=0.1$ ($n=2,3,5,10$).

\subsection{Metrics}
To quantify the properties of ensemble and subspace methods, we have used the following metrics: \textbf{Learning Accuracy} $LA_{T}=\frac{1}{T}\displaystyle{\sum_{\tau=1}^{T}}a_{\tau,\tau} \; $, where $a_{\tau,\tau}$ represents the accuracy on task $\tau$ after learning on task $\tau$ the first time, \textbf{Forgetting} $F_T=\frac{1}{T-1}\sum_{\tau=1}^{T-1} \max_{t=\{1..T-1 \}} (a_{t,\tau}-a_{T,\tau})$ and \textbf{Forgetting Improvement} $FI_{T}=F_{T}(\textit{single model})-F_{T}(\textit{ensemble/subspace model})$


\newpage
\subsection{Diversity of performance of the Vanilla Ensemble methods}
\begin{figure}[h!]
    \centering
    \includegraphics[width=1.0\linewidth,height=0.8\linewidth,keepaspectratio=True]{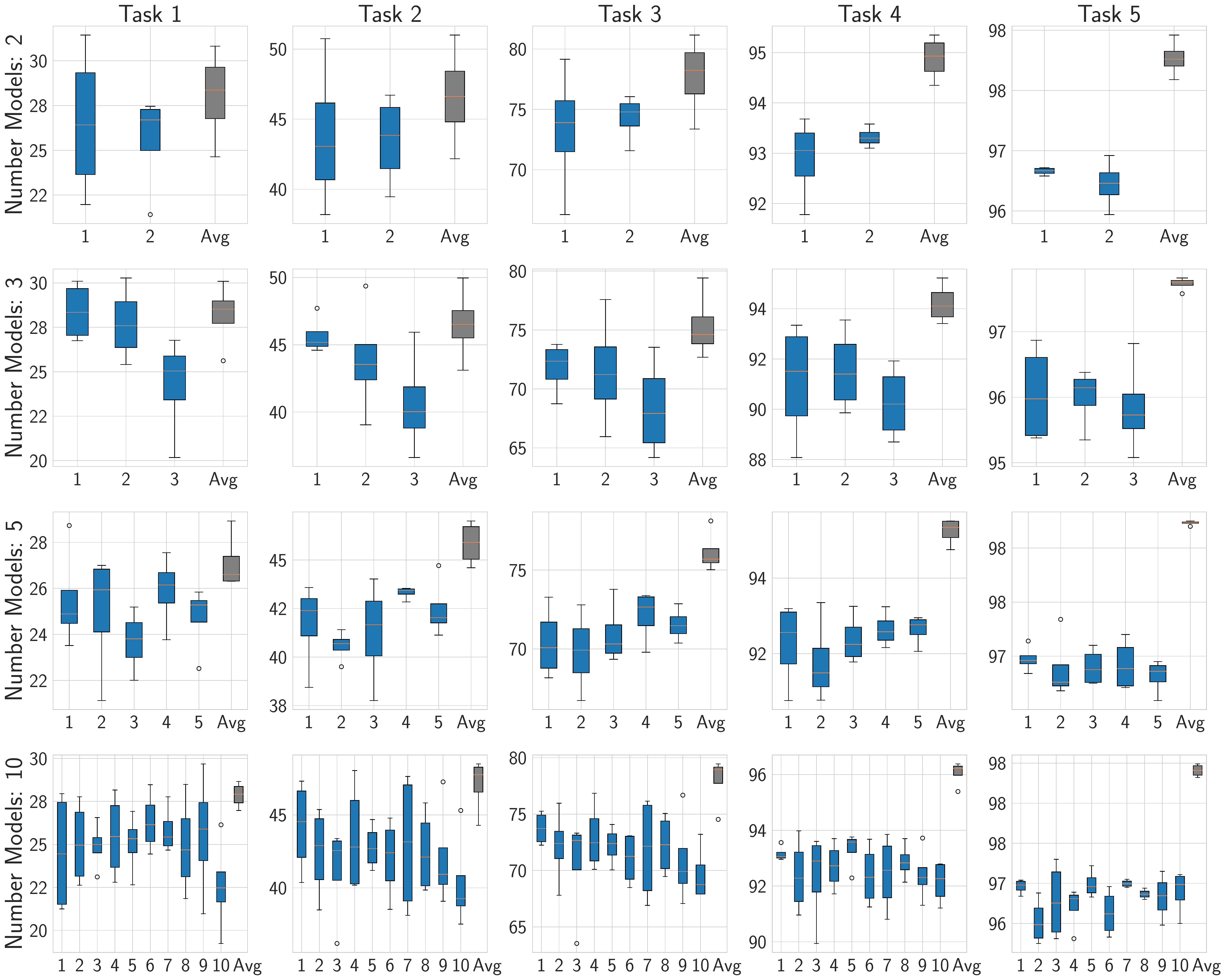}
    \caption{Final accuracy for each model of the ensemble (blue) and the final ensemble prediction (gray) for each task for Rotated MNIST. The final ensemble prediction accuracy is almost always higher than the best accuracy of the best model. As we increase the number of models, we can see a diversity in the prediction of each model, each of them seems to specialize naturally in diverse tasks (look at $n=5,10$). This might contribute to the good performance of the ensemble method.}
    \label{fig:diversity_ensemble_rotation}
\end{figure}

To measure the diversity of each member in the ensemble method (Ensemble Continual Learning Alg.~\ref{alg:enemble_cl}), we show the boxplot of the final average accuracy for each member of the ensemble (line) on each task (column) in Figure~\ref{fig:diversity_ensemble_rotation}. The x-axis represents each individual in the ensemble while the last index represents the final prediction of the ensemble (noted "Avg"). As we increase the number of models, the gap between the highest accuracy of each individual and the final prediction increases (For instance, compare gap between blue and gray boxplot for the 3rd column). The diversity in each model (initialized differently) might explain good performance of ensemble methods.
\newpage

\begin{figure}[h!]
    \centering
    \includegraphics[width=1.0\linewidth,height=0.8\linewidth,keepaspectratio=True]{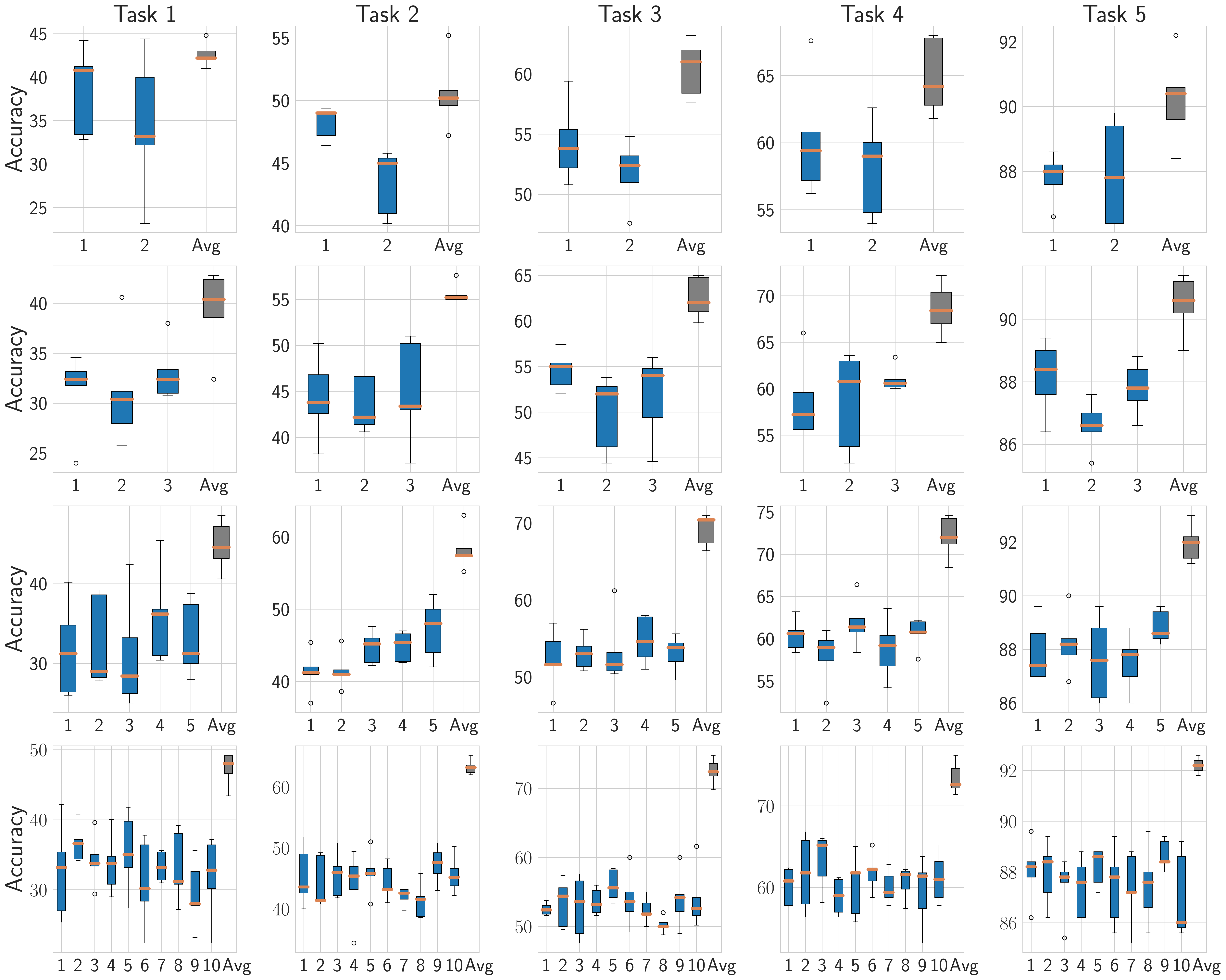}
    \caption{Final accuracy for each model of the ensemble (blue) and the final ensemble prediction (gray) for each task for Split CIFAR-100. Note how the ensemble accuracy (grey) is almost always much beter than the accuracy of the best member (blue).}
    \label{fig:diversity_ensemble_cifar}
\end{figure}


\subsection{Diversity in prediction of Vanilla Ensemble methods}
\begin{figure}[h!] \centering \includegraphics[width=1.0\linewidth,keepaspectratio=True]{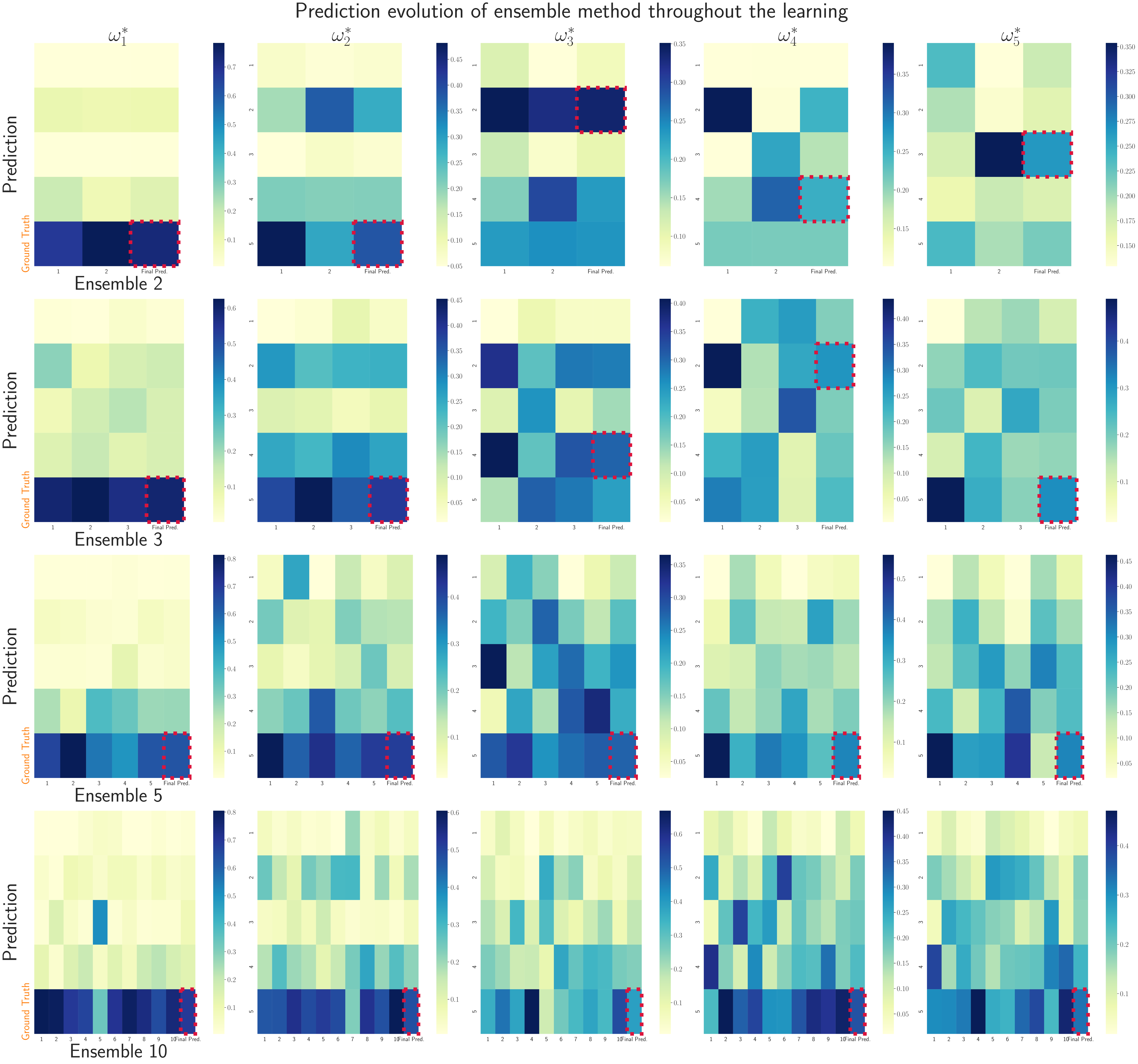} 
\negspace{-2mm}
\caption{Evolution of prediction for Split CIFAR-100 on one sample of task $1$. The y-axis represents the prediction for each class $1$ to $5$ while the x-axis represents the behavior of member $i$. Throughout the training (read from left to right), predictions of former's solution is forgotten leading to diversity as well. The more member an ensemble has, the more likely the former solution's can be remembered since the average prediction is used as the final output (look at last row 4th and 5th column).}
\label{fig:prediction_evolution}
\negspace{-2mm}
\end{figure}

We here provide a visualization of the prediction's evolution throughout the learning experience for the $5$ tasks ablation on Split CIFAR-100 for Vanilla Ensemble (average prediction). Figure~\ref{fig:prediction_evolution} shows the output of each member of an ensemble for a given sample of task $1$ ( the column represents the weights $\omega_{i}^{*}$ corresponding to the current task $i$). The y-axis represents the possible prediction (class $1$ to $5$) with the Ground truth being highlighted in orange (class $1$ here). The dashed red square is the final prediction made by the ensemble.

Throughout the learning, one can see the forgetting about solution's $1$ prediction that leads to diversity in prediction (read the 2nd row for instance). This obviously leads to forgetting of former's solution. The more member an ensemble has (take last row for instance), the more likely it will be able to keep its former's solution has a main core member sticks around the former's solution (compare 4th column).

\clearpage
\newpage
\subsection{Understanding subspace properties}
\label{sec:subspace_properties}

This section aims at showing the dynamic of subspace through the training process. We will show that applying naively Subspace (SE) and adding Experience Replay (ER) still show high forgetting.

To this end, we raise two questions: (1) ``What are the interesting properties of the learned subspaces in continual learning?'' and (2) ``How do subspaces evolve throughout the learning experience?''. In Sec.~\ref{sec:inside-subspace}, we study the first question and highlight an important property of the subspace method, that is, the center of the subspace contains more accurate and robust solutions. To answer the second question, in Sec.~\ref{sec:subspace-dynamic}, we show the limit of the subspace method that can still suffer from the forgetting problem when the number of tasks increases.  

For this purpose we design \algoname exploits the connectivity of the subspaces throughout the continual learning. By connectivity, we mean that there is a path between two solutions along which the loss value and the test error stay low~\cite{draxler2018essentially,modeconnectivity_main}.

\subsection{ Subspace midpoint gets the best accuracy }
\label{sec:inside-subspace}
In this section, we further investigate the behavior  of the learned subspaced ensembles by monitoring the accuracy within the subspace.

\begin{figure*}[h]
\centering
\begin{subfigure}[t]{.3\textwidth}
    \centering
    \includegraphics[width=00.78\textwidth,keepaspectratio=True]{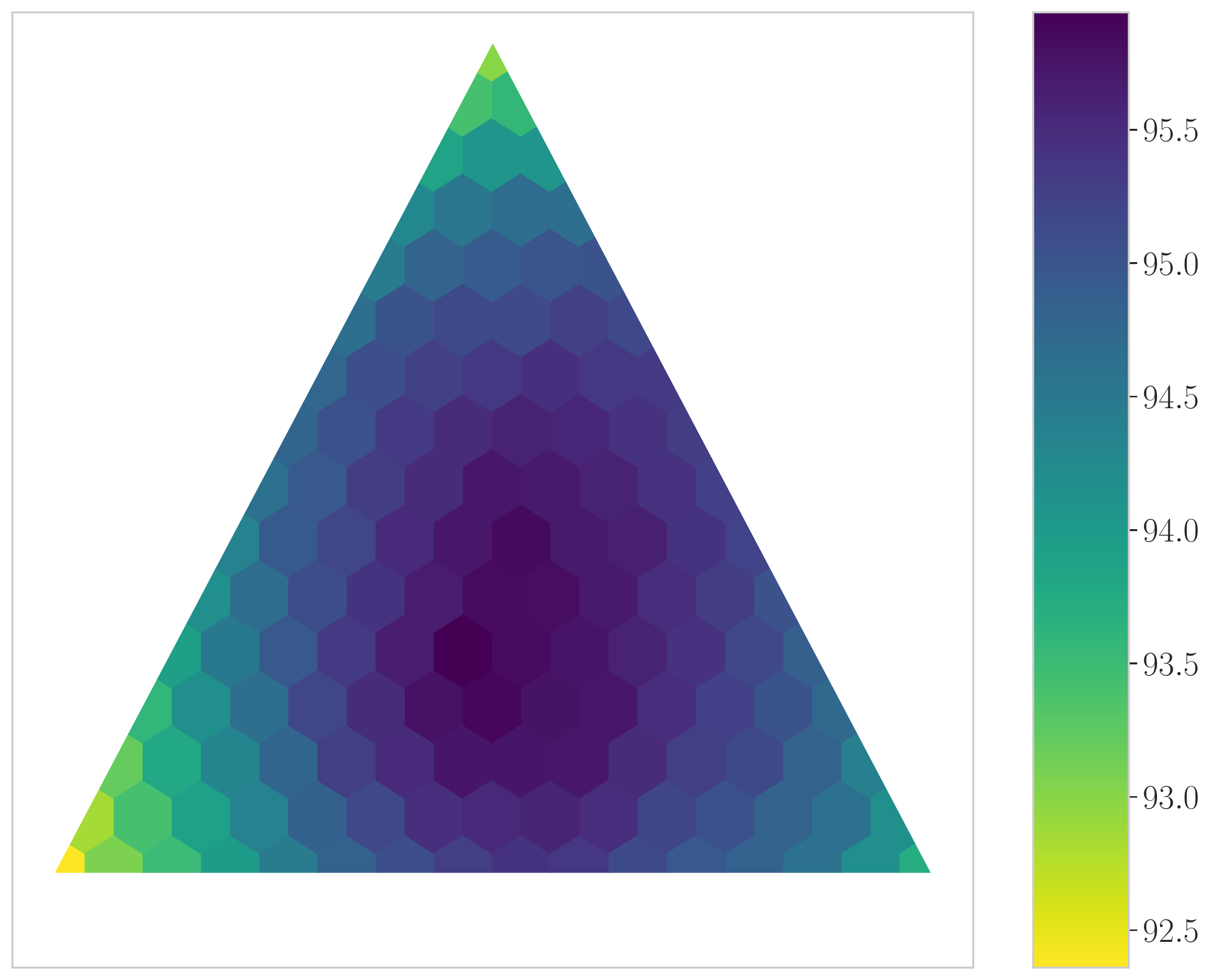}
    \caption{Learning accuracy of task 1}
    \label{fig:simplex_accuracy_1}
\end{subfigure}\hfill
\begin{subfigure}[t]{.3\textwidth}
    \centering
    \includegraphics[width=00.78\textwidth,keepaspectratio=True]{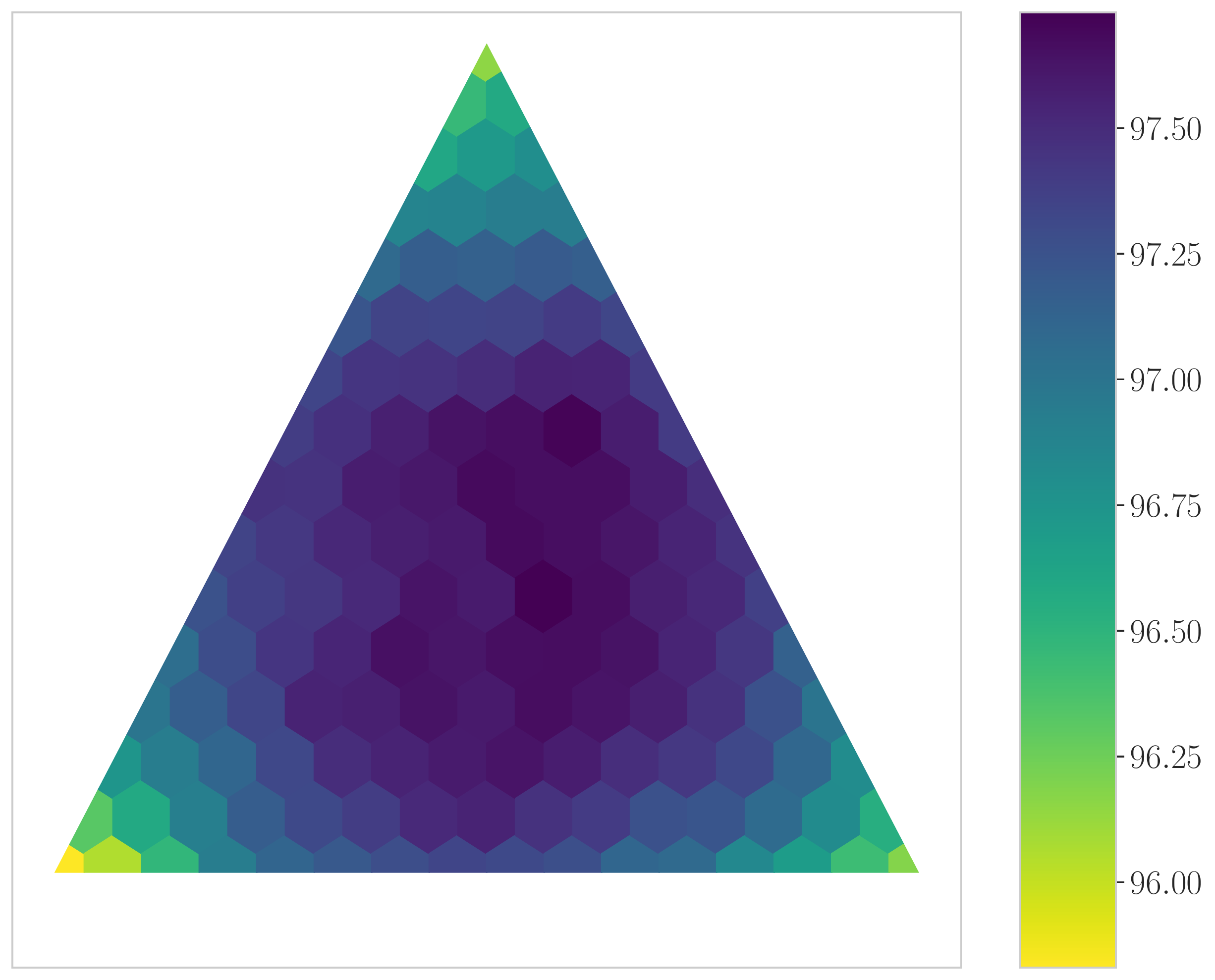}
    \caption{Learning accuracy of task 10}
    \label{fig:simplex_accuracy_2}
\end{subfigure}
\hfill
\begin{subfigure}[t]{.3\textwidth}
    \centering
    \includegraphics[width=0.80\textwidth,keepaspectratio=True]{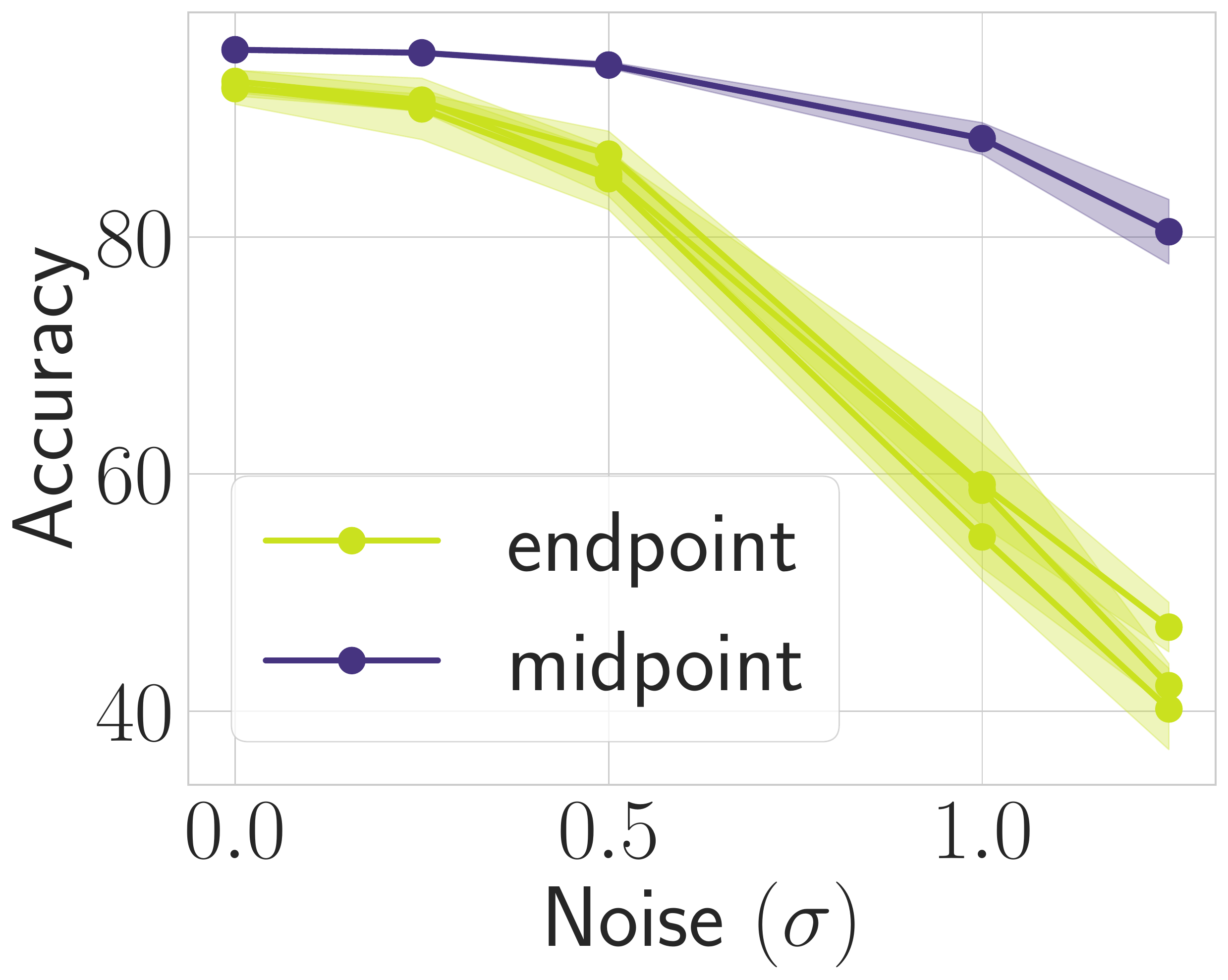}
    \caption{Robustness to Gaussian noise around solution's minima.}
    \label{fig:robust_to_noise}
\end{subfigure}
\centering
\caption{Rotated MNIST: Learning Accuracy for different tasks inside a subspace region with 3 models. Midpoints (center) show the best performance as opposed to the endpoints (corners) \textbf{(left, middle)}. Robustness to uniform noise for the case of $n=3$ models. Midpoint is more robust to weight perturbation than endpoints \textbf{(right)}}
\label{fig:simplex_accuracy}
\negspace{-3mm}
\end{figure*}

To this end, we use Subspace Continual Learning algorithm (Alg.~\ref{alg:subspace_cl}) with $n=3$ models. In Figs.~\ref{fig:simplex_accuracy_1} and~\ref{fig:simplex_accuracy_2} we show the accuracy across the subspace at the beginning of learning (task 1) and in the middle of learning (task 10), respectively. The plots illustrate that the center (midpoint) has higher learning accuracy within each region. This also translates into more robustness of the midpoint versus the endpoints (Fig.~\ref{fig:robust_to_noise}).

\begin{figure}[t] \centering
\includegraphics[width=0.7\linewidth,keepaspectratio=True]{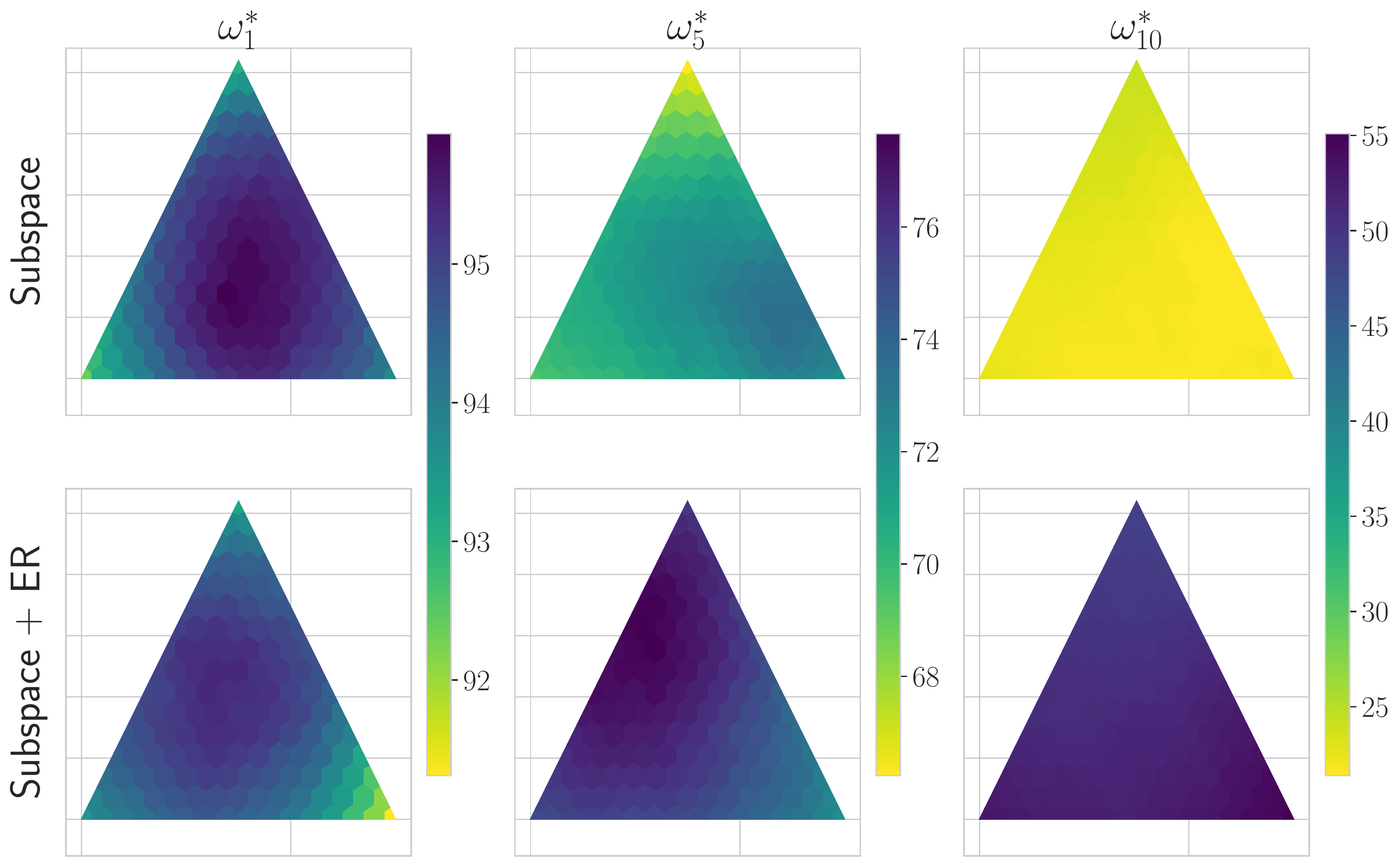}
\negspace{-2mm}
\caption{Rotated MNIST: The evolution of accuracy of task 1 throughout the training. Even though the subspace method restricts weight's movement, there is still a noticeable degradation in performance (hence a large amount of forgetting).}
\label{fig:evolution_simplex_accuracy_main}
\negspace{-2mm}
\end{figure}

\subsection{Subspaces still forget}
\label{sec:subspace-dynamic}

Now, we shift our focus on the evolution of the learned subspaces throughout the continual learning experience. To this end, we compare the subspace method (Alg.~\ref{alg:subspace_cl}) with and without Experience Replay, ER \cite{riemer2018learning}, (Alg.~\ref{alg:subspace_er_cl}) . Fig.~\ref{fig:evolution_simplex_accuracy_main} shows snapshots of task $1$'s accuracy across different times of the training (task $\tau=1,5,10$) on Rotated MNIST ($20$ tasks). We can observe that while adding ER improves the accuracy, over long sequences of tasks, there is still a degradation of performance. For instance, on task $1$ Subspace + ER incurs a performance decrease to $55\%$ (last column $\omega^{*}_{10}$).

\begin{figure}[h!] \centering \includegraphics[width=0.4\linewidth,keepaspectratio=True]{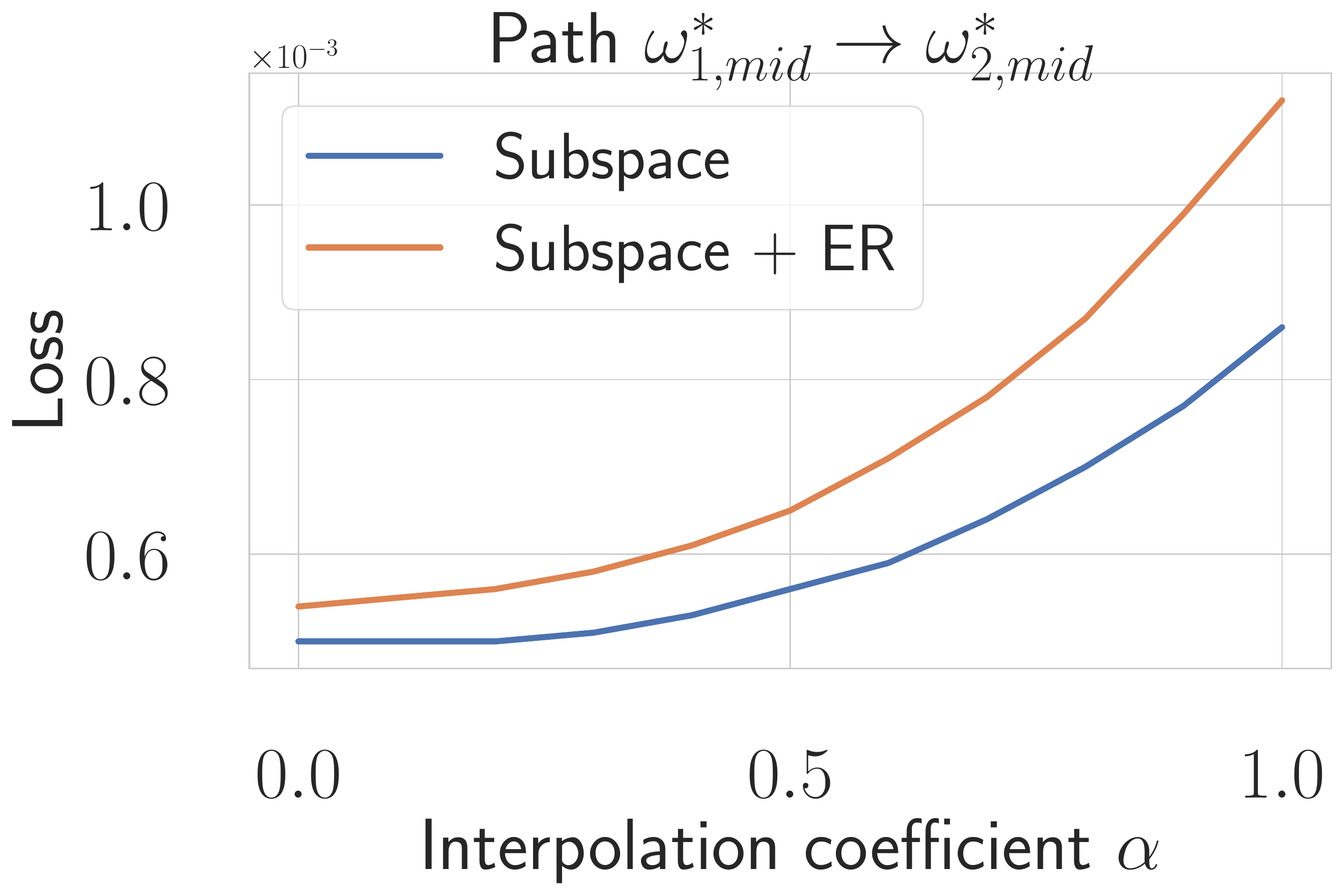} 
\negspace{-2mm}
\caption{Rotated MNIST: The sequential subspace solutions are not linearly connected.}
\label{fig:connectivity_naive_subspace}
\negspace{-2mm}
\end{figure}

To investigate this performance degradation, we employ the recent technique introduced by~\cite{mirzadeh2021MCSGD} where the authors show that the linear mode connectivity can explain the performance gap between continual and multitask solutions. Since the subspace method is inherently motivated by the mode connectivity, we believe investigating the connectivity across subspaces can explain the performance degradation. Hence, we visualize the mode connectivity between successive solutions (represented by midpoints) of task 1 and task 2, evaluated on the loss of task $1$. In Fig.~\ref{fig:connectivity_naive_subspace} we can see that the midpoints of subsequent subspaces' solutions are not \textit{linearly connected} meaning the linear interpolated weights between these two solutions does not stay in a low-loss region \cite{mirzadeh2021MCSGD}. Fig~\ref{fig:connectivity_mosaiq} in Appendix provides more detailed analysis and metrics (including accuracy) with the same conclusion.

Thus far, we have investigated the properties of the subspaces and their evolution throughout the learning experience. 

By tracking the evolution of the subspace, we have also observed a performance degradation, which can be explained by the lack of connectivity between solutions. These results motivate us to design \algoname that takes these findings into action. In the next section, we highlight the improve in mode connectivity of our algorithm against Subspace Ensemble and Subspace + ER.

\clearpage
\subsubsection{Subspace connectivity}
\label{sec:subspace_connectivity}
\begin{figure}[h!]
    \centering
    \includegraphics[width=0.9\linewidth,height=0.7\linewidth,keepaspectratio=True]{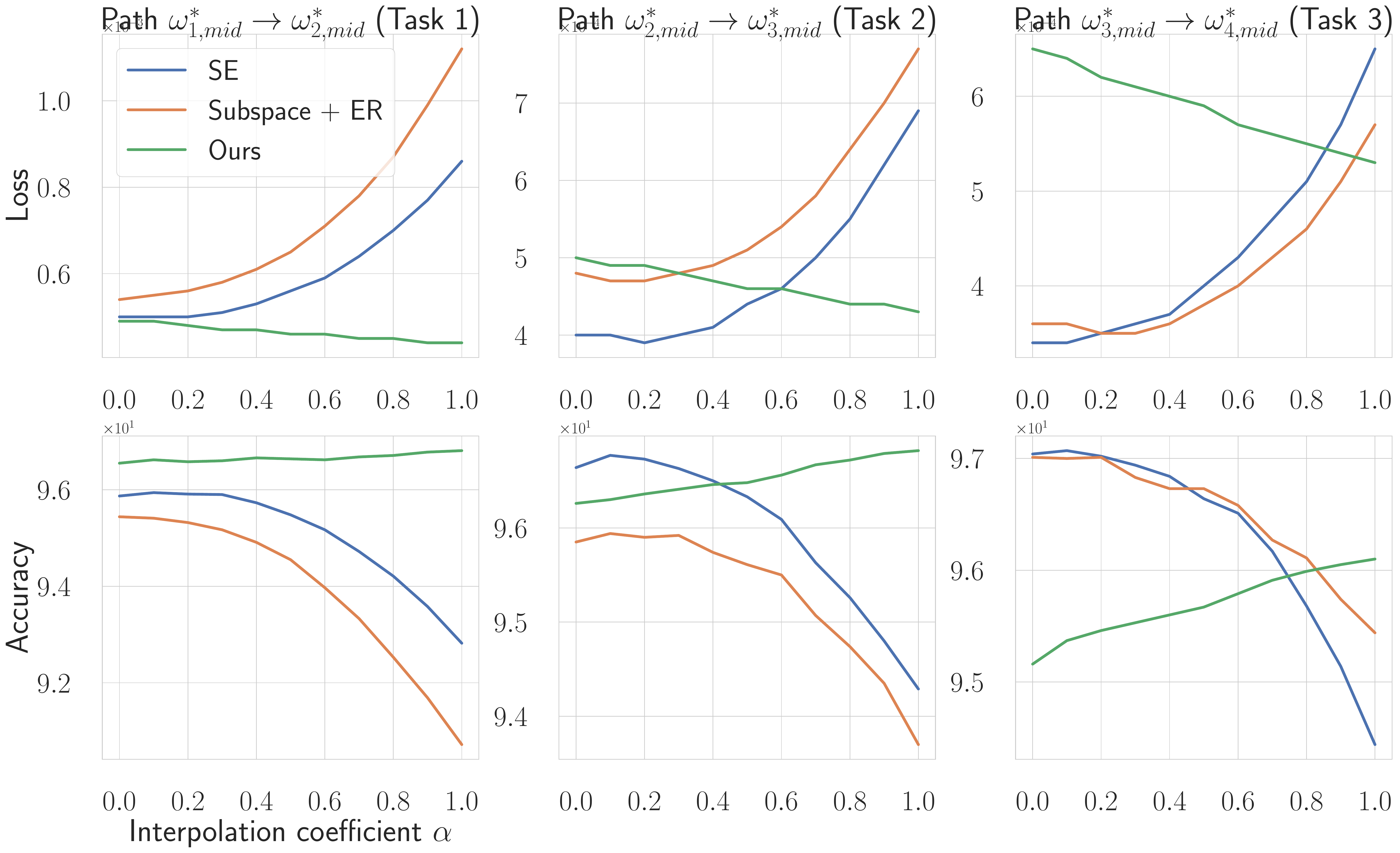}
    \caption{Connectivity comparison between \algoname and other baselines. Naive and ER+Subspace incurs high variation when interpolating between successive solutions since they do not enforce mode connectivity.}
    \label{fig:connectivity_mosaiq}
\end{figure}

 To measure the mode connectivity~\cite{mirzadeh2021MCSGD} between subsequent solutions, we evaluate the loss function between interpolated solutions $\alpha \omega_{i}^{*} + (1-\alpha) \omega_{i+1}^{*}$, $\forall \alpha \in [0,1]$. Fig~\ref{fig:connectivity_mosaiq} provides a comparison between our method \algoname against Subspace CL (Alg.~\ref{alg:subspace_cl}) and Subspace + ER (Alg.~\ref{alg:subspace_er_cl}). We clearly see that the two latter incurs a high variation in their loss and accuracy between successive solutions on task $1$'loss. As an example, while \algoname incurs a slight variation of $1\%$ in accuracy when interpolating between $\omega^{*}_{2}$ and $\omega^{*}_{3}$ (2nd column) the other methods incur a drop of more than $2\%$ of accuracy. This drop is even larger when interpolating between  $\omega^{*}_{3}$ and $\omega^{*}_{4}$ (3rd column). Now we are interested to investigate the mode connectivity for further task's solution of \algoname.

\begin{figure}[h!]
    \centering
    \includegraphics[width=1.0\linewidth,height=1.0\linewidth,keepaspectratio=True]{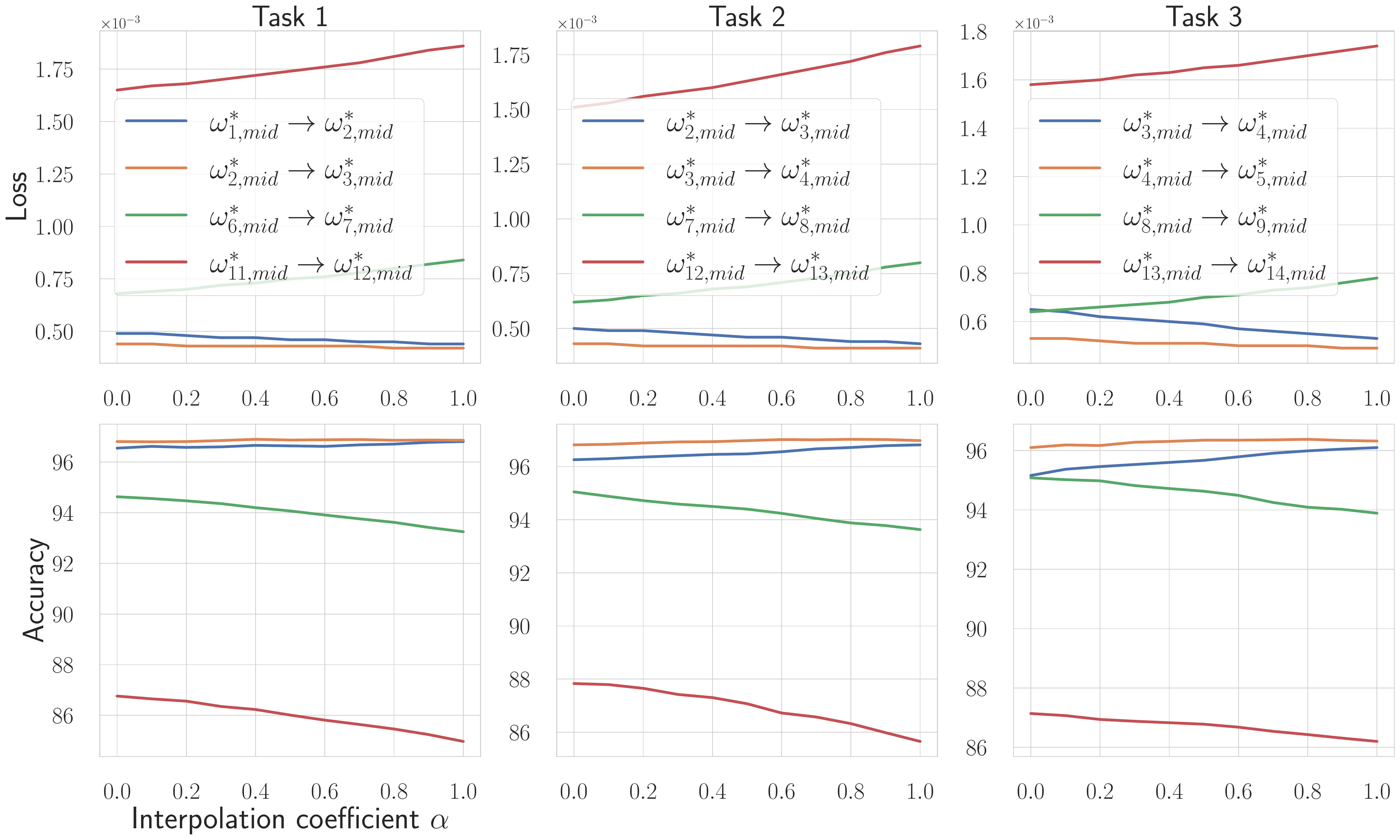}
    \caption{Subsequent solutions of \algoname maintain more or less connectivity (horizontality of each curve). }
    \label{fig:connectivity_ours_mosaiq}
\end{figure}

Figure~\ref{fig:connectivity_ours_mosaiq} shows the linear connectivity between solutions more than $5$ tasks further in time. Although performance might have decreased between $\omega^{*}_{11}$ and $\omega^{*}_{2}$ on task $1$, we can see that the linear path between $\omega^{*}_{11}$ and $\omega^{*}_{12}$ (red line) is close to horizontal losing roughly $2 \%$ of accuracy (first column) and $1 \%$ if one considers the path between $\omega^{*}_{6}$ and $\omega^{*}_{7}$ (green line). Note the high accuracy of $\omega^{*}_{6} \rightarrow \omega^{*}_{7}$ on task $1$ (first  column, green line) which reaches $\sim 94 \%$ of accuracy while the accuracy of the first task (with $\omega^{*}_{1}$) was around $96 \%$.

\begin{figure*}[t!]
\centering
\begin{subfigure}{.32\textwidth}
    \centering
    \includegraphics[width=1.0\textwidth,height=1.0\textwidth,keepaspectratio=True]{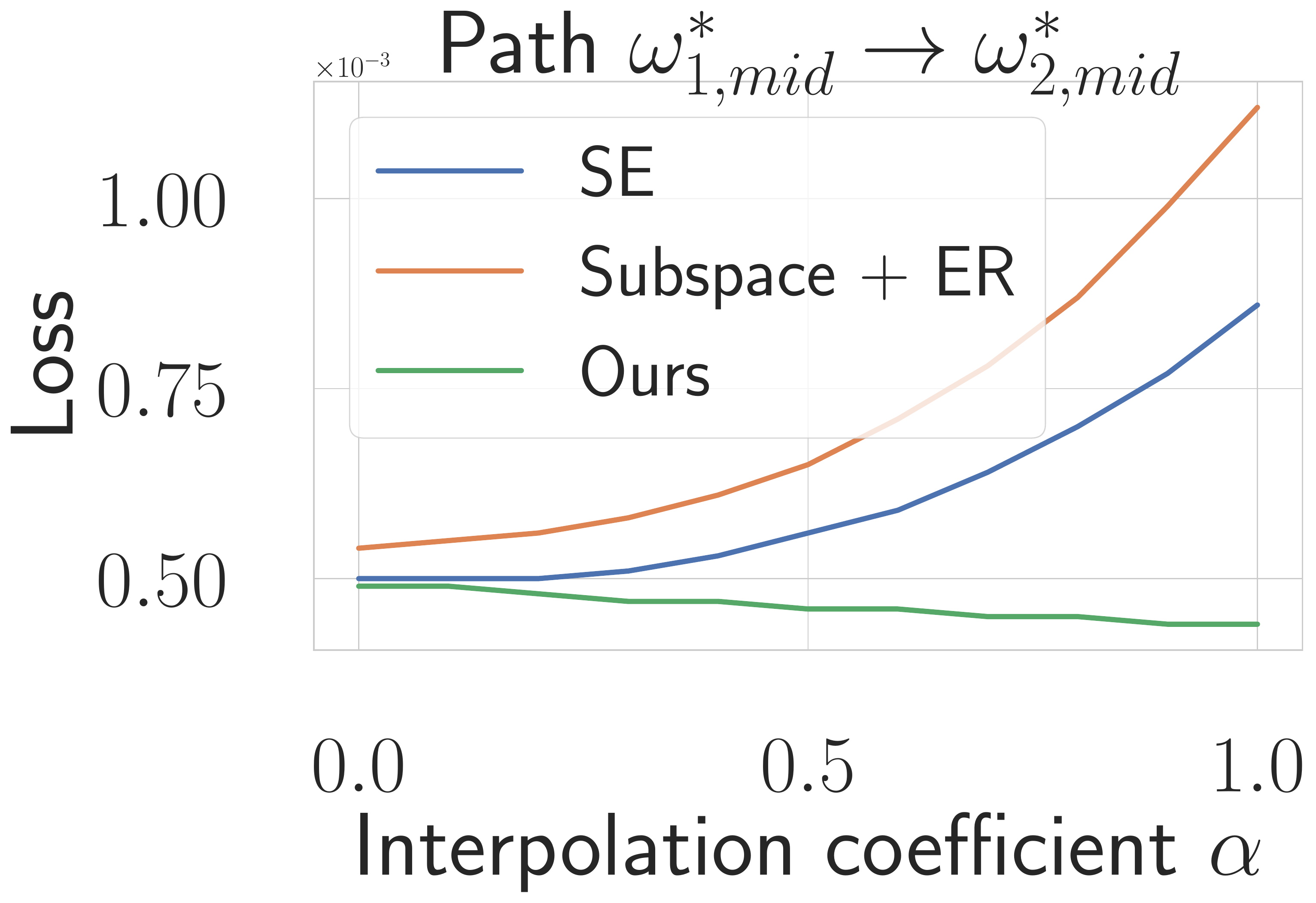}
    \caption{}
    \label{fig:connectivity_loss_1}
\end{subfigure}\hfill
\begin{subfigure}{.32\textwidth}
    \centering
    \includegraphics[width=1.0\textwidth,height=0.7\textwidth,keepaspectratio=True]{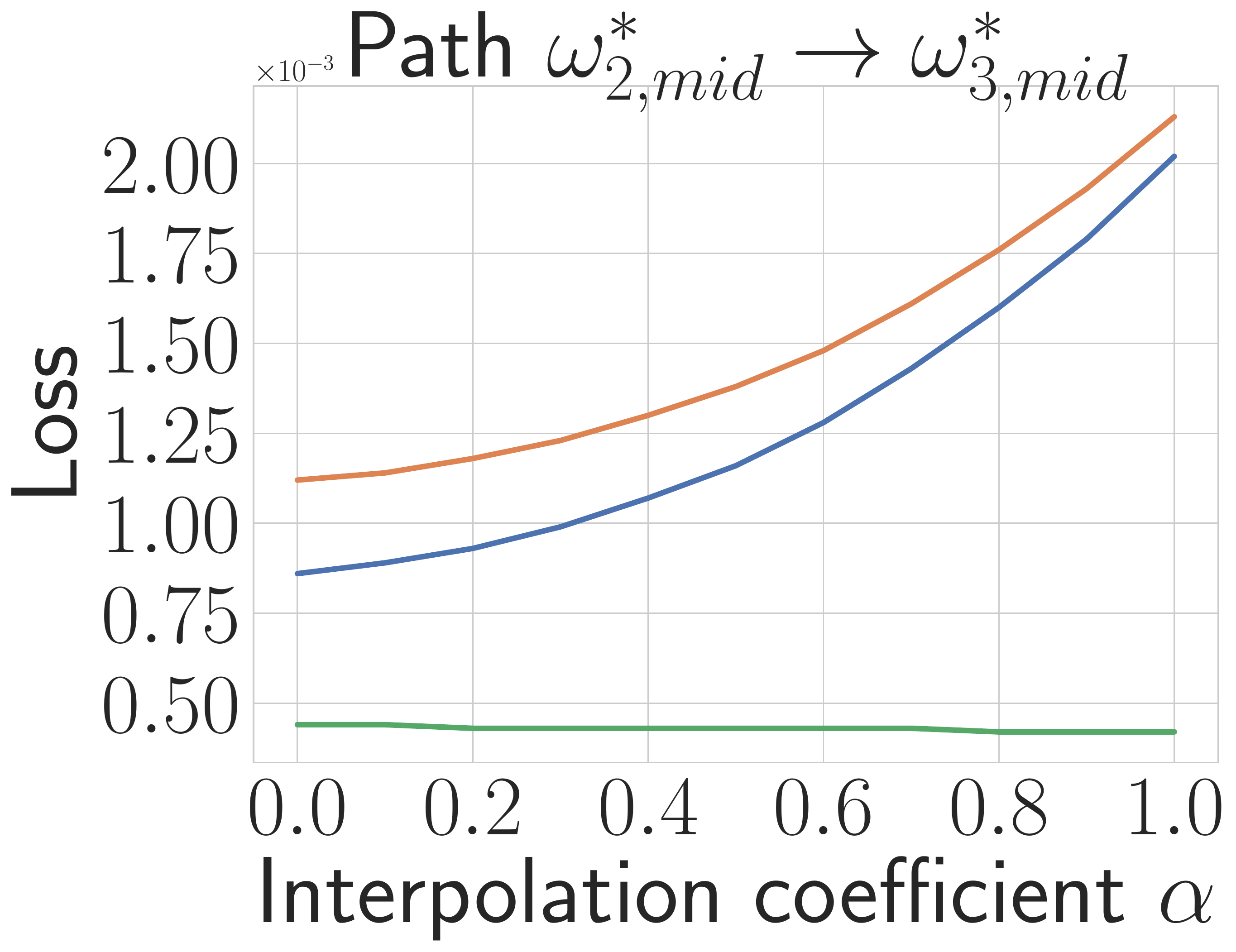}
    \caption{}
    \label{fig:connectivity_loss_2}
\end{subfigure}
\hfill
\begin{subfigure}{.32\textwidth}
    \centering
    \includegraphics[width=1.0\textwidth,height=0.7\textwidth,keepaspectratio=True]{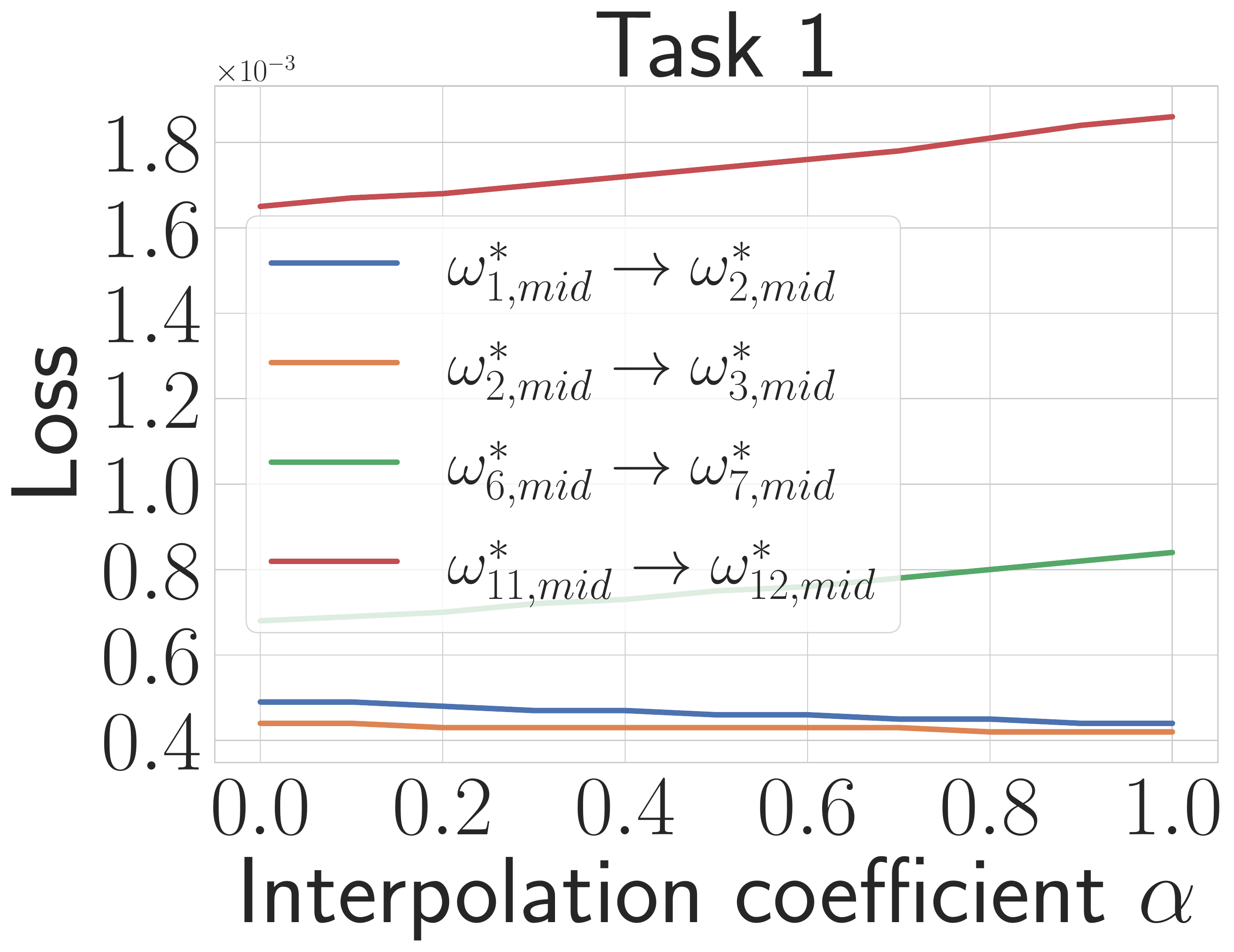}
    \caption{}
    \label{fig:connectivity_vs_ours}
\end{subfigure}

\caption{While applying naively subspace (Subspace and Subspace + ER) does not allow mode connectivity (loss function is increasing between two subsequent solutions) (left), our method (denoted "Ours") allows linear mode connectivty between successive solution'.}
\label{fig:connectivity}
\end{figure*}

\clearpage

\clearpage
\negspace{-2mm}
\section{Additional results and setup details}

\label{sec:appendix-experiments-results}
In this section, we first provide our experimental setup and benchmarks used. We then check if our method connects the subspace of previous solutions together and finally compare our algorithm \algoname against the main baselines of the literature.

\subsection{Experimental setup}
\label{sec:experimental_setup}
\paragraph{Setup}
The experimental setup, such as benchmarks, network architectures, continual learning setting (e.g., number of tasks, episodic memory size, and training epochs per task), hyper-parameters, and evaluation metrics are chosen to be similar to several other studies~\cite{AGEM,understanding_continual,Chaudhry2019OnTE,farajtabar2020orthogonal}. For all experiments, we report the average and standard deviation over five runs with different random seeds. 

\paragraph{Benchmarks}
We use the standard benchmarks similarly to \cite{Goodfellow2013Forgetting} and \cite{Chaudhry2019OnTE}.
Permuted-MNIST \cite{Goodfellow2013Forgetting} consists of a series of MNIST supervised learning tasks,
where the pixels of each task are permuted with respect to a fixed permutation.
Rotation-MNIST \cite{farajtabar2020orthogonal} consists of a series of MNIST classification tasks, where the images are rotated with respect to a fixed angle, monotonically. For both MNIST dataset the task ID does not need to be provided since it is a domain incremental task.
We increment the rotation angle by $9$ degrees at each new task.
Split CIFAR-100 \cite{Chaudhry2019OnTE} is constructed by splitting the original CIFAR-100 dataset
\cite{Krizhevsky2009LearningML} into 20 disjoint subsets. In order to assess the robustness to catastrophic forgetting over long tasks sequences, all datasets have $20$ tasks. Split miniImageNet is a variant of the ImageNet
dataset~\cite{ILSVRC15}, also splitted in 20 disjoint subsets where each subset is formed by sampling without replacement of 5 classes out of 100. Both CIFAR-100 and miniImageNet contains $20$ tasks with $500$ samples for each of the $5$ classes and request the task ID to be provided as input to the model.

\paragraph{Architectures} For MNIST dataset, we used a fully connected neural networks with two hidden layers of $256$  ReLU hidden units as in ~\cite{AGEM,understanding_continual,Chaudhry2019OnTE,farajtabar2020orthogonal}. For Split CIFAR-100, we used the same reduced Resnet18 as in~\cite{understanding_continual} (with three times less features maps accros all layers). For Split miniImageNet, we adapted the network used with CIFAR-100 by adapting the input dimension of the last fully connected layers since both dataset have different input dimensions ((3,84,84) for miniImageNet versus (3,32,32) for CIFAR-100).

\paragraph{Evaluation metrics}
\label{paragraph:evalation_metrics}
To assess the performance of each baseline, we report two metrics used in the literature which are the \textbf{Final Accuracy} and \textbf{Forgetting Measure}.
The Final Accuracy after $T$ tasks is the average validation accuracy over all the tasks $\tau=1...T$ defined as: $A_T=\displaystyle{\frac{1}{T}\sum_{\tau=1}^{T}a_{T,\tau}}$ where $a_{T,\tau}$ is the validation accuracy of task $\tau$ after the model finished learning on task $T$. The Forgetting Measure is defined as:
$F_T=\frac{1}{T-1}\displaystyle{\sum_{\tau=1}^{T-1} \max_{t=\{1..T-1 \}} (a_{t,\tau}-a_{T,\tau}})$


\subsection{Ensemble model ablations}
This section provides various ablation experiments between Scaled MC-SGD, Ensemble MCSGD and \algoname:
\begin{itemize}
    \item Table~\ref{tab:varying_n} shows the performance for \algoname with different number of models $n$
    \item Table~\ref{tab:parameters_cifar} compare \algoname against Scaled single model and Ensemble MC-SGD varying: number of models $n$, parameters and for different level of compute cost (FLOPS) and different strategies for Vanilla Ensemble (bagging and not bagging). 

\end{itemize}

\label{sec:comparison_ensembles}
 
\paragraph{Scaled MC-SGD} In order to compare to the performance on a single model ($n=1$), we increase its capacity to match the number of parameters of the ensemble methods. While fully connected networks (MNIST dataset), we increase the number of hidden units ($h=256,450,600$), for the Restnet18 network (Split CIFAR-100), we increase the number of channels ($nf=20,29,35$).

\paragraph{Bagging strategy for Ensemble MC-SGD}  Every time a data batch arrives, we sample uniformly one model among the other to receive the data and be updated. This allows to compare Ensemble MC-SGD at the same level of compute as MC-SGD since each member of the ensemble get to see $1/n$ of the whole dataset.

\paragraph{Calculation of the FLOPS} The Floating Point Operation per Second is a metric to quantify the computational cost for an algorithm. Our reported FLOPS metric represents a forward pass cost given a  batch size of $10$. To have an idea of the total training cost, one can approximate the backward pass as $\sim 2$ times the forward pass cost ~\cite{Kaplan2019NotesOC}. The relative FLOPS is the ratio between our reported Inference (forward pass) FLOPS for a given algorithm and its Single version, i.e the reference algorithm is MC-SGD in our case.

\begin{table*}[h!]
\centering

\resizebox{\textwidth}{!}{%
\begin{tabular}{lcccccccc}
\hline
\multirow{2}{*}{\textbf{Number of model}} &
  \multicolumn{2}{c}{\textbf{Permuted MNIST}} &
  \multicolumn{2}{c}{\textbf{Rotated MNIST}} &
  \multicolumn{2}{c}{\textbf{Split CIFAR-100}} &  \multicolumn{2}{c}{\textbf{Split miniImageNet}}\\ \cline{2-9} 
        & Accuracy $\uparrow$ & Forgetting $\downarrow$&Accuracy   $\uparrow$ & Forgetting  $\downarrow$ & Accuracy   $\uparrow$ & Forgetting  $\downarrow$   & Accuracy   $\uparrow$ & Forgetting  $\downarrow$  \\ \hline
2        & 87.1 ($\pm$0.19) & 0.07 ($\pm$0.01) & 86.6 ($\pm$0.45) & 0.07 ($\pm$0.01) & 60.97 ($\pm$1.53) & 0.05 ($\pm$0.01)  & 57.10 ($\pm$1.53) & 0.05 ($\pm$0.01) \\
  3 & \multicolumn{1}{l}{87.8 ($\pm$0.53)} &0.06 ($\pm$0.01) & \multicolumn{1}{l}{86.7 ($\pm$0.67)} & 0.07 ($\pm$0.01) &61.74 ( $\pm$0.80) & 0.05 ($\pm$ 0.01)  & 58.17 ($\pm$0.84) & 0.03  ($\pm$0.01) \\
5         & 87.8 ($\pm$0.3) & 0.07 ($\pm$0.01) & 86.8 ($\pm$0.46) & 0.07 ($\pm$0.01) & 60.85 ($\pm$0.73) & 0.05 ($\pm$ 0.01) & 58.11 ($\pm$1.23) & 0.03 ($\pm$ 0.01)  \\ \hline
\end{tabular}%

}
\centering
\caption{Final accuracy for subspace method with a different number of models $n$. For $n \geq 2$ models, there is not much difference in performance for MNIST dataset while for Split CIFAR-100, $n=3$ gives the best performance.}
\label{tab:varying_n}
\end{table*}

\begin{table*}[h!]
\centering
\resizebox{\textwidth}{!}{%
\begin{tabular}{lccccc}
\hline
\multirow{2}{*}{\textbf{Number of model}} & \multicolumn{2}{c}{\textbf{Permuted MNIST}} &
  \multicolumn{2}{c}{\textbf{Rotated MNIST}}   &  \multirow{2}{*}{\textbf{Relative FLOPS ratio}}
 \\ \cline{2-5} 
        & Accuracy $\uparrow$ & Forgetting $\downarrow$&Accuracy   $\uparrow$ & Forgetting  $\downarrow$   &\\ \hline
\textcolor{blue}{$|\theta|=268 K$}         &  &  &  &  &  \\  \hline
Scaled MC-SGD ($256$)       & 83.17 ($\pm$0.63) & 0.10 ($\pm$0.01) & 81.70 ($\pm$0.43) & 0.08 ($\pm$0.01)   &  $1$\\ \hline

\textcolor{blue}{$|\theta|=560 K$}         &  &  &  &    \\  \hline
Scaled MC-SGD ($450$)         & 86.80 ($\pm$0.93) & 0.07 ($\pm$0.01) & 82.67 ($\pm$0.20) & 0.08 ($\pm$0.01 )   & $2.1$\\
Ensemble MC-SGD ($n=2$, bagging$^{*}$)           & 86.23 ($\pm$0.52) & 0.06 ($\pm$0.01) & 80.7 ($\pm$0.50) & 0.09 ($\pm$0.01)   &  $1$\\
Ensemble MC-SGD ($n=2$)         & 86.5 ($\pm$0.33) & 0.08 ($\pm$0.01) & 83.00 ($\pm$0.45) & 0.08 ($\pm$0.01)  &   $2$\\
\algoname ($n=2$) & \multicolumn{1}{l}{87.1 ($\pm$0.19)} &0.07 ($\pm$0.01) & \multicolumn{1}{l}{86.6 ($\pm$0.45)} & 0.07 ($\pm$0.01)  &  $1.1$  \\  \hline

\textcolor{blue}{$|\theta|=836 K$}         &  &  &  &    & \\  \hline
Scaled MC-SGD ($600$)         & 88.03 ($\pm$0.36) & 0.06 ($\pm$0.01) & 83.67 ($\pm$0.40) & 0.07 ($\pm$0.01)  &  \textcolor{red}{$3.1$} \\
Ensemble MC-SGD ($n=3$, bagging)              & 86.60 ($\pm$0.46) & 0.05 ($\pm$0.01) & 80.00 ($\pm$0.24) & 0.08 ($\pm$0.01)  & $1$ \\ 
Ensemble MC-SGD ($n=3$)              & 88.30 ($\pm$0.48) & 0.06 ($\pm$0.01) & 83.63 ($\pm$0.39) & 0.07 ($\pm$0.01) &   \textcolor{red}{$3$}\\ 
\algoname ($n=3$) & \multicolumn{1}{l}{87.8 ($\pm$0.30)} &0.07 ($\pm$0.01) & \multicolumn{1}{l}{86.7 ($\pm$0.67)} & 0.07 ($\pm$0.01)      & $1.2$ \\\hline
\end{tabular}%
}

\end{table*}

\begin{table*}[h!]
\centering

\resizebox{\textwidth}{!}{%
\begin{tabular}{lccccccc}
\hline
\multirow{2}{*}{\textbf{Number of model}} &  \multicolumn{2}{c}{\textbf{Split CIFAR-100}} 
 & \multicolumn{2}{c}{\textbf{Split miniImageNet}}
     & \multirow{2}{*}{\textbf{Relative FLOPS ratio}}  \\ \cline{2-5} 
    & Accuracy $\uparrow$ & Forgetting $\downarrow$     & Accuracy $\uparrow$ & Forgetting $\downarrow$ &   \\ \hline
\textcolor{blue}{$|\theta|=500 K$}         &  &  &  & \\  \hline
Scaled MC-SGD ($nf=20$)       & 58.22 ($\pm$0.91) & 0.08 ($\pm$0.01) &  54.80 ($\pm$1.04) & 0.05 ($\pm$0.01) & $1$  \\ \hline
\textcolor{blue}{$|\theta|=1 M$}         &  &  &  &    \\  \hline
Scaled MC-SGD ($nf=29$)         & 60.12 ($\pm$0.97) & 0.07 ($\pm$0.01) &  55.30 ($\pm$0.83) & 0.06 ($\pm$0.01) &  $2.08$\\
Ensemble MC-SGD ($n=2$, bagging)      &    56.87 ($\pm$0.80) & 0.06 ($\pm$0.01) &   54.00 ($\pm$0.77) & 0.05 ($\pm$0.01) &  $1$ \\
Ensemble MC-SGD ($n=2$)    &  60.83 ($\pm$0.99) & 0.09 ($\pm$0.01)    &  57.60 ($\pm$0.55) & 0.04 ($\pm$0.01)    & $2$\\
\algoname ($n=2$) &  \multicolumn{1}{l}{ 60.97 ($\pm$1.53)} & 0.05 ($\pm$0.01)&   \multicolumn{1}{l}{ 57.10 ($\pm$0.79)} & 0.05 ($\pm$0.01)& $\approx 1.00$ \\  \hline
\textcolor{blue}{$|\theta|=1.5 M$}         &  &  &  &    \\  \hline
Scaled MC-SGD ($nf=35$)         & 60.50 ($\pm$0.84) & 0.06 ($\pm$0.01)  & 55.44 ($\pm$1.36) & 0.05 ($\pm$0.01) & \textcolor{red}{$3.01$}\\
Ensemble MC-SGD ($n=3$, bagging)      &   55.48 ($\pm$1.20) & 0.06 ($\pm$0.01)    &  52.39 ($\pm$0.60) & 0.03 ($\pm$0.01)   &   $1$ \\
Ensemble MC-SGD ($n=3$)    &   64.12 ($\pm$1.16) & 0.06 ($\pm$0.01)   &   59.1 ($\pm$1.1) & 0.04 ($\pm$0.01) & \textcolor{red}{$3$}\\
\algoname ($n=3$) & \multicolumn{1}{l}{61.74 ( $\pm$0.80)} & 0.05 ($\pm$ 0.01)  &  \multicolumn{1}{l}{58.17 ( $\pm$0.84)} & 0.03 ($\pm$ 0.01)  &  $\approx 1.00$ \\\hline
\end{tabular}%

}
\caption{ Performance comparison against Scaled MC-SGD, Ensemble MC-SGD and \algoname. Ensemble MC-SGD gets the best performance but at a high compute cost. However, if we compare with a fair compute cost (see bagging) Ensemble MC-SGD performs worst than \algoname.}
\label{tab:parameters_cifar}
\end{table*}

\clearpage

\clearpage
\section{Baselines hyperparameters}
\label{sec:final_exp_parameters}
We first enumerate the hyperparameters used for the $20$ tasks experiments in Table~\ref{tab:main_results} before describing in detail the ablation in this section. 

\subsection{Hyper-Parameters}
For the experiment in Section~\ref{sec:5-experiments-results}, we have used the following grid for each model. We note that for other algorithms (e.g., A-GEM, and EWC), we ensured that our grid contains the optimal values that the original papers reported. If applicable, all baselines used a buffer memory of $1$ element per class per task (which translates in a total replay buffer memory of $200$ for MNIST dataset and $100$ for CIFAR-100 and miniImageNet). We used the same single training epoch per task setting as in~\cite{Chaudhry2019OnTE,understanding_continual,mirzadeh2020dropout}.
\vspace{3mm}
\subsubsection*{Naive SGD}
\begin{itemize}
    \item learning rate:  [0.25, \textbf{0.1} (MNIST), \textbf{0.03} (miniImageNet), \textbf{0.01} (CIFAR-100), 0.001]
    \item batch size: 10
\end{itemize}
\vspace{3mm}
\subsubsection*{EWC}
\begin{itemize}
    \item learning rate: [0.25, \textbf{0.1} (MNIST, CIFAR-100), 0.01, 0.001]
    \item batch size: [64, \textbf{10}]
    \item $\lambda$ (regularization): [100, \textbf{10} (MNIST, CIFAR-100), 1]
\end{itemize}
\vspace{3mm}

\subsubsection*{A-GEM}
    \begin{itemize}
        \item learning rate: [0.1, \textbf{0.1} (MNIST), \textbf{0.01} (CIFAR-100), 0.001]
        \item batch size: [64, \textbf{10}]
    \end{itemize}
    
\vspace{3mm}
\subsubsection*{ER-Reservoir}
    \begin{itemize}
        \item learning rate: [0.25, \textbf{0.1} (MNIST, miniImageNet), \textbf{0.01} (CIFAR-100), 0.001]
        \item batch size: [64, \textbf{10}]
    \end{itemize}

\subsubsection*{Batch Ensemble}
Although \cite{Wen2020BatchEnsemble} reported an optimal value for an ensemble size of $n=4$, we found out in our setting that $n=2$ provided the best results.
\begin{itemize}
    \item learning rate: [0.25, \textbf{0.1} (CIFAR-100,miniImageNet, Permuted), 0.01 (Rotated), 0.001]
   
    \item learning rate decay: [0.95, \textbf{0.9} (CIFAR-100, miniImageNet), 0.85,\text{0.8} (MNIST)]
     \item batch size: [128, 64, 32, \textbf{10}]
\end{itemize}
\vspace{3mm}

\vspace{3mm}
\subsubsection*{Stable SGD}
\begin{itemize}
    \item initial learning rate: [0.25, \textbf{0.1} (MNIST, CIFAR-100), 0.01, 0.001]
    \item learning rate decay: [0.95, \textbf{0.9}(CIFAR-100), 0.85,\text{0.8} (miniImageNet), \textbf{0.6}(MNIST)]
    \item batch size: [64, \textbf{10}]
    \item dropout: [\textbf{0.25} (MNIST), 0.1,\textbf{0.0} (CIFAR-100, miniImageNet)]
\end{itemize}
\vspace{3mm}
\subsubsection*{Mode Connectivity SGD}
To obtain continual minima (i.e., $\hat{\omega}^{*}_{1}$ to $\hat{\omega}^{*}_{20}$), we use the following hyperparameters:
\begin{itemize}
    \item initial learning rate: [0.25, \textbf{0.1} (MNIST, CIFAR-100), 0.01, 0.001]
     \item momentum: [0.9, 0.85,  \textbf{0.8} (MNIST), \textbf{0.7} (miniImageNet), \textbf{0.4} (CIFAR-100)]
    \item learning rate decay: [\textbf{0.95} (Rotated MNIST, CIFAR-100), \textbf{0.9} (miniImageNet), 0.85, \textbf{0.8} (Permuted MNIST), 0.7 ]
    \item batch size: [\textbf{64} (MNIST), 32, \textbf{10} (CIFAR-100, miniImageNet)]
    \item dropout: [\textbf{0.25} (Permuted MNIST), 0.1, \textbf{0.0} ( Rotated MNIST, CIFAR-100, miniImageNet, )]
\end{itemize}

To  obtain $\bar{\omega}^{*}_1$ to $\bar{\omega}^{*}_{20}$, we use the following grid:
\begin{itemize}
    \item number of samples: [10, \textbf{5}, 3] for both MNIST and CIFAR experiments.
    \item learning rate: [0.2, 0.1, \textbf{0.05} (MNIST), \textbf{0.01} (CIFAR-100, miniImageNet), 0.001].
\end{itemize}
\vspace{3mm}
\subsubsection*{\algoname}
To obtain the subspace solution from the first step $\{ \hat{\omega}^{*}_{i}\}_{i=1}^{n}$ from Eq~\ref{eq:w_hat}, we used the following hyperparameters:
\begin{itemize}
    \item initial learning rate: [\textbf{0.3}\footnote{this value must be multiplied by the number of model $n$ to get the final learning rate used. Same logic is applied for miniImageNet dataset.} (CIFAR-100), 0.2, \textbf{0.15} (miniImageNet) , \textbf{0.1} (MNIST)]
     \item momentum: [0.9, 0.85,  \textbf{0.8} (Rotated MNIST), \textbf{0.4} (Permuted MNIST), \textbf{0} (CIFAR-100, miniImageNet)]
    \item learning rate decay: [\textbf{0.95} (Rotated MNIST, CIFAR-100, miniImageNet), 0.9, \textbf{0.8} (Permuted MNIST)]
    \item batch size: [32, \textbf{10} ( MNIST, CIFAR-100, miniImageNet)]
    \item dropout: [\textbf{0.25} (Permuted MNIST), 0.1, \textbf{0.0} ( Rotated MNIST, CIFAR-100, miniImageNet)]
\end{itemize}
The subspace connectivity steps leading to $\{ \omega^{*}_{i}\}_{i=1}^{n}$ ( Eq~\ref{eq:w_bar}) used the following hyperparameters:
\begin{itemize}
    \item number of samples: [10, \textbf{5} (MNIST), \textbf{3} (CIFAR-100, miniImageNet)] 
    \item learning rate: [0.2, \textbf{0.1} (CIFAR-100), \textbf{0.05} (MNIST, miniImageNet), ].
\end{itemize}

\paragraph{Implementation details of \algoname}
\label{sec:\algoname}

While the first loss (Eq.~\ref{eq:w_hat}) is a fine-tuning on the incoming task $\tau-1$ (Vanilla SGD), the second one (Eq.~\ref{eq:w_bar}) is done in two steps. First, we initialize the weight using a convex combination of weights around the two former midpoints as $\bar{\omega}_{i}=\alpha \omega^{*}_{\tau-1,mid}+ (1-\alpha)\hat{\omega}_{\tau,mid}$ then we add multiplicative noise $\bar{\omega}_{i} * \epsilon , \quad i=1...n$, $\epsilon \sim \mathcal{N}(1,\sigma)$ (with $\sigma=0.005$ for MNIST and $\sigma=0.01$ for CIFAR-100 and miniImageNet) where $*$ represents element-wise multiplication. The multiplicative noise has the nice property to scale well with the weights magnitude. The $\alpha$ values taken are:  [0.9, \textbf{0.85} (Rotated MNIT), \textbf{0.8} (CIFAR-100),  \textbf{0.7} (miniImageNet), \textbf{0.25} (Permuted MNIST),]

\begin{table*}[h!]
\centering
\resizebox{\textwidth}{!}{%
\begin{tabular}{lcccccc}
\hline
\multirow{2}{*}{\textbf{Method}} &
  \multicolumn{2}{c}{\textbf{Permuted MNIST}} &
  \multicolumn{2}{c}{\textbf{Rotated MNIST}} &
  \multicolumn{2}{c}{\textbf{Split CIFAR-100}} \\ \cline{2-7} 
        & Accuracy $\uparrow$ & Forgetting $\downarrow$&Accuracy   $\uparrow$ & Forgetting  $\downarrow$ & Accuracy   $\uparrow$ & Forgetting  $\downarrow$  \\ \hline
Naive SGD           & 44.4 ($\pm$2.46) & 0.53 ($\pm$0.03) & 46.3 ($\pm$1.37) & 0.52 ($\pm$0.01) & 40.4 ($\pm$2.83) & 0.31 ($\pm$0.02) \\
EWC \text{\citep{EWC}}           & 70.7 ($\pm$1.74) & 0.23 ($\pm$0.01) & 48.5 ($\pm$1.24) & 0.48 ($\pm$0.01) & 42.7 ($\pm$1.89) & 0.28 ($\pm$0.03)\\
A-GEM \text{\citep{AGEM}}              & 65.7 ($\pm$0.51) & 0.29 ($\pm$0.01) & 55.3 ($\pm$1.47) & 0.42 ($\pm$0.01) & 50.7 ($\pm$2.32) & 0.19 ($\pm$0.04) \\
ER-Reservoir  \text{\citep{Chaudhry2019OnTE}}      & 72.4 ($\pm$0.42) & 0.16 ($\pm$0.01) & 69.2 ($\pm$1.10) & 0.21 ($\pm$0.01) & 46.9 ($\pm$0.76) & 0.21 ($\pm$0.03) \\
Stable SGD  \text{\citep{understanding_continual}}       & 80.1 ($\pm$0.51) & 0.09 ($\pm$0.01) & 70.8 ($\pm$0.78) & 0.10 ($\pm$0.02) & 56.9 ($\pm$1.52) & 0.11 ($\pm$0.01) \\
MC-SGD \text{\citep{mirzadeh2021MCSGD}} &
  \multicolumn{1}{l}{82.9 ($\pm$0.40)} &0.10 ($\pm$0.01) & 81.9 ($\pm$0.46) & 0.08 ($\pm$0.01) &58.2 ($\pm$0.91) & 0.08 ($\pm$0.01) \\   \hline
    Ensemble MC-SGD  & 88.30 ($\pm$0.48) & 0.06 ($\pm$0.01) & 83.63 ($\pm$0.39) & 0.07 ($\pm$0.01)  & 64.12 ($\pm$1.16) & 0.06 ($\pm$0.01) \\
  Batch Ensemble \text{\citep{Wen2020BatchEnsemble}}  & 63.37 ($\pm$1.32)  & 0.27 ($\pm$0.01) & 57.54 ($\pm$0.53)  & 0.17 ($\pm$0.01)  & 53.08 ($\pm$1.70)  & 0.08 ($\pm$0.01) \\

  \algoname  (ours) &
  \multicolumn{1}{l}{87.8 ($\pm$0.53)} &0.06 ($\pm$0.01) & \multicolumn{1}{l}{86.7 ($\pm$0.67)} & 0.07 ($\pm$0.01) &61.7 ($\pm$0.80) & 0.05 ($\pm$0.01)\\ \hline
Multitask Learning & 89.5 ($\pm$0.21) & 0.0              & 89.8($\pm$0.37)  & 0.0              & 66.8($\pm$1.42)  & 0.0            \\ \hline

\end{tabular}%
}
\caption{Comparison between the proposed method (\algoname) and other single model baselines on $5$ random seeds ($\pm$ std).}
\label{tab:main_results}
\end{table*}

\begin{figure}[t!]
\centering
\begin{subfigure}{.32\textwidth}
    \centering
    \includegraphics[width=0.9\textwidth,height=0.8\textwidth,keepaspectratio=True]{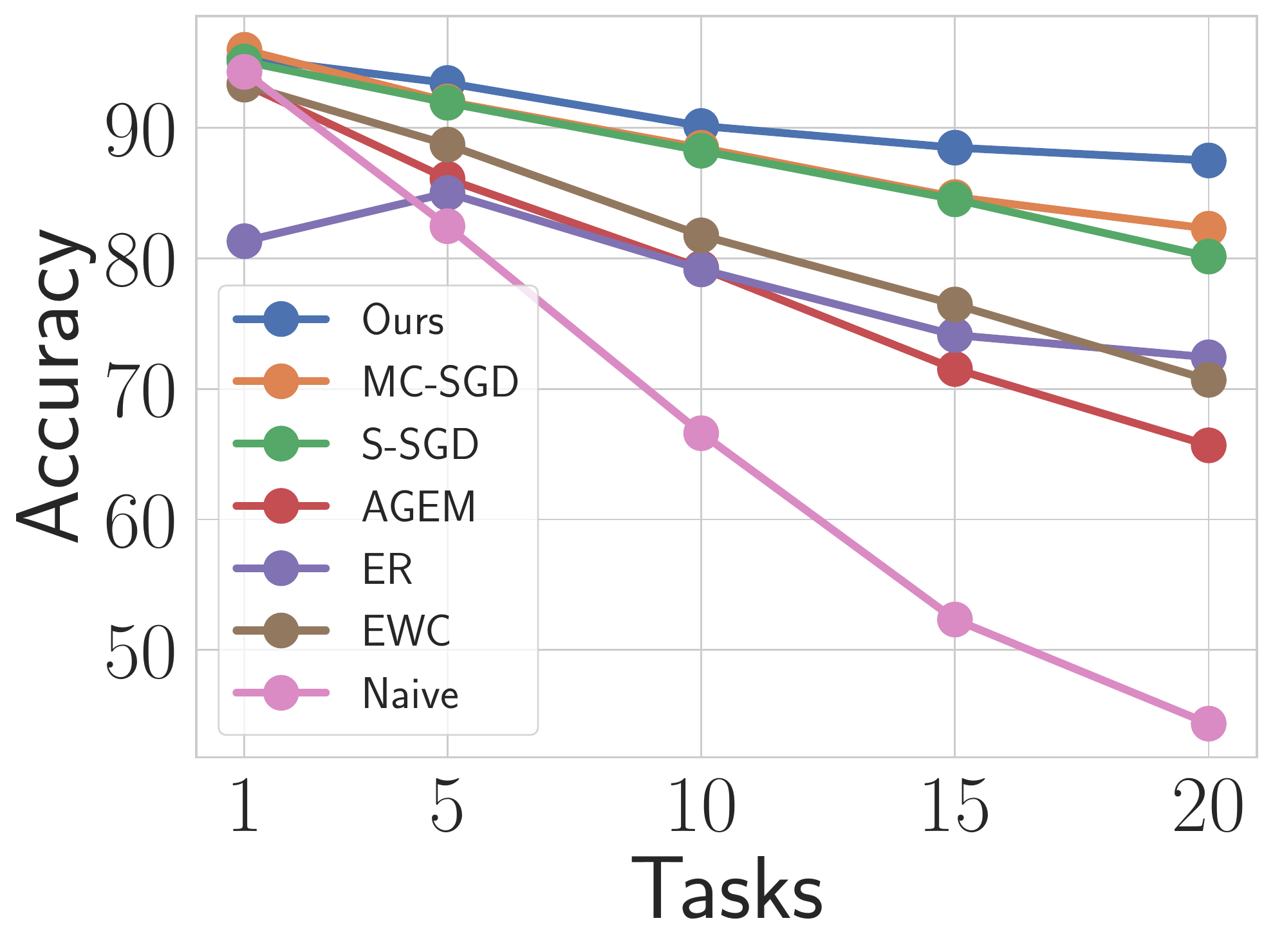}
    \caption{Permuted MNIST}
    \label{fig:curve_permuted}
\end{subfigure}\hfill
\begin{subfigure}{.32\textwidth}
    \centering
    \includegraphics[width=0.9\textwidth,height=0.8\textwidth,keepaspectratio=True]{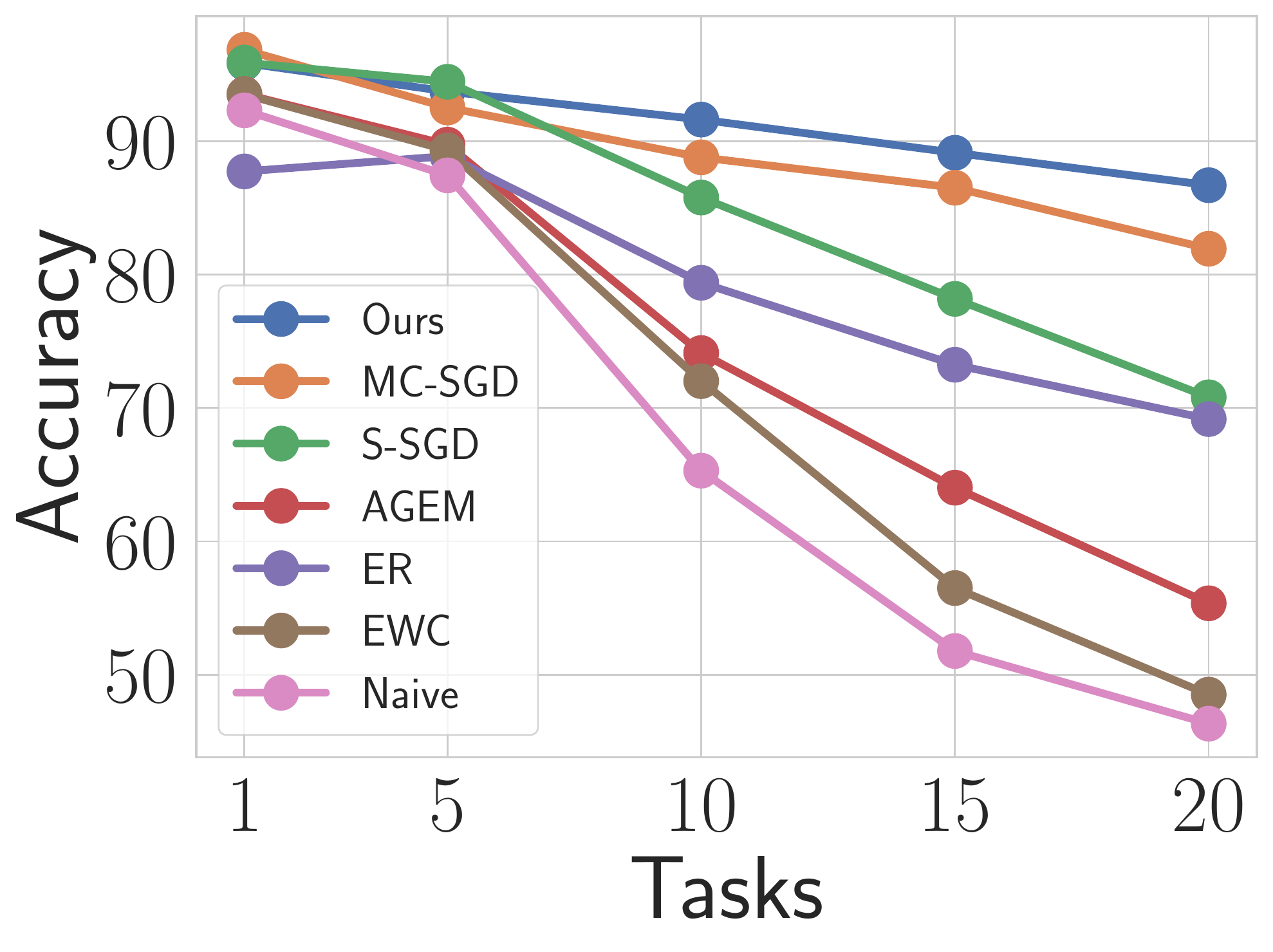}
    \caption{Rotated MNIST}
    \label{fig:curve_rotated}
\end{subfigure}
\hfill
\begin{subfigure}{.32\textwidth}
    \centering
    \includegraphics[width=0.9\textwidth,height=0.8\textwidth,keepaspectratio=True]{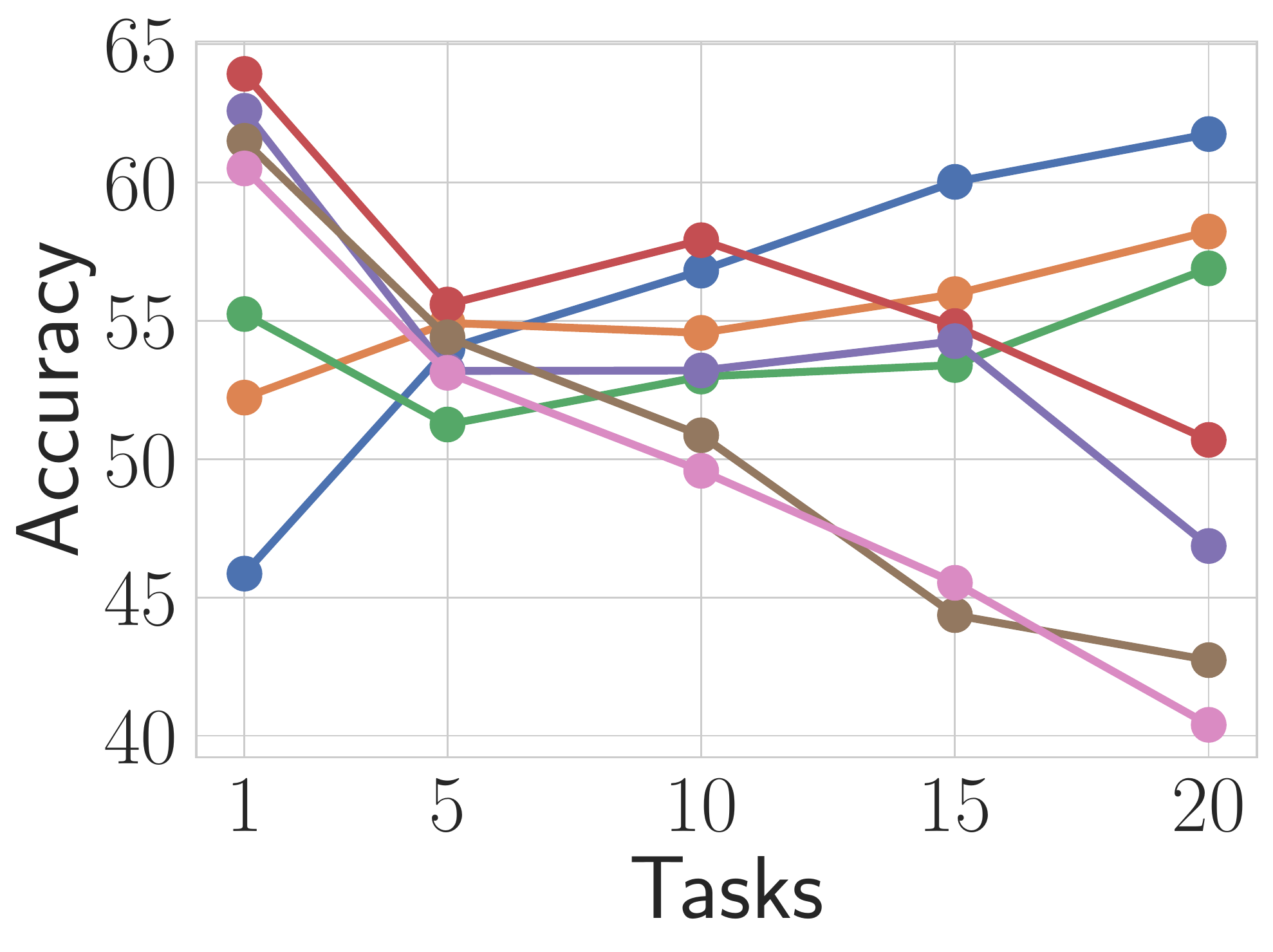}
    \caption{Split CIFAR-100}
    \label{fig:curve_cifar}
\end{subfigure}
\caption{Evolution of the average accuracy throughout the training.}
\label{fig:evolution_performance}
\end{figure}

\begin{table*}[h!]
\centering

\begin{tabular}{lcc}
\hline
\multirow{2}{*}{\textbf{Method}} &
  \multicolumn{2}{c}{\textbf{Split miniImageNet}} \\ \cline{2-3} 
        & Accuracy $\uparrow$ & Forgetting $\downarrow$ \\ \hline
Naive SGD                                                 & 43.66 ($\pm$1.65) & 0.22 ($\pm$0.02)  \\
  ER Reservoir \text{\citep{Chaudhry2019OnTE}}             & \multicolumn{1}{l}{51.7 ($\pm$2.53)} &0.11 ($\pm$0.02)  \\
  Stable SGD   \text{\citep{understanding_continual}}     & \multicolumn{1}{l}{53.76 ($\pm$1.13)} &0.07 ($\pm$0.01)  \\
  MC-SGD \text{\citep{mirzadeh2021MCSGD}}                 & \multicolumn{1}{l}{54.80 ($\pm$1.04)} &0.05 ($\pm$0.01)  \\\hline
  Ensemble MC-SGD  & \multicolumn{1}{l}{59.2 ($\pm$1.1)} &0.04 ($\pm$0.01)    \\
  Batch Ensemble \text{\citep{Wen2020BatchEnsemble}}                 & \multicolumn{1}{l}{51.78 ($\pm$1.74)} &0.05 ($\pm$0.01)  \\
    \algoname (ours)                                            & \multicolumn{1}{l}{58.17 ($\pm$0.84)} & 0.03 ($\pm$0.01)   \\
    \hline
Multitask Learning                                        & 62.82 ($\pm$1.77) & 0.0 \\ \hline
\end{tabular}%

\centering
\caption{Comparison between \algoname and other baselines on 5 random seeds for Split miniImageNet.}
\label{tab:image_results}
\end{table*}